\documentclass[runningheads]{llncs}

\usepackage[mobile]{eccv}

\usepackage{eccvabbrv}

\usepackage{graphicx}
\usepackage{booktabs}

\usepackage[accsupp]{axessibility}  

\usepackage[pagebackref]{hyperref}

\usepackage{orcidlink}

\usepackage{amsmath} 	
\usepackage{amsfonts} 	
\usepackage{dsfont} 	
\usepackage[ruled]{algorithm2e} 
\usepackage{xfrac} 

\usepackage{soul}       
\usepackage{microtype} 	

\usepackage{gensymb} 	
\usepackage{units} 	    
\usepackage{pifont} 	

\usepackage{array} 	    
\usepackage{booktabs}
\usepackage{multirow} 	
\usepackage{makecell} 	
\usepackage{subcaption} 
\usepackage{wrapfig}  	
\usepackage{float} 	    
\usepackage[percent]{overpic}

\usepackage[export]{adjustbox}

\usepackage{enumitem}   

\usepackage[x11names, table]{xcolor} 	
\usepackage{tcolorbox}    
\usepackage{fancybox}
\usepackage{dashbox}
\usepackage{transparent}
\usepackage{arydshln}

\newcommand{\cmark}{\ding{51}}%
\newcommand{\xmark}{\ding{55}}%

\newcommand{\ours}{SyntheticDoc}

\newlength{\tmpintextsep}
\newlength{\tmpcolumnsep}
\tmpintextsep=\intextsep
\tmpcolumnsep=\columnsep

\definecolor{gold}{HTML}{ffd700}
\definecolor{silver}{HTML}{c0c0c0}
\definecolor{bronze}{HTML}{CD7F32}
\definecolor{other}{HTML}{FFFFFF}
\definecolor{original}{gray}{0.8}
\newcommand{\best}[1]{\setlength{\fboxsep}{0pt}\fbox{\setlength{\fboxsep}{2pt}\colorbox{gold!50}{#1}}}
\newcommand{\second}[1]{\setlength{\fboxsep}{0pt}\fbox{\setlength{\fboxsep}{2pt}\colorbox{silver!70}{#1}}}
\newcommand{\third}[1]{\setlength{\fboxsep}{0pt}\fbox{\setlength{\fboxsep}{2pt}\colorbox{bronze!50}{#1}}}
\newcommand{\other}[1]{\setlength{\fboxsep}{0pt}\fbox{\setlength{\fboxsep}{2pt}\colorbox{other}{#1}}}

\definecolor{originalborder}{rgb}{1,0,0}
\newcommand{\original}[1]{\setlength{\fboxsep}{0pt}\fcolorbox{original}{white}{\setlength{\fboxsep}{2pt}\colorbox{original}{#1}}}

\newsavebox{\largestimage}

\begin{document}

\title{\ours{}: A Large Synthetic Dataset for Document Unwarping and Illumination Correction} 

\titlerunning{\ours{}}

\newcommand\samethanks[1][\value{footnote}]{\footnotemark[#1]}
\author{Daniel Woortmann\thanks{Equal contribution.}\inst{1}\orcidlink{0009-0009-3959-5631} \and
Tanguy Magne\samethanks\inst{1}\orcidlink{0009-0001-0231-026X} \and
Olga Sorkine-Hornung\inst{1}\orcidlink{0000-0002-8089-3974}}

\authorrunning{D.~Woortmann et al.}

\institute{
\footnotesize{ETH Zurich, Department of Computer Science, \\Universitätstrasse 6, 8092 Zurich, Switzerland}\\
\email{d.woortmann@gmail.com}, 
\email{\{tanguy.magne,olga.sorkine\}@inf.ethz.ch}
}

\maketitle

\begin{abstract}
    Deep learning models have become the standard tool for document rectification and illumination correction, yet their performance is fundamentally bound by their training data. For nearly a decade, the community has heavily relied on Doc3D, a pioneering but increasingly limited document unwarping dataset in terms of scale and quality. 
    To address this bottleneck, we introduce \ours{}, a massive, high-quality dataset designed to push the boundaries of document unwarping. \ours{} is composed of 1,000,000 high-resolution procedurally generated training samples, alongside extensive validation and test sets. Each sample is paired with rich, pixel-perfect annotations, including UV maps, normal maps, albedo and shading. To ensure physical accuracy and photorealism, the paper geometries are generated via a physics-based simulator and rendered using a path tracer. 
    To demonstrate the benefit of our dataset, we train a simple baseline model on \ours{} and report on its performance in comparison to state-of-the-art methods on both document unwarping and illumination correction tasks.
    Our dataset is available at https://igl.ethz.ch/projects/SyntheticDoc/ and the code used to generate it at https://github.com/tanguymagne/SyntheticDoc.

    \keywords{Document unwarping \and Illumination correction \and Dataset}
\end{abstract}

\section{Introduction}
\label{sec:intro}

\begin{figure}[t]
    \centering
	\small
	\setlength{\tabcolsep}{0.5pt}
    \setlength{\fboxsep}{0pt}
    \begin{tabular}{cccccc}
      \includegraphics[width=0.163\linewidth]{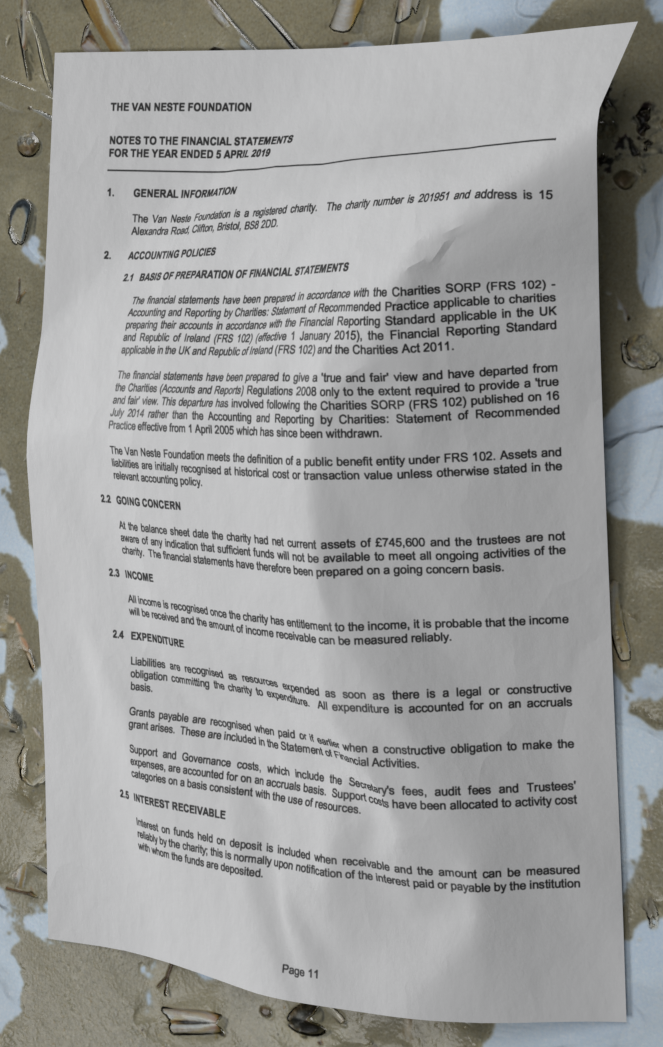}
    & \includegraphics[width=0.163\linewidth]{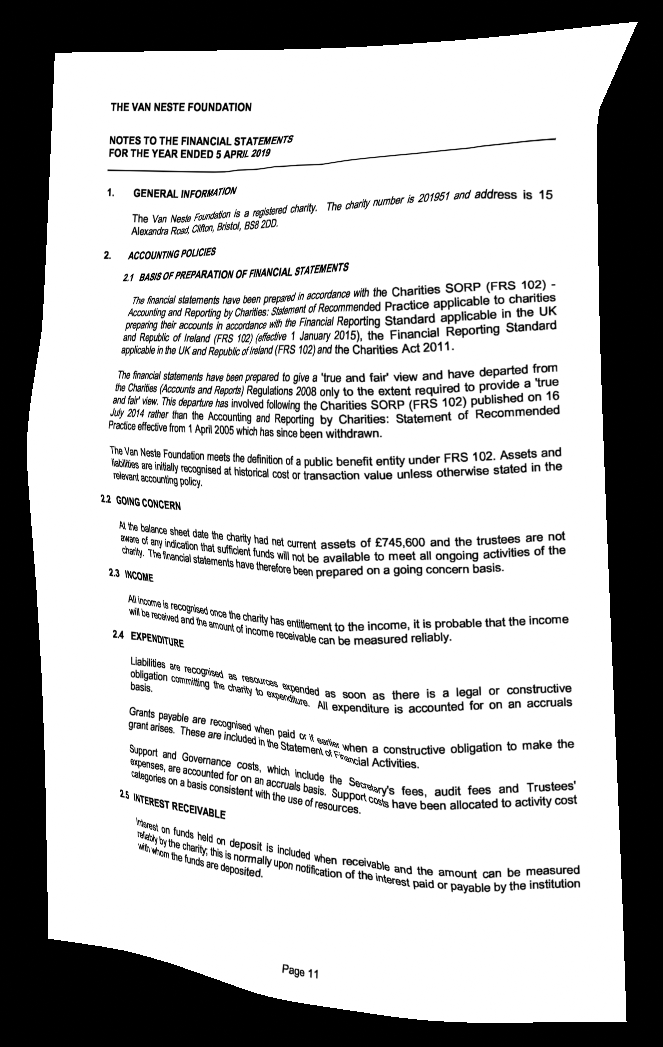}
    & \includegraphics[width=0.163\linewidth]{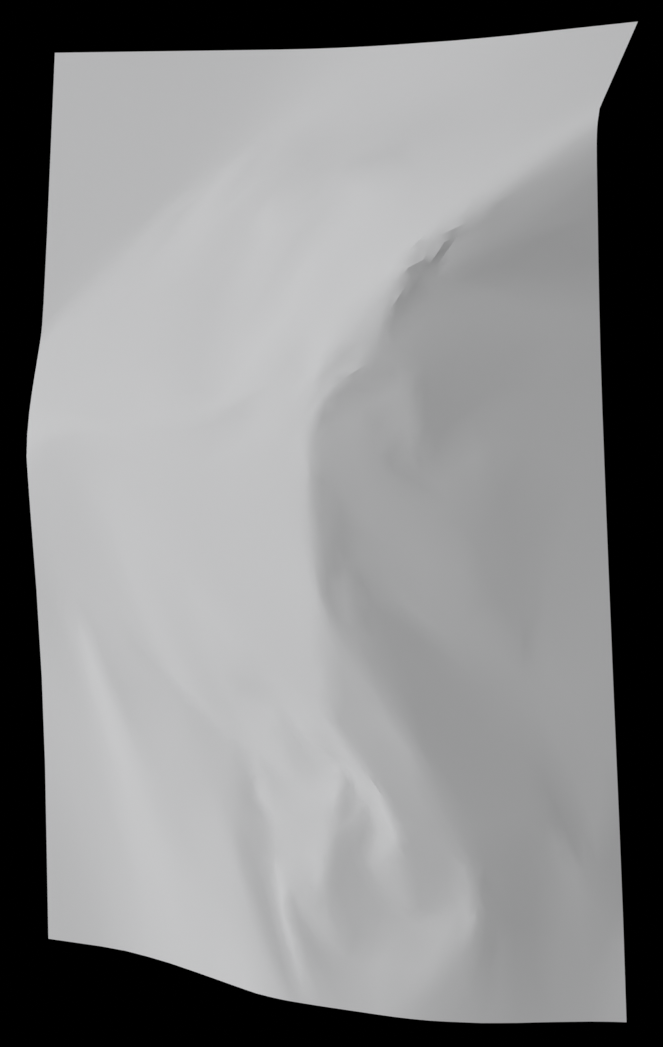}
    & \includegraphics[width=0.163\linewidth]{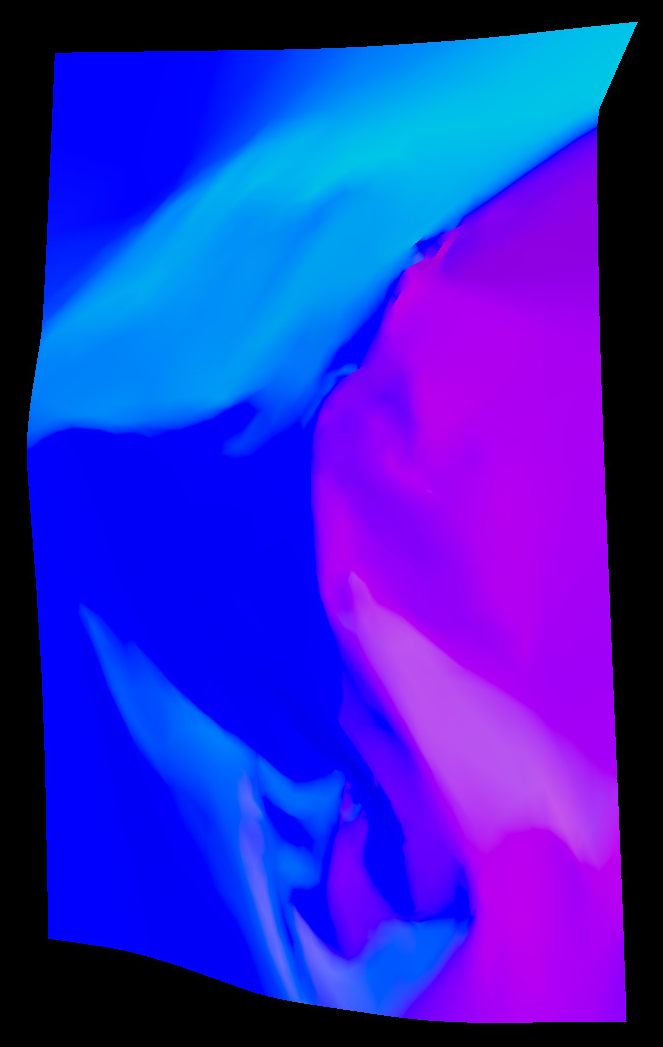}
    & \includegraphics[width=0.163\linewidth]{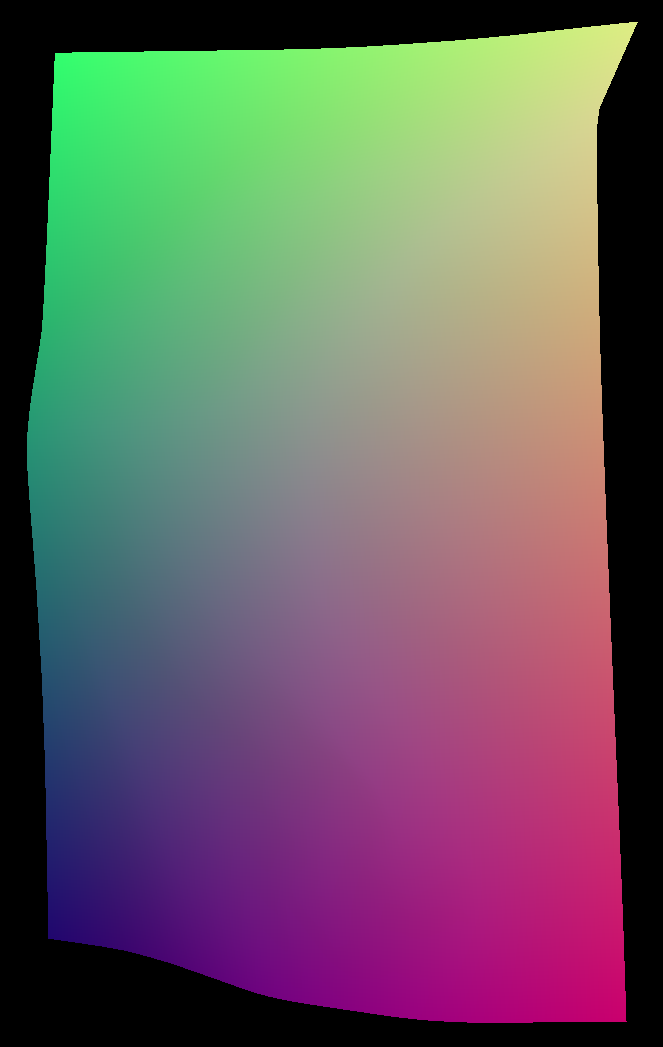}
    & \includegraphics[width=0.163\linewidth]{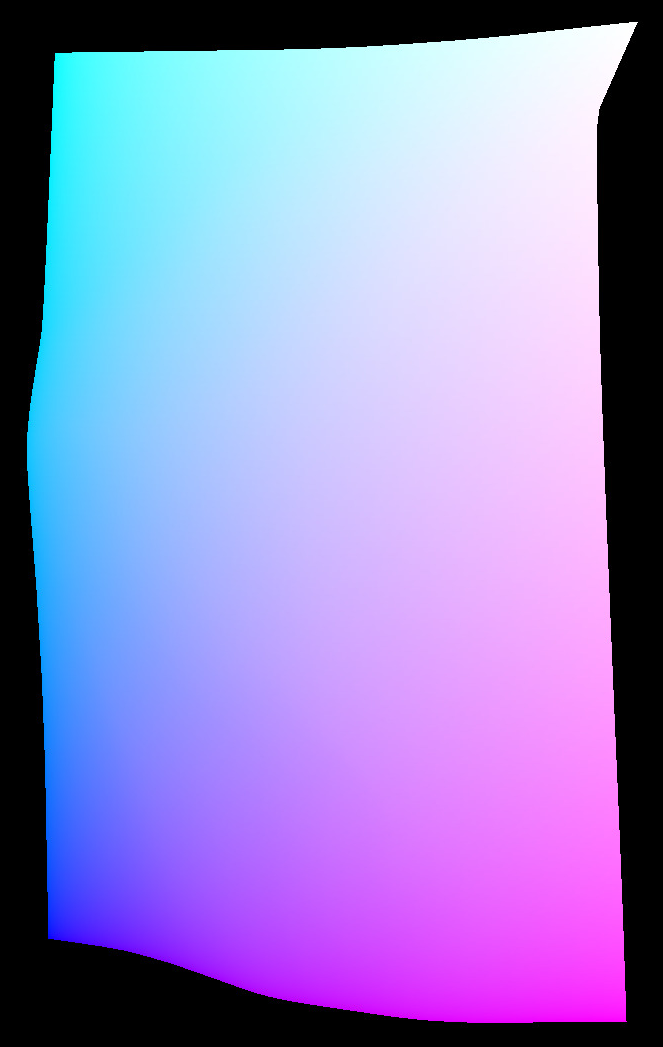}\\
    \scriptsize Rendered image &\scriptsize Albedo &\scriptsize Shading &\scriptsize Normal map &\scriptsize 3D coordinates &\scriptsize UV map
    \end{tabular}
    \caption{Example of a sample from our \ours{} dataset, with all its annotations. 
    }
    \label{fig:dataset_sample}
    \vspace{-5pt}
\end{figure}

The digitization of physical documents has become an essential task in bridging the gap between legacy paper formats and modern digital workflows. Because of their ease of use and flexibility, mobile devices are increasingly utilized for this purpose. However, in contrast to flatbed scanners, which capture images of perfectly flat documents under controlled lighting, photos of documents taken with mobile devices such as smartphones suffer from multiple visual degradations. Since the paper is rarely perfectly planar and the camera viewpoint can vary significantly, geometric distortions are almost always present. In addition, under unconstrained and non-uniform lighting, these 3D deformations cause unwanted shadows and variable shading. These artifacts reduce the visual quality of the captured document and make its automatic processing significantly more difficult.

To obtain a visually appealing and practically useful scan, the captured document needs to undergo geometric unwarping and illumination correction. Historically, document unwarping was tackled through model-based approaches, which attempt to estimate the document's 3D geometry using constrained representations and flatten it by solving an optimization problem. However, these methods are limited in the types of deformations they can handle. Deep learning-based approaches have become the standard way to solve both document unwarping and illumination correction. Yet, the performance of these techniques is fundamentally bounded by the quality and scale of their training datasets. Generating training data for these tasks is challenging, as networks require pixel-perfect annotations, such as mappings from the document photograph to the unwarped image or shading maps, which are nearly impossible to acquire for real-world photographs of documents. The most commonly used dataset, Doc3D~\cite{das.etal2019}, containing 100,000 samples, is synthetic, and relies on manually scanned 3D meshes, rendered at a relatively low image resolution ($448 \times 448$ pixels). An alternative dataset, UVDoc~\cite{verhoeven.etal2023} (20,000 samples), employs a hybrid approach, compositing document textures with real photographs of blank deformed paper, but it provides only coarse annotations for the unwarping mapping. Both datasets require extensive manual labor to create, making them inherently difficult to scale.

To overcome this scalability and quality bottleneck, we present \ours{}, the first large-scale dataset for document unwarping and illumination correction. \ours{} contains 1,000,000 training samples, 100,000 validation samples and more than 38,000 test samples. To achieve this unprecedented scale, we rely on rendering and use scalable processes at each step of the data generation pipeline. The deformed 3D document meshes are simulated using ArcSim~\cite{narain.etal2012, narain.etal2013}, a physical simulation engine for which we design multiple parameterized scenarios to ensure diverse outputs. The document textures and background materials are sourced from various repositories with permissive licenses. Finally, the samples are rendered using Blender~\cite{blender}, enabling the procedural generation of lighting environments, camera poses and paper material properties. Each sample is rendered at a high resolution ($1024 \times 1440$ pixels) and includes pixel-perfect annotations, such as albedo, shading, normal maps, UV maps and 3D coordinates. A sample from our dataset, along with its annotations, is presented in~\cref{fig:dataset_sample}. 

To demonstrate the practical benefits of our dataset, we train a lightweight baseline model to simultaneously perform document unwarping and illumination correction. We evalute the inference speed and performance of this model on the standard DocUNet benchmark~\cite{ma.etal2018}, comparing it to the state-of-the-art.

\section{Related works}
\label{sec:related_works}

\subsection{Document unwarping}

Early model-based approaches to document unwarping rely on a two-step process: estimating the 3D document surface via specialized hardware~\cite{brown.seales2001, brown.seales2004, zhang.etal2008, meng.etal2014}, multi-view imagery~\cite{ulges.etal2004, koo.etal2009, you.etal2018} or visual cues~\cite{tian.narasimhan2011, meng.etal2018}, followed by simulated or geometric flattening~\cite{brown.seales2001, brown.seales2004, zhang.etal2008, you.etal2018, koo.etal2009, tsoi.brown2007, liang.etal2008, meng.etal2014}.
However, data-driven approaches have largely replaced these methods due to their superior generalization capabilities. Ma et al.~\cite{ma.etal2018} introduce the first deep learning-based unwarping method, while DewarpNet~\cite{das.etal2019} proposes an architecture based on 3D coordinates, inspired by classical techniques. Following these foundational works, the field innovates rapidly. Several works explore patch-based~\cite{li.etal2019, das.etal2021} and iterative~\cite{zhang.etal2022} architectures, integrate text-line supervision~\cite{jiang.etal2022, feng.etal2022, li.etal2023} and employ transformers~\cite{feng.etal2021, zhou.etal2025} to capture global context. Other works focus on predicting coarse mappings~\cite{xie.etal2021, xue.etal2022, verhoeven.etal2023}, adopting multi-task frameworks to handle various document processing steps~\cite{zhang.etal2024, tang.etal2024, zhao.etal2025} and, more recently, leveraging diffusion models~\cite{zhang.etal2025, kumari.das2025}. While these deep learning methods show strong performance, their effectiveness strongly depends on the scale and quality of the training datasets, which motives us to develop our \ours{} dataset.

\begin{table}[b]
	\setlength{\tabcolsep}{5pt}
    \centering
    \caption{Comparison of the various document unwarping datasets. \textit{3D geometry source} characterizes the type of geometry used for rendering in a synthetic dataset, \textit{Mapping} represents the kind of mapping annotations and \textit{3D} signifies the inclusion of 3D annotations.
    }
    \resizebox{\linewidth}{!}{%
    \begin{tabular}{l r r @{ $\times$ } l l l l r}
    \toprule
        Dataset & \# Samples & \multicolumn{2}{c}{Resolution} & Type & 3D geometry source & Mapping & 3D\\ \midrule
        DocUNet \cite{ma.etal2018} & 100,000 & 600 & 800 & Synthetic 2D & N/A & Dense & \xmark\\
        Doc3D \cite{das.etal2019} & 100,000 & 448 & 448 & Synthetic & 3D scanned & Dense & \cmark\\
        DocProj \cite{li.etal2019} & 2,450 & 1800 & 2400 & Synthetic & Geometric deformations & Dense & \xmark \\
        DIW \cite{ma.etal2022} & 5,000 & 512 & 512 & Real & N/A & --- & \xmark\\
        Inv3D \cite{hertlein.etal2023}  & 25,000 & 1600 & 1600 & Synthetic & 3D scanned & Dense & \cmark \\
        Book3D \cite{liu.etal2026} & 56,000 & 1200 & 800 & Synthetic & Geometric deformations & Dense & \cmark\\
        UVDoc \cite{verhoeven.etal2023} & 20,000 & 488 & 712 & Pseudo-real & Depth camera & Coarse & \cmark\\
        \ours{} & 1,000,000 & 1024 & 1440 & Synthetic & Physics simulation & Dense & \cmark\\ \bottomrule
    \end{tabular}
    }
    \label{tab:unwarping_dataset_comparison}
\end{table}

\begin{figure}[t]
    \centering
	\small
	\setlength{\tabcolsep}{1pt}
	\renewcommand{\arraystretch}{0.0}  
    \setlength{\fboxsep}{0pt}
    \newcommand{\figheight}{0.18\linewidth}
    \begin{tabular}{cccccc}
      \includegraphics[height=\figheight]{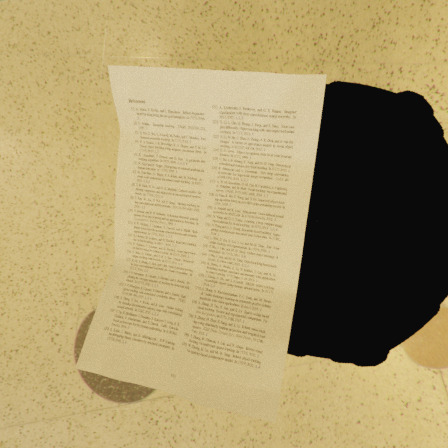}
    & \includegraphics[height=\figheight]{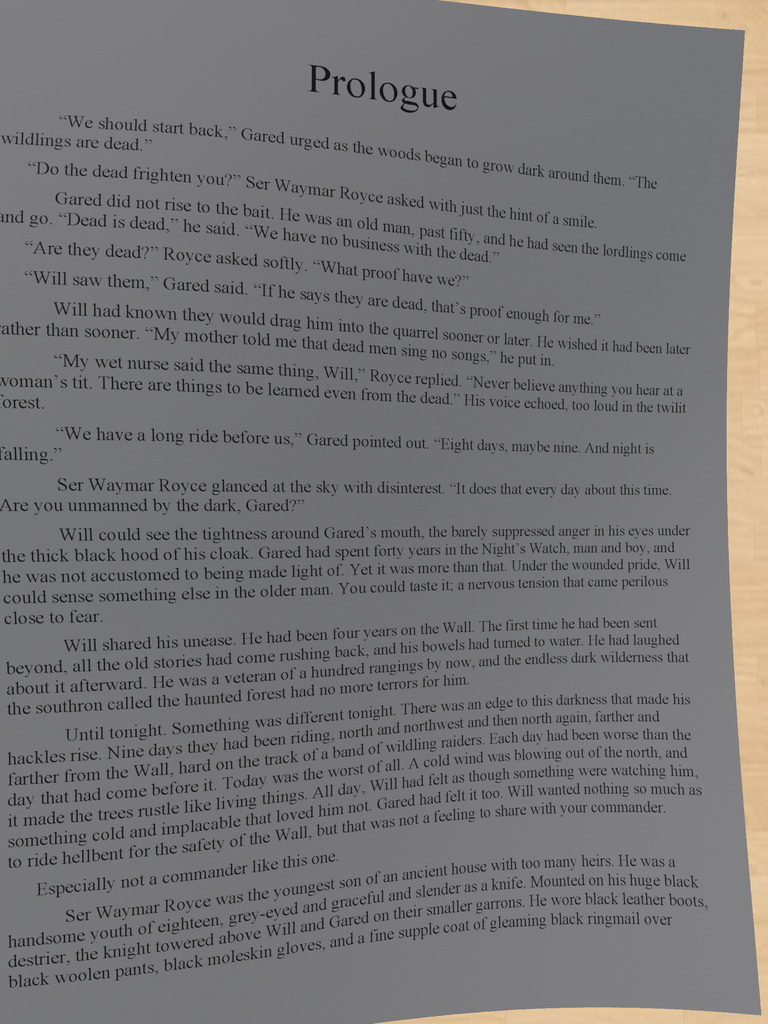}
    & \includegraphics[height=\figheight]{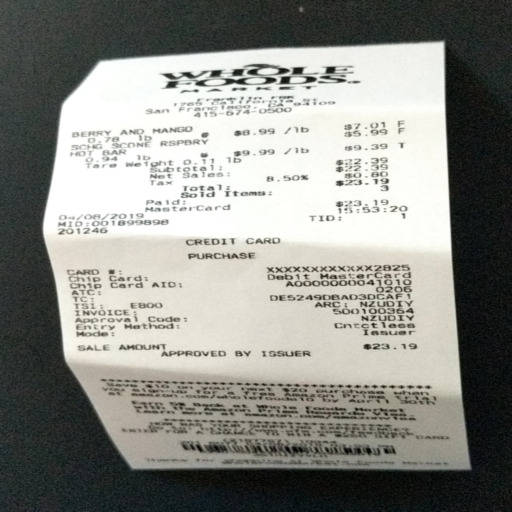}
    & \includegraphics[height=\figheight]{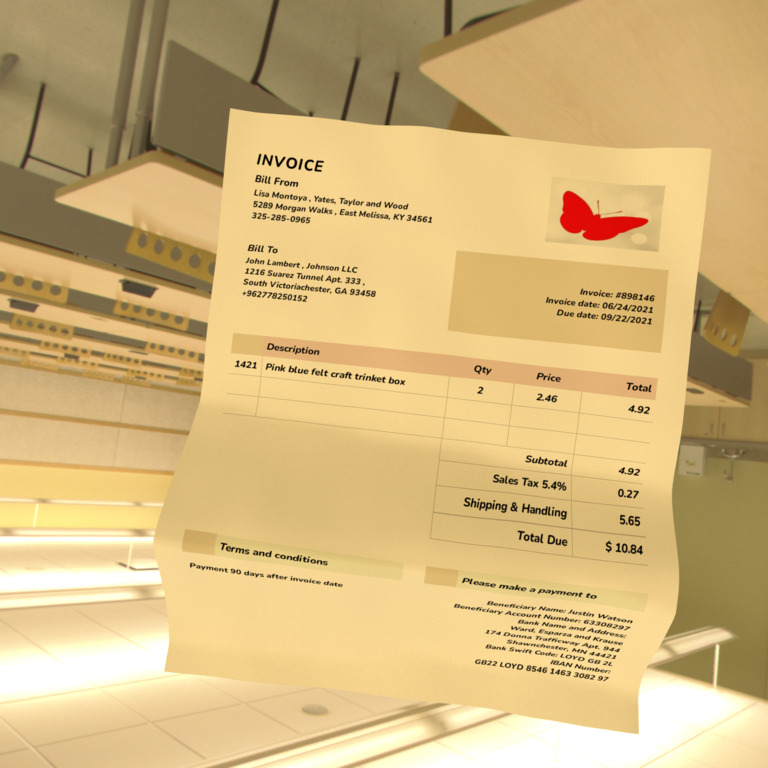}
    & \includegraphics[height=\figheight]{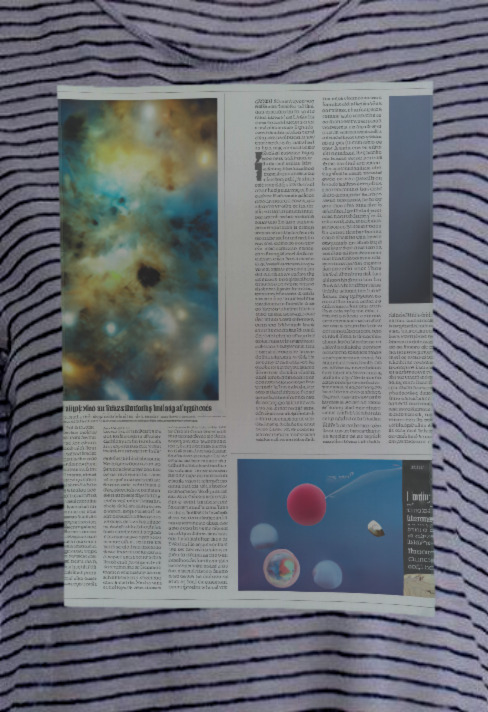}
    & \includegraphics[height=\figheight]{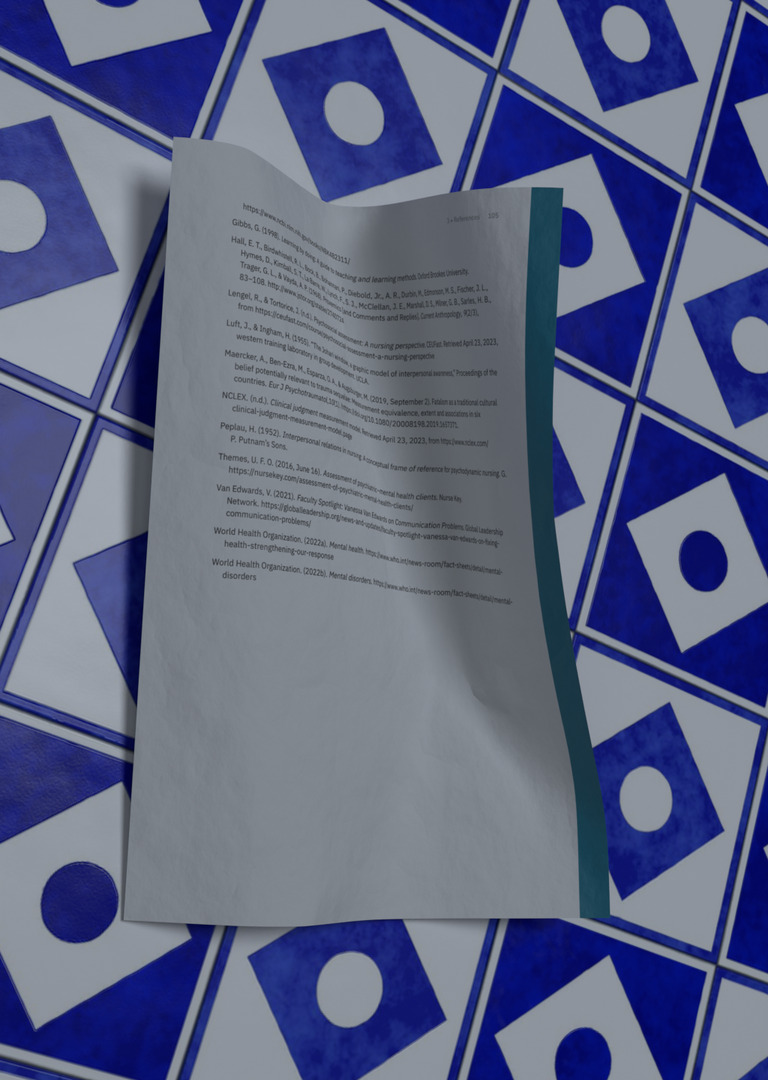}\\[2pt]
      \includegraphics[height=\figheight]{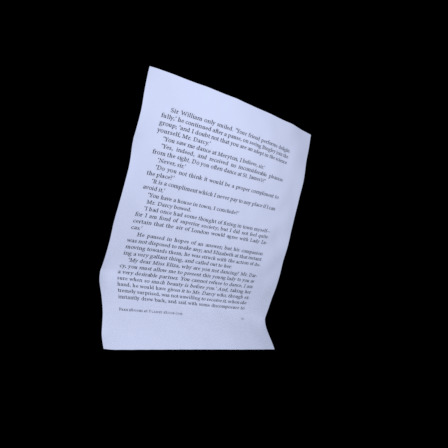}
    & \includegraphics[height=\figheight]{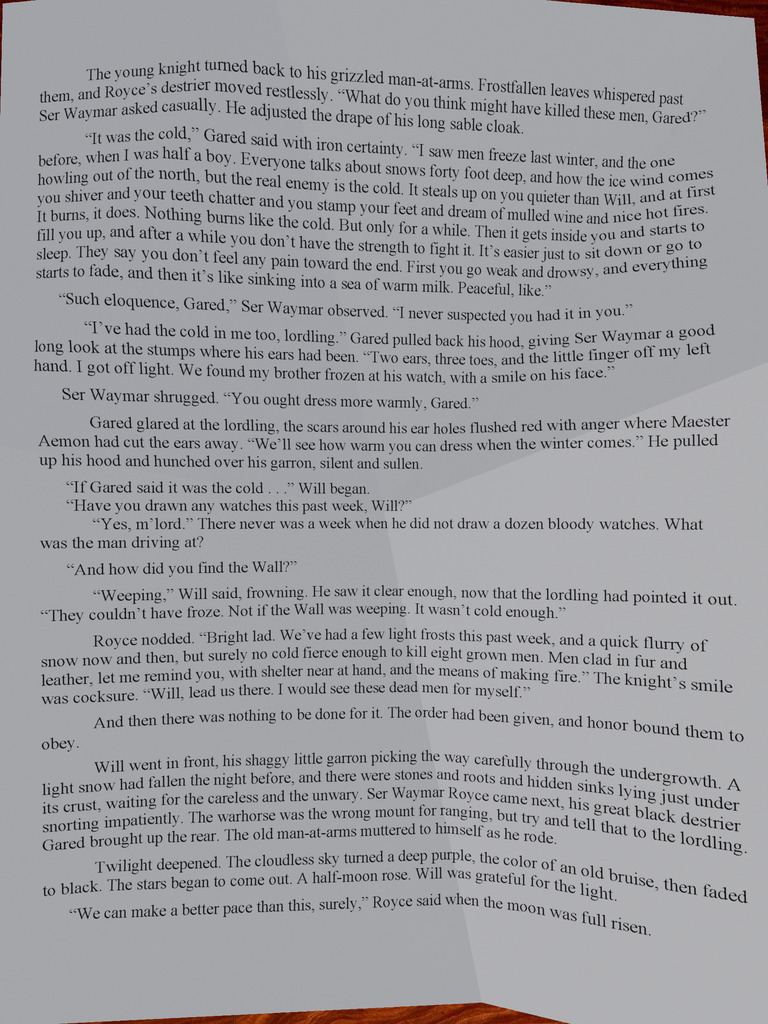}
    & \includegraphics[height=\figheight]{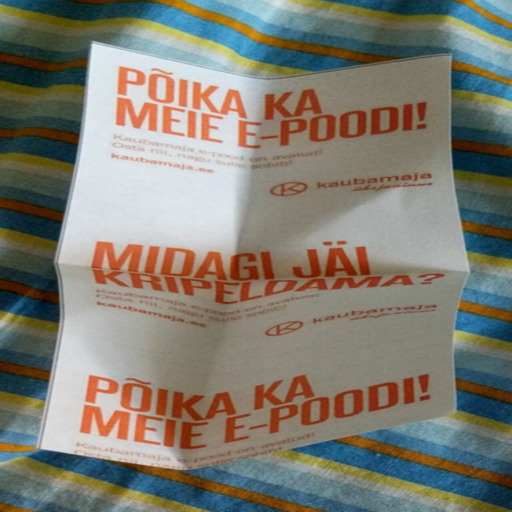}
    & \includegraphics[height=\figheight]{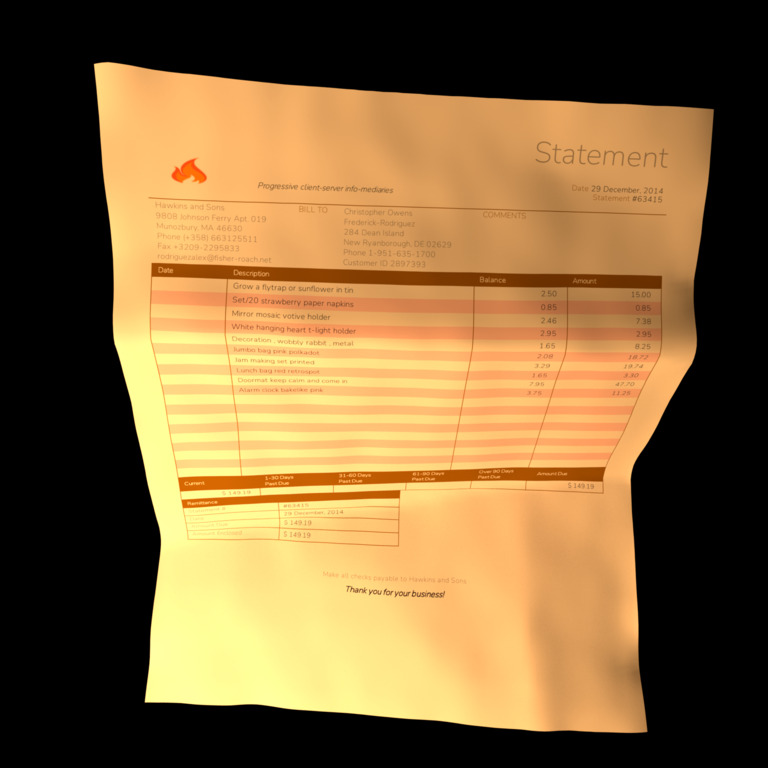}
    & \includegraphics[height=\figheight]{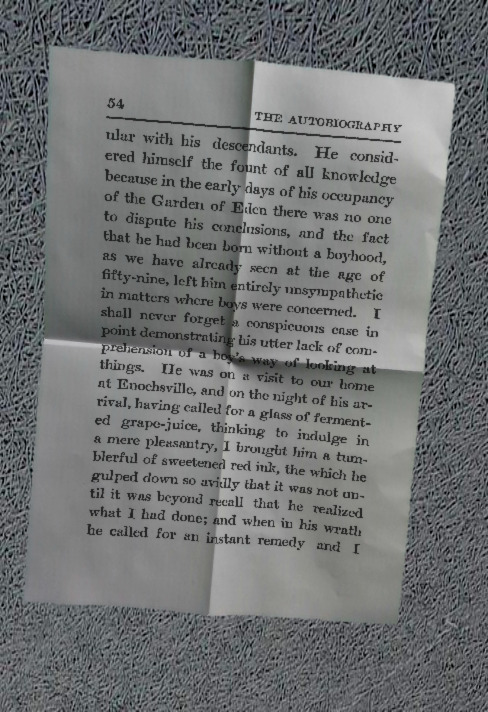}
    & \includegraphics[height=\figheight]{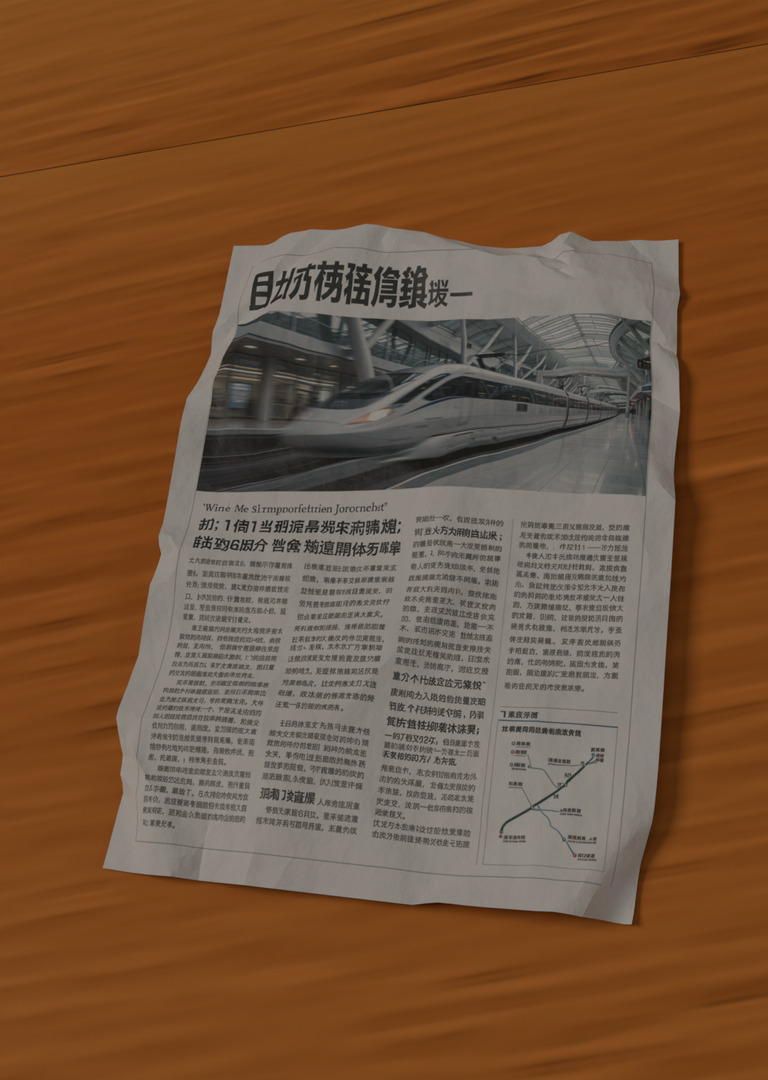}\\[2pt]
      \includegraphics[height=\figheight]{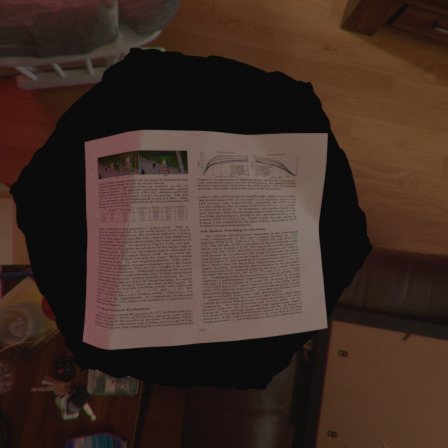}
    & \includegraphics[height=\figheight]{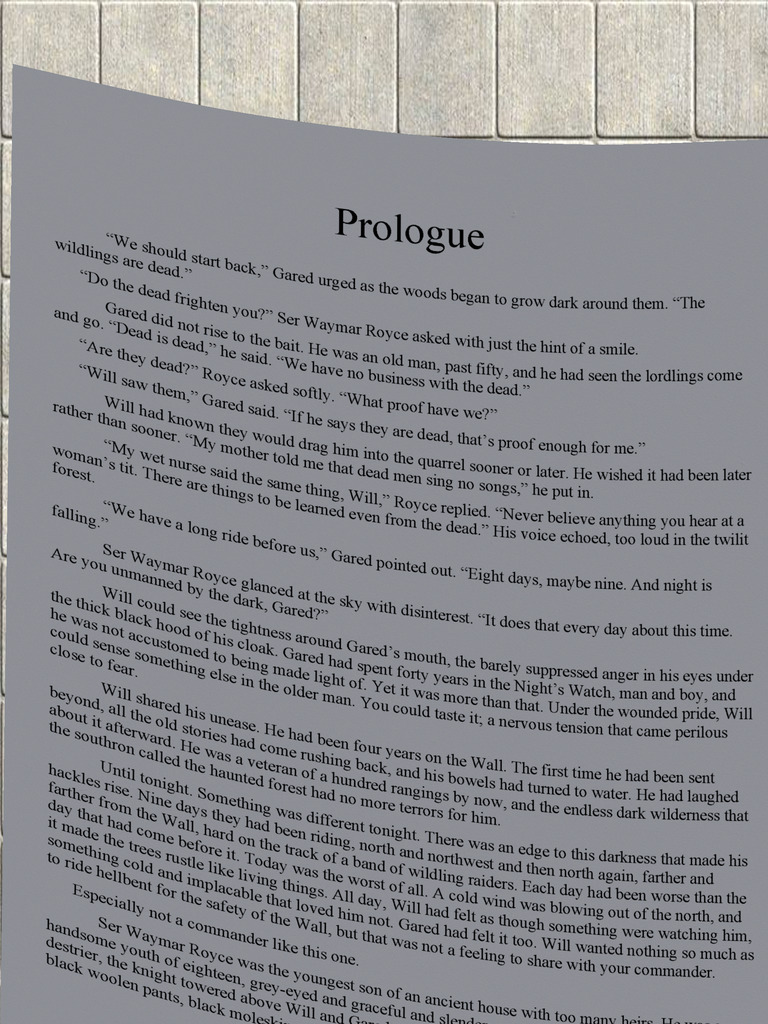}
    & \includegraphics[height=\figheight]{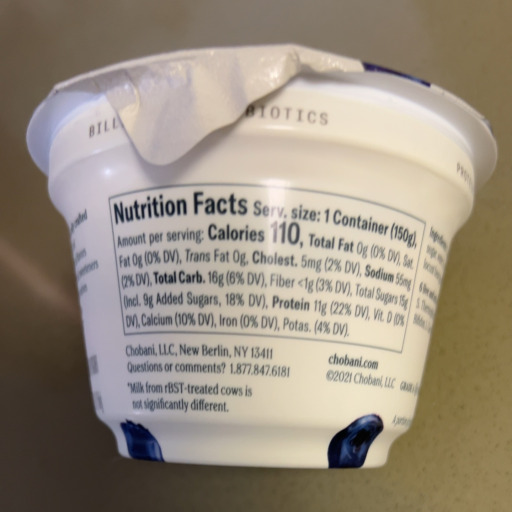}
    & \includegraphics[height=\figheight]{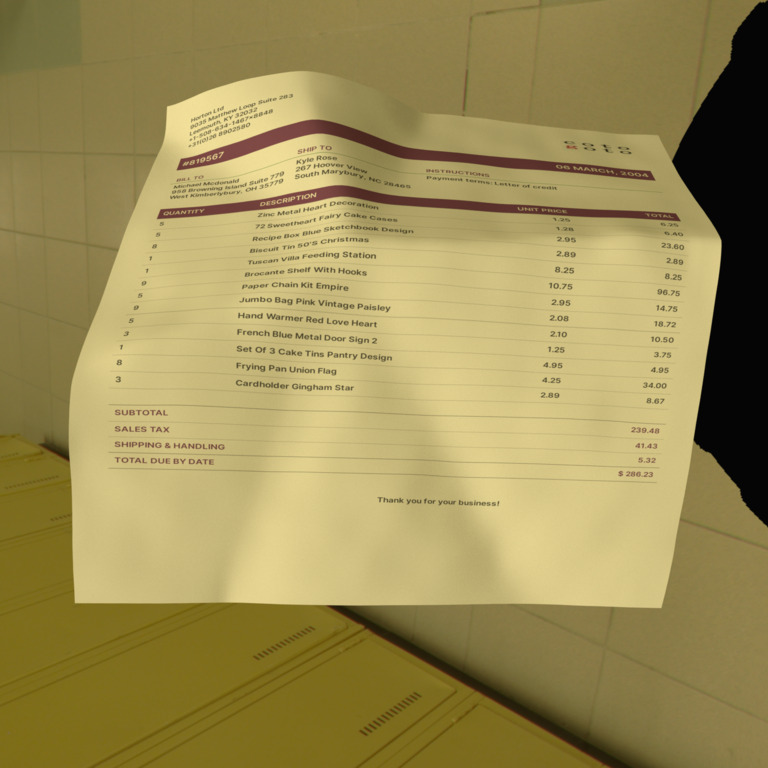}
    & \includegraphics[height=\figheight]{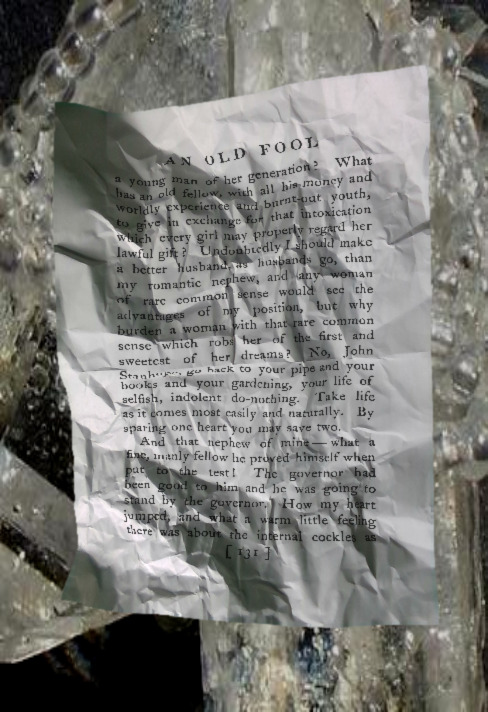}
    & \includegraphics[height=\figheight]{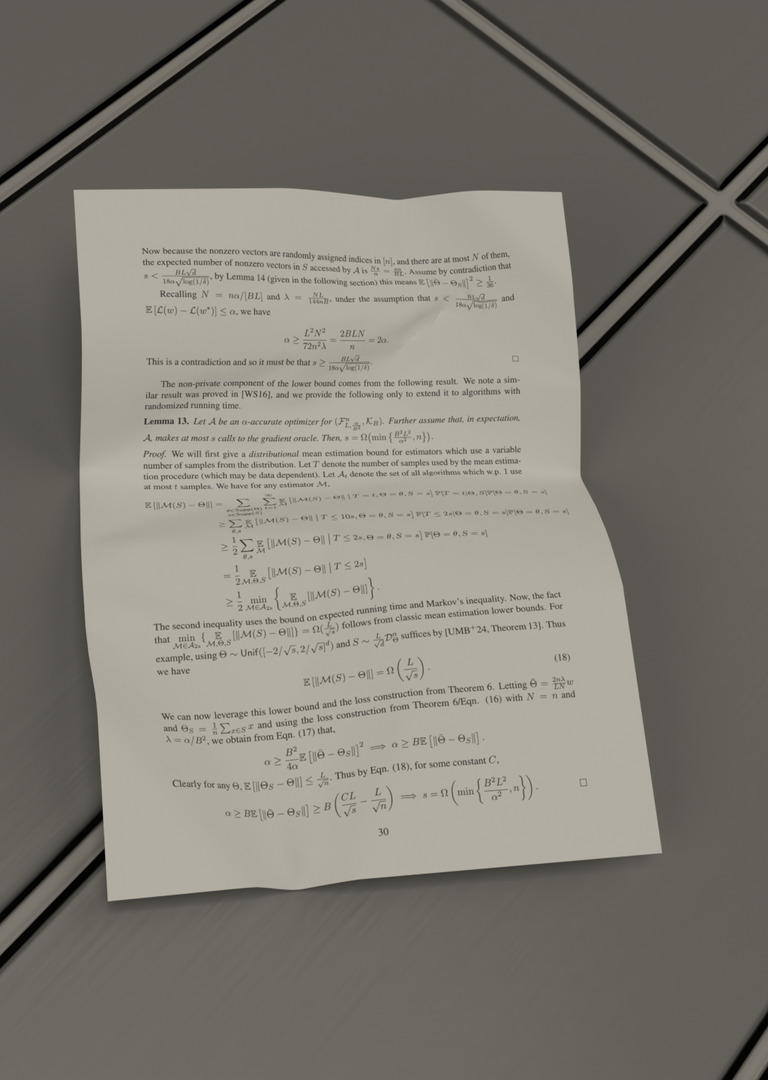}\\[4pt]
      \scriptsize Doc3D \cite{das.etal2019} 
    & \scriptsize DocProj \cite{li.etal2019} 
    & \scriptsize DIW \cite{ma.etal2022}  
    & \scriptsize Inv3D \cite{hertlein.etal2023} 
    & \scriptsize UVDoc \cite{verhoeven.etal2023} 
    & \scriptsize Ours 
    \end{tabular}
    \caption{Examples of data points from various document unwarping datasets.}
    \label{fig:unwarping_datasets_samples}
\end{figure}

\paragraph{Datasets.} Since modern models typically predict a dense backward mapping from the warped input to the flat document, their training datasets require complex, pixel-perfect annotations that are difficult to obtain for real-world photographs. Earlier methods rely on a synthetic dataset generated via non-physically plausible 2D deformations~\cite{ma.etal2018}, later augmented with supplementary annotations~\cite{xie.etal2020, xie.etal2021}. For years, Doc3D~\cite{das.etal2019} has served as the standard dataset for document unwarping. It has been extended with additional ground-truth labels, such as text lines~\cite{markovitz.etal2020, feng.etal2022} or cropped images~\cite{feng.etal2024}. While rendered in Blender, similar to our \ours{}, Doc3D contains far fewer samples at a significantly lower resolution. Moreover, its reliance on manual 3D scanning is not scalable. The captured 3D scans required geometric post-processing that resulted in over-smoothing, removing fine-grained details and in some cases making the surfaces non-developable. DocProj~\cite{li.etal2019} uses an approach similar to ours but features simplistic geometries obtained by geometrically deforming a mesh and a very limited size. UVDoc~\cite{verhoeven.etal2023} prioritizes visual and geometric realism, being the only non-rendered dataset providing mapping annotations. Inv3D~\cite{hertlein.etal2023} and Book3D~\cite{liu.etal2026} focus on certain document types and remain severely limited in scale and/or deformation variety. Finally, datasets of real-world documents~\cite{ma.etal2022} only provide the flatbed scan as ground truth, making them impractical for training mapping-based networks. As summarized in~\cref{tab:unwarping_dataset_comparison}, \ours{} stands out thanks to its massive scale and high image quality. It also contains the largest variety of document types and deformations and, as illustrated in~\cref{fig:unwarping_datasets_samples}, the rendering of \ours{} is more realistic than previous datasets, with shadows correctly cast on the background.

\subsection{Illumination correction}

Beyond geometric unwarping, document restoration encompasses photometric corrections to convert camera-captured document images into clean, flatbed-style scans. In this context, deshadowing and illumination correction are two distinct but frequently confounded tasks. Deshadowing removes shadows cast on the document by external objects~\cite{lin.etal2020, li.etal2023b}, while illumination correction rectifies shading. Since \ours{} focuses on the latter, we review the relevant illumination correction literature here.

Classical methods rely on estimating a shading map by interpolating background border colors~\cite{brown.etal2006} or inpainting document content~\cite{zhang.etal2007}. Modern data-driven approaches largely outperform these techniques. Initial deep learning methods address the problem sequentially, correcting colorimetry only after rectifying the document geometry~\cite{das.etal2019, li.etal2019, feng.etal2021}. Subsequent works focus specifically on dedicated illumination networks~\cite{das.etal2020}, utilizing coarse-to-fine strategies~\cite{zhang.etal2024b} as well as generative adversarial networks (GANs) combined with either cycle consistency~\cite{wang.etal2022} or vision-language priors~\cite{quan.etal2024} like CLIP~\cite{radford.etal2021}. More recently, multi-task approaches simultaneously solving unwarping and illumination correction have emerged~\cite{tang.etal2024, wang.etal2024}. The latest unified document enhancement models tackle all forms of document degradation concurrently, such as deshadowing, deblurring and illumination correction, by leveraging advanced feature generators~\cite{zhang.etal2024} or diffusion modules~\cite{zhao.etal2025}.

\begin{table}[t]
	\setlength{\tabcolsep}{5pt}
    \centering
    \caption{Comparison of the various illumination correction datasets. For RealDAE, the resolution ranges from $398 \times 164$ to $5344 \times 5312$ pixels.}
    \resizebox{\linewidth}{!}{%
    \begin{tabular}{l r c l l l c c}
    \toprule
        Dataset & \# Samples & Resolution & Type & 3D geometry source & Albedo & Shading\\ \midrule
        Doc3D \cite{das.etal2019} & 100,000 & $448 \times 448$ & Synthetic & 3D scanned & \cmark & \xmark\\
        DocProj \cite{li.etal2019} & 2,450 & $1800 \times 2400$ & Synthetic & Geometric deformations  & \cmark & \xmark\\
        Doc3DShade \cite{das.etal2020} & 90,000 & $640 \times 480$ & Synthetic & 3D scanned & \cmark & \xmark \\
        RealDAE \cite{zhang.etal2024b} & 450 & Various & Real & N/A & \cmark & \xmark\\
        \ours{} & 1,000,000 & $1024 \times 1440$ & Synthetic & Physical simulation & \cmark & \cmark\\ \bottomrule
    \end{tabular}
    }
    \label{tab:illumination_dataset_comparison}
\end{table}

\paragraph{Datasets.} Since illumination correction has received considerably less attention than geometric unwarping, dedicated training datasets are rare. Consequently, many deep learning-based illumination correction methods rely on the warped albedo or original document textures provided by unwarping datasets such as Doc3D~\cite{das.etal2019} and DocProj~\cite{li.etal2019}. While a few datasets target this specific task, they exhibit significant limitations. Doc3DShade~\cite{das.etal2020} follows a capture procedure similar to Doc3D~\cite{das.etal2019} but additionally captures real-world shading. However, it inherits Doc3D's scalability issues and contains very few distinct document textures and geometries, with the different samples varying mostly in the applied shading. RealDAE~\cite{zhang.etal2024b} is the only real-world dataset for this task. Its ground truth images were obtained by manually correcting the deformed images using Adobe Photoshop, making the collection process difficult and restricting its size to just 450 training samples. Unlike \ours{}, none of these datasets provide explicit shading annotations, offering only the albedo (see \cref{tab:illumination_dataset_comparison}). In addition, \ours{} is created with over 500,000 distinct geometries and an equal number of unique document images, all rendered using a high-quality path tracer and diverse lighting configurations for enhanced realism.

\section{\ours{} dataset generation pipeline}
\label{sec:dataset}

The generation of the \ours{} dataset consists of two main steps. First, the 3D geometries representing the deformed sheets of paper are simulated using ArcSim~\cite{narain.etal2012, narain.etal2013}, a simulation engine specialized for sheets of deformable materials. Then, the resulting meshes are rendered using Blender~\cite{blender} to produce the final photorealistic images along with their corresponding ground-truth annotations.

\subsection{Warped paper meshes generation}

\begin{figure}[t]
    \centering
	\small
	\setlength{\tabcolsep}{0.5pt}
    \setlength{\fboxsep}{0pt}
    \begin{tabular}{cccccc}
    \includegraphics[width=0.163\linewidth]{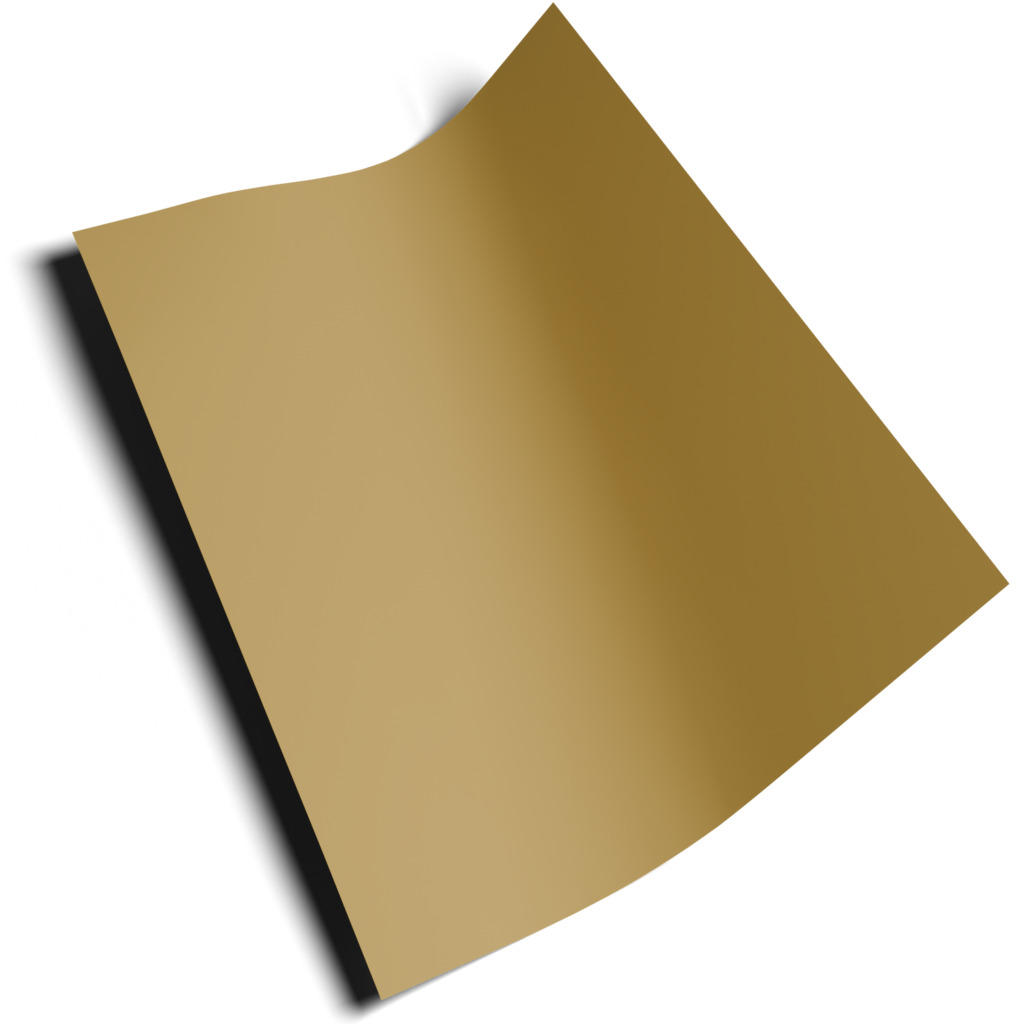}
    & \includegraphics[width=0.163\linewidth]{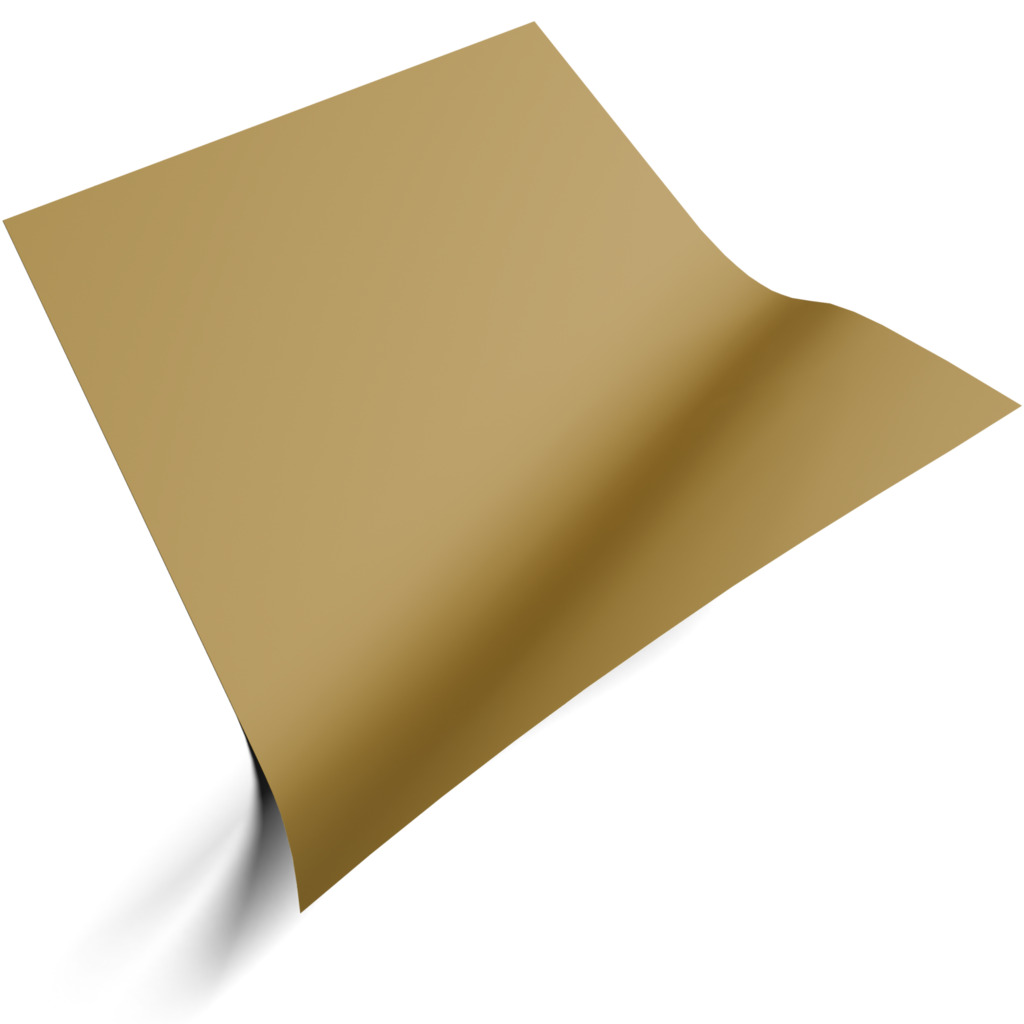}
    & \includegraphics[width=0.163\linewidth]{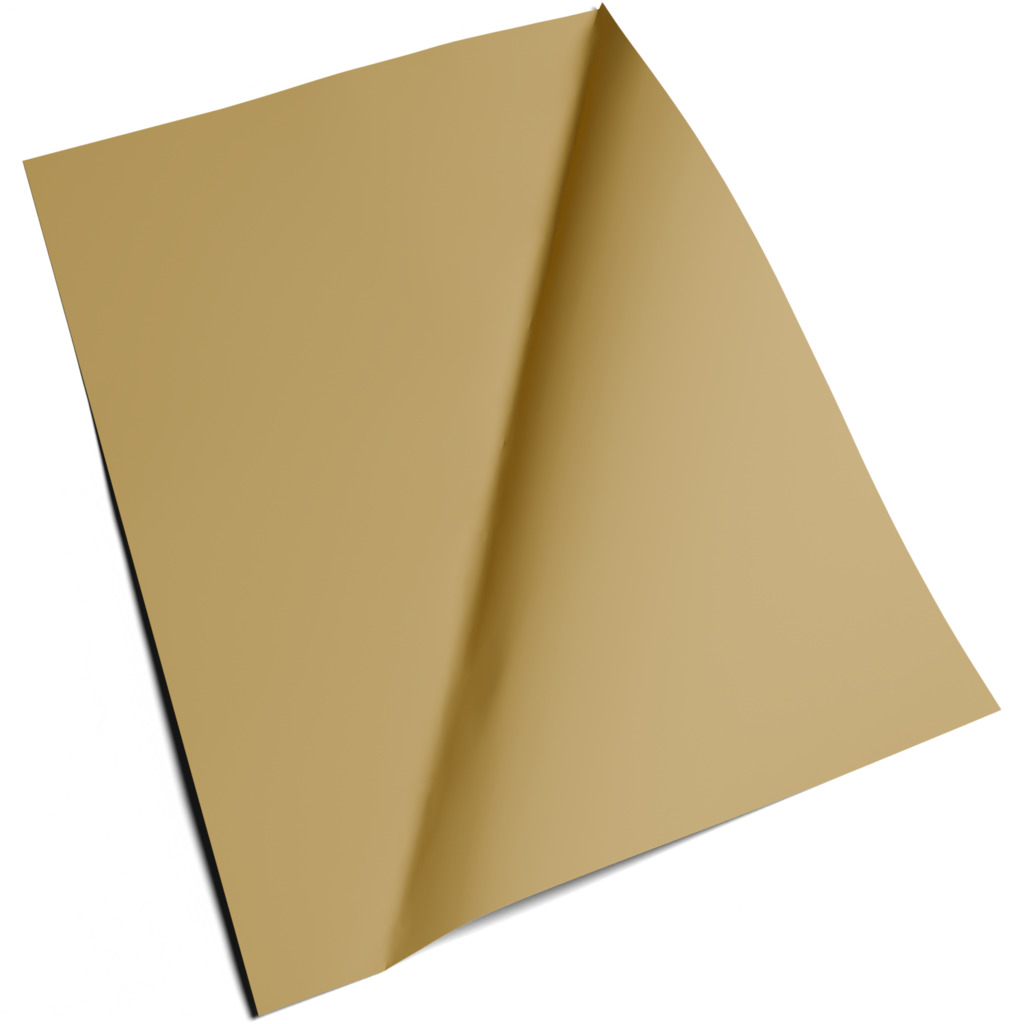}
    & \includegraphics[width=0.163\linewidth]{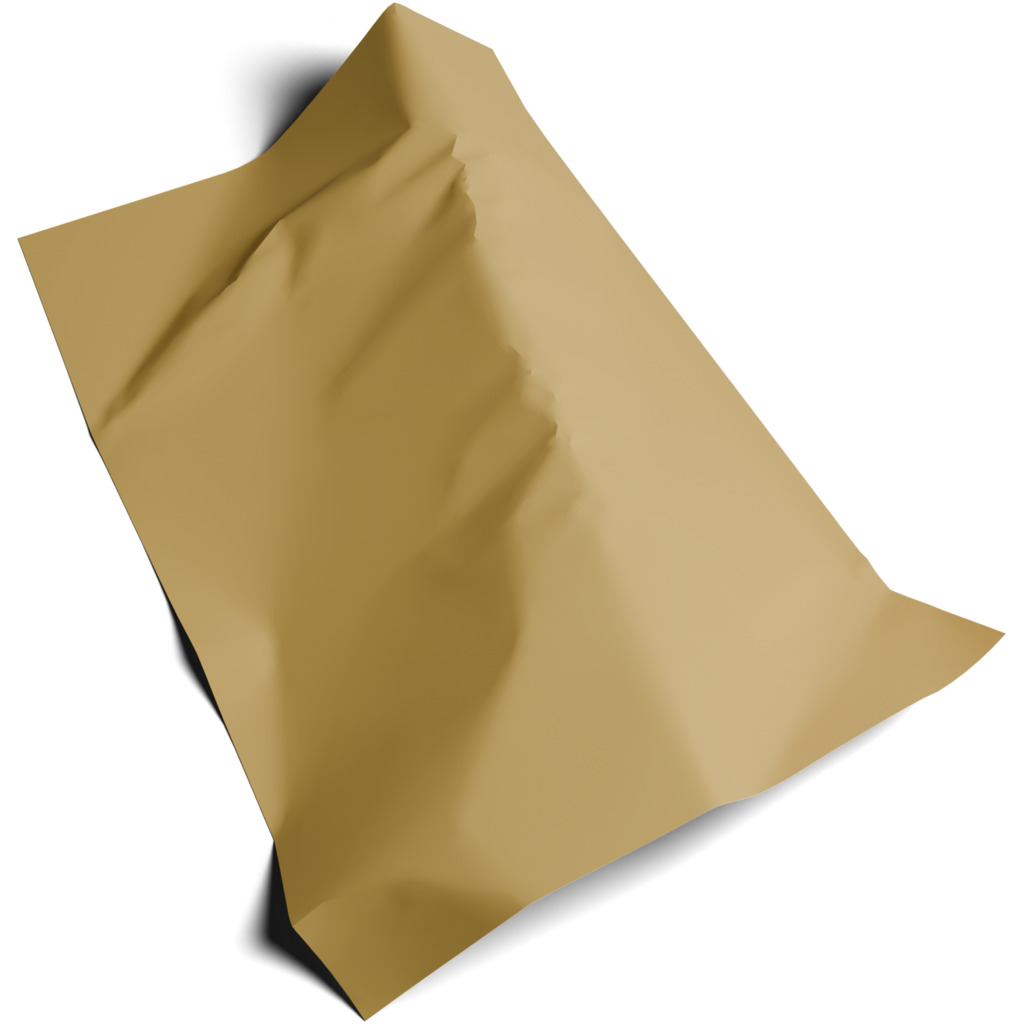}
    & \includegraphics[width=0.163\linewidth]{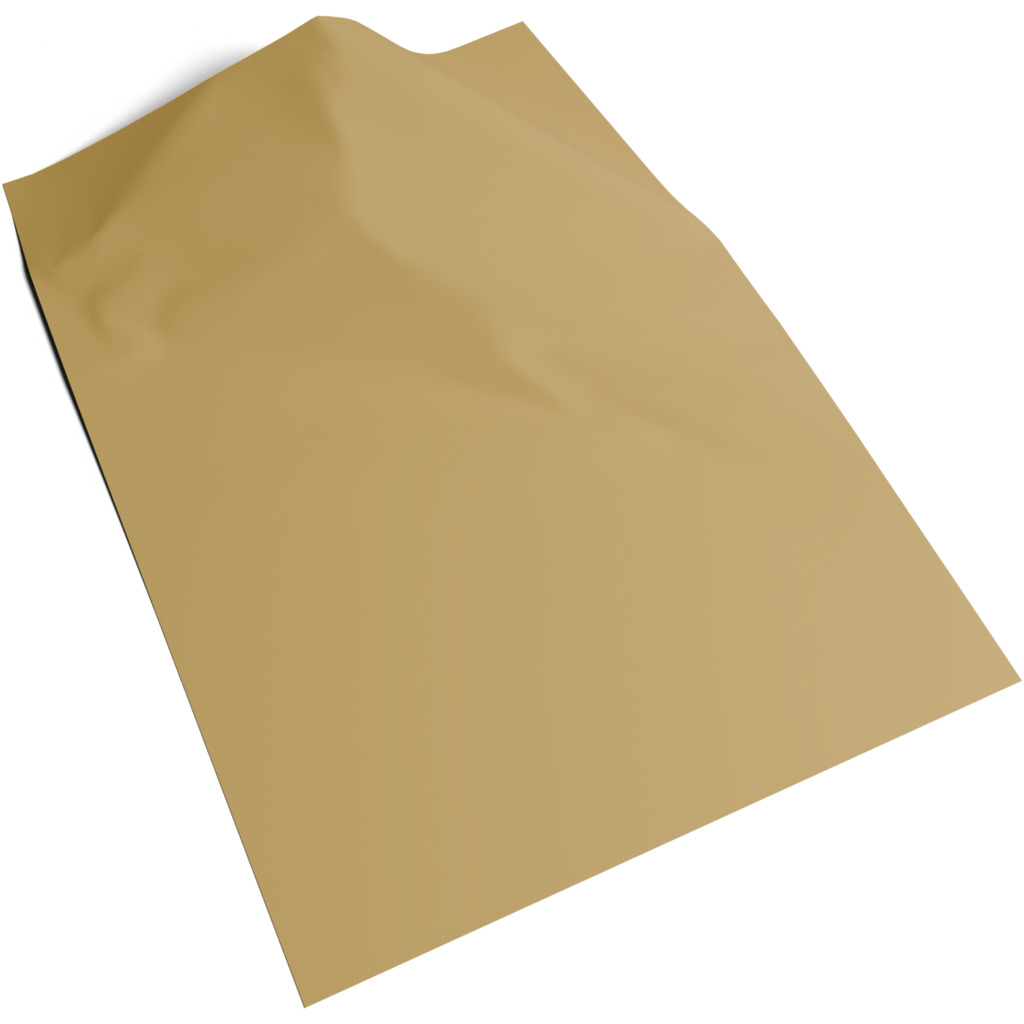}
    & \includegraphics[width=0.163\linewidth]{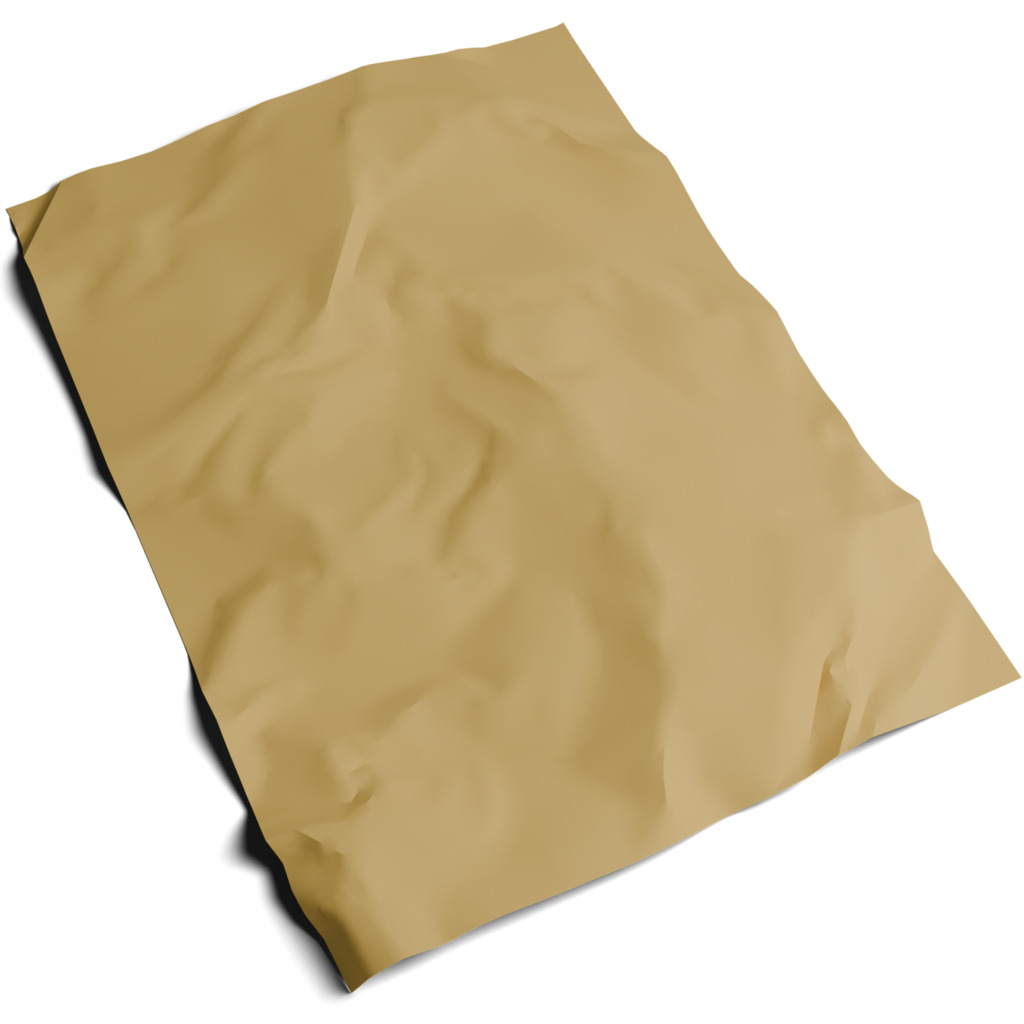}\\
    \includegraphics[width=0.163\linewidth]{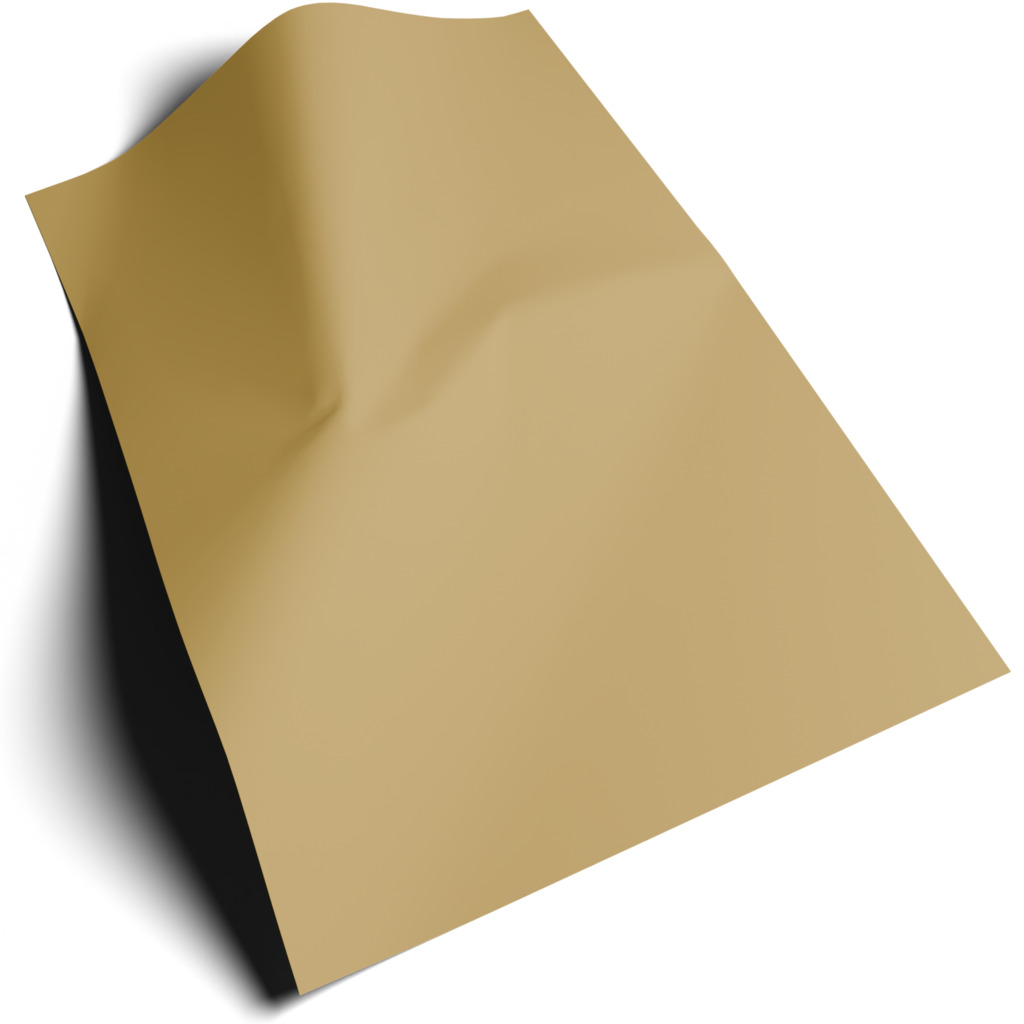}
    & \includegraphics[width=0.163\linewidth]{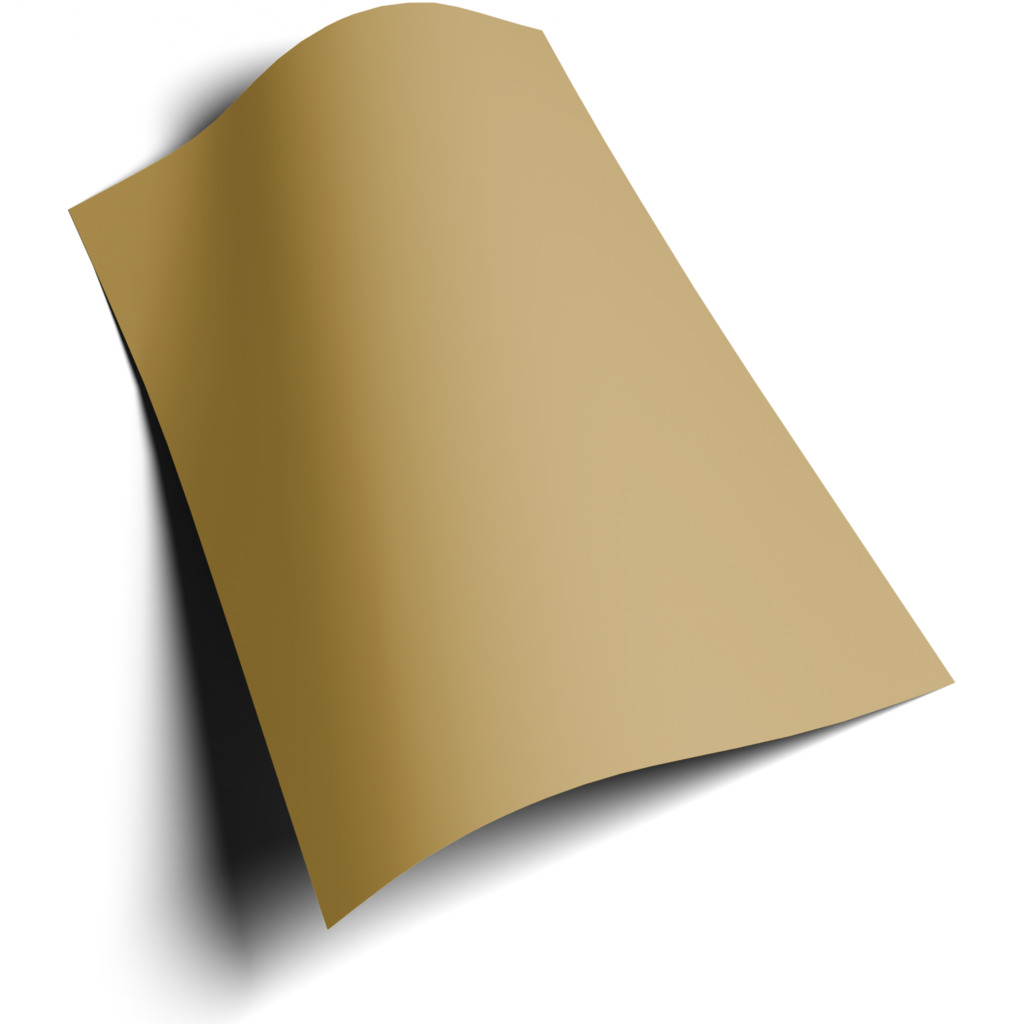}
    & \includegraphics[width=0.163\linewidth]{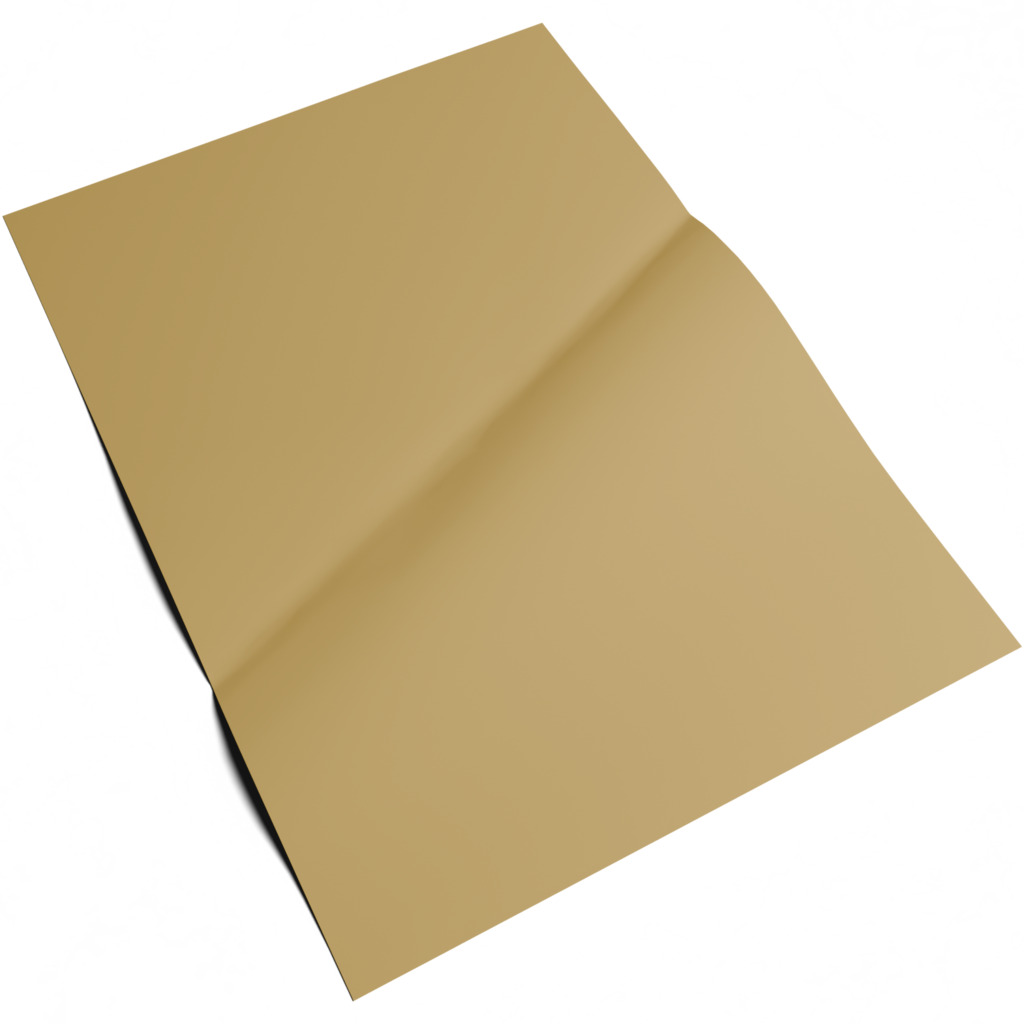}
    & \includegraphics[width=0.163\linewidth]{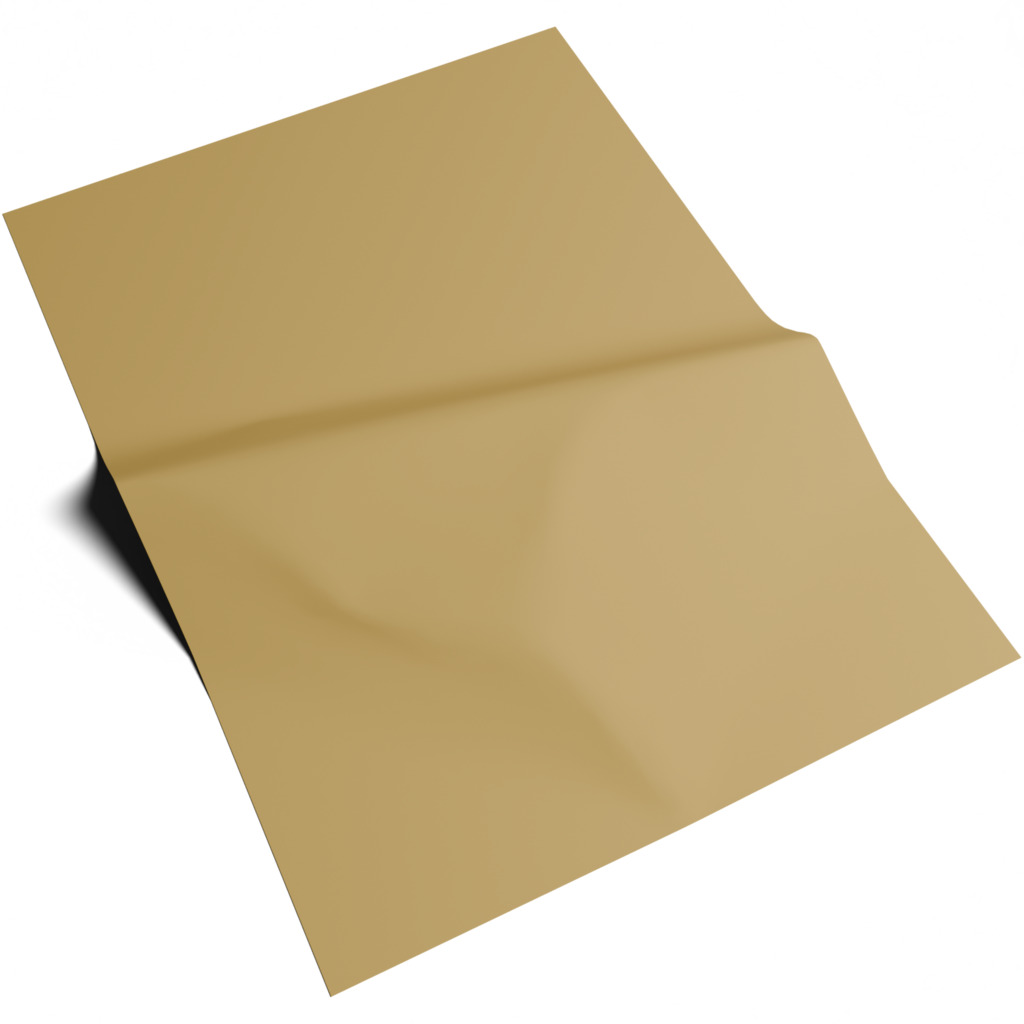}
    & \includegraphics[width=0.163\linewidth]{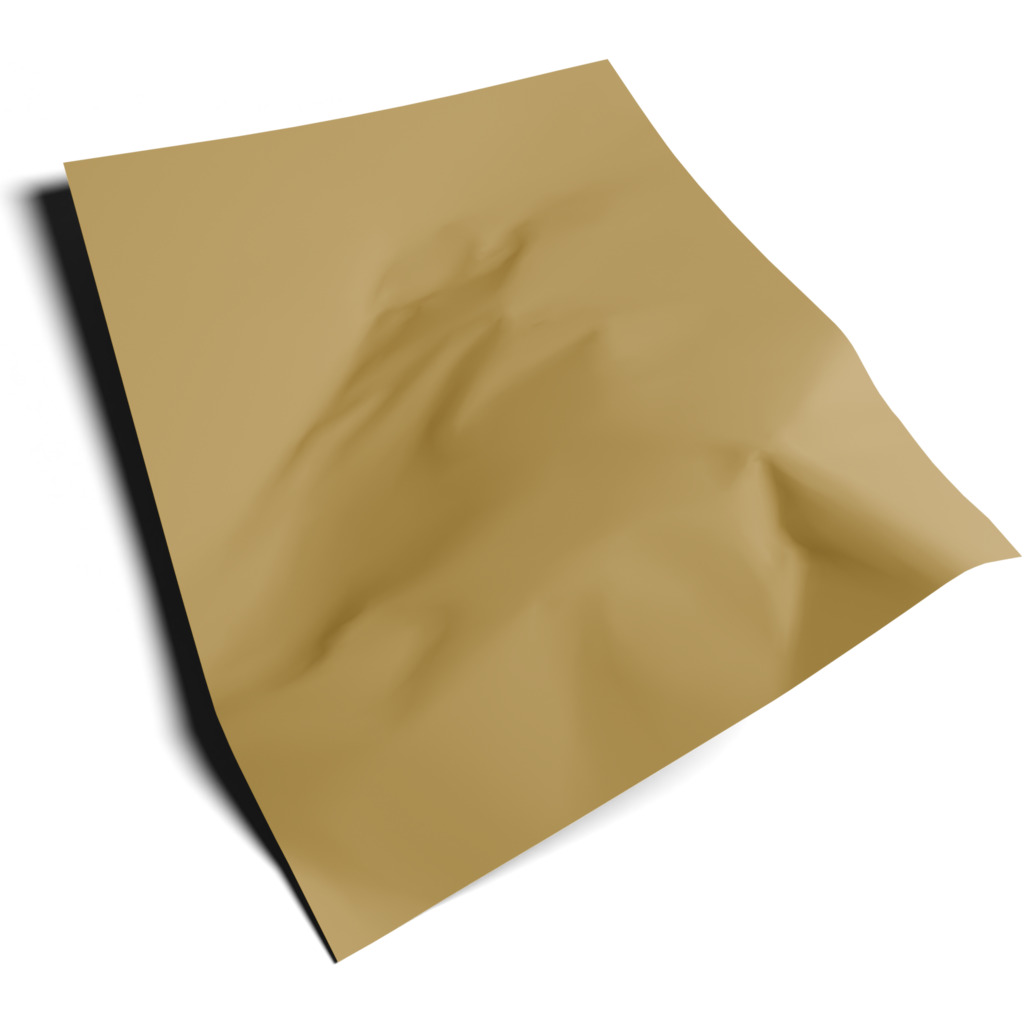}
    & \includegraphics[width=0.163\linewidth]{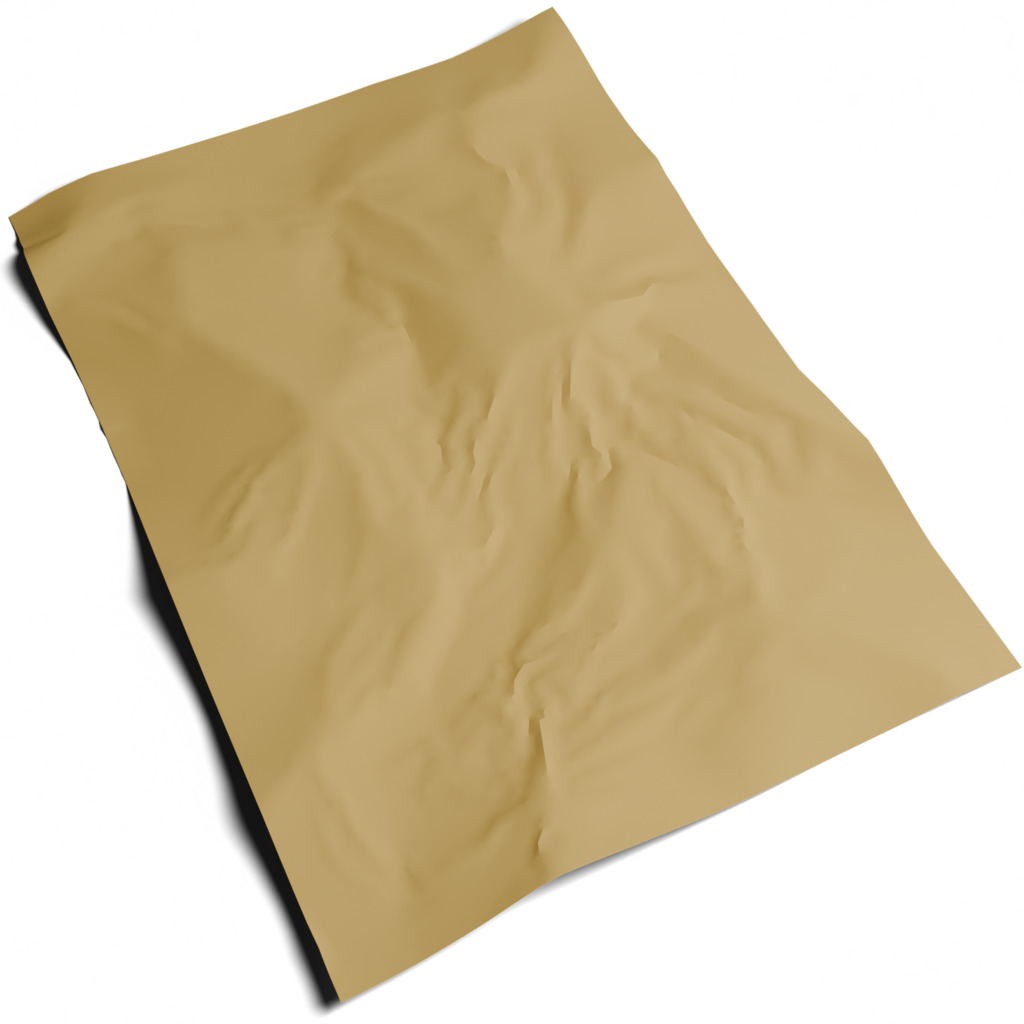}\\
    \includegraphics[width=0.163\linewidth]{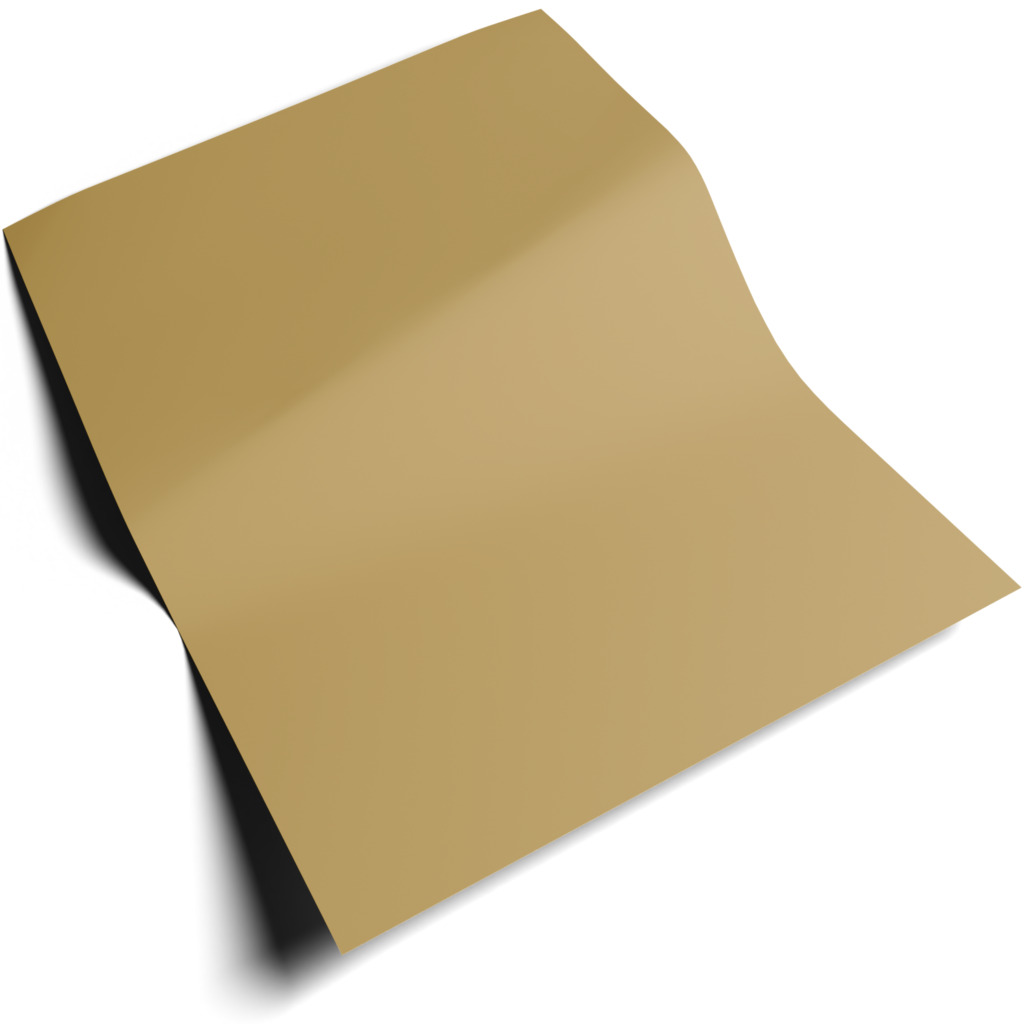}
    & \includegraphics[width=0.163\linewidth]{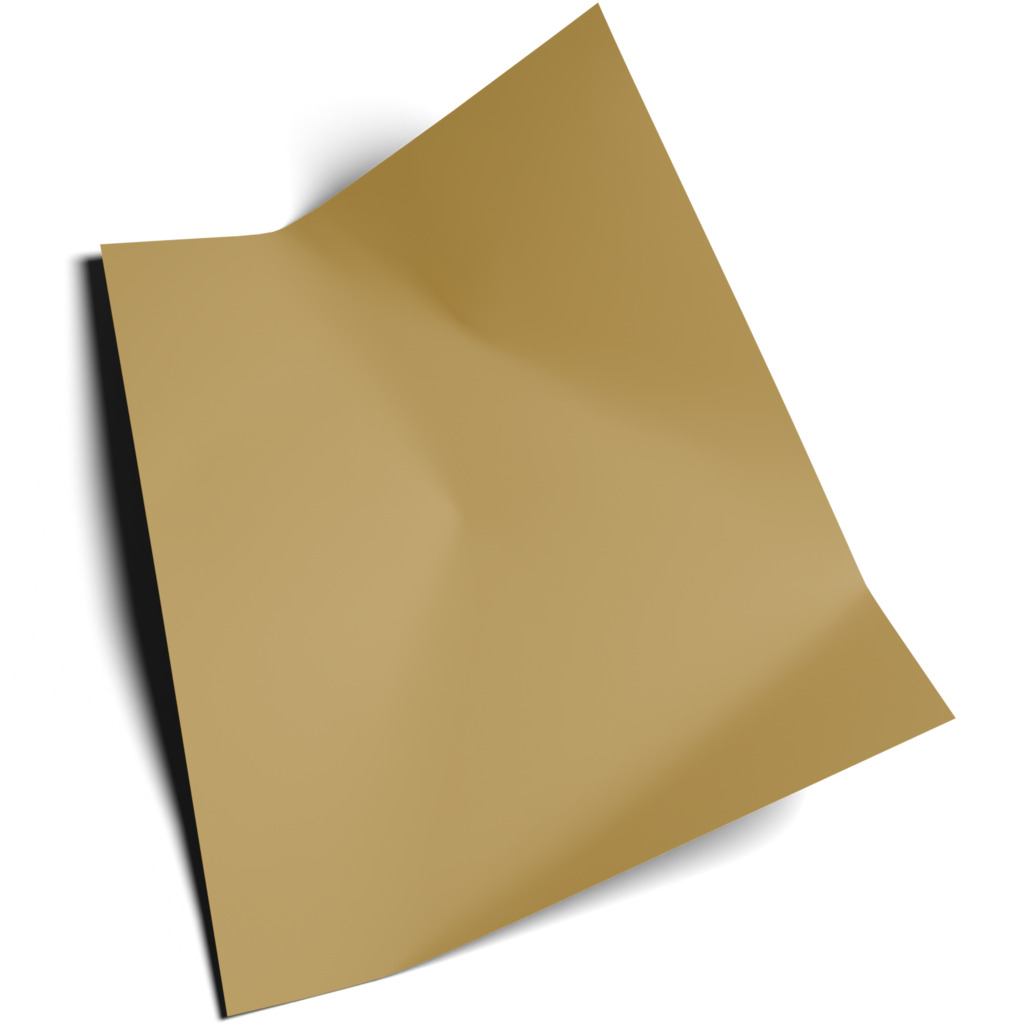}
    & \includegraphics[width=0.163\linewidth]{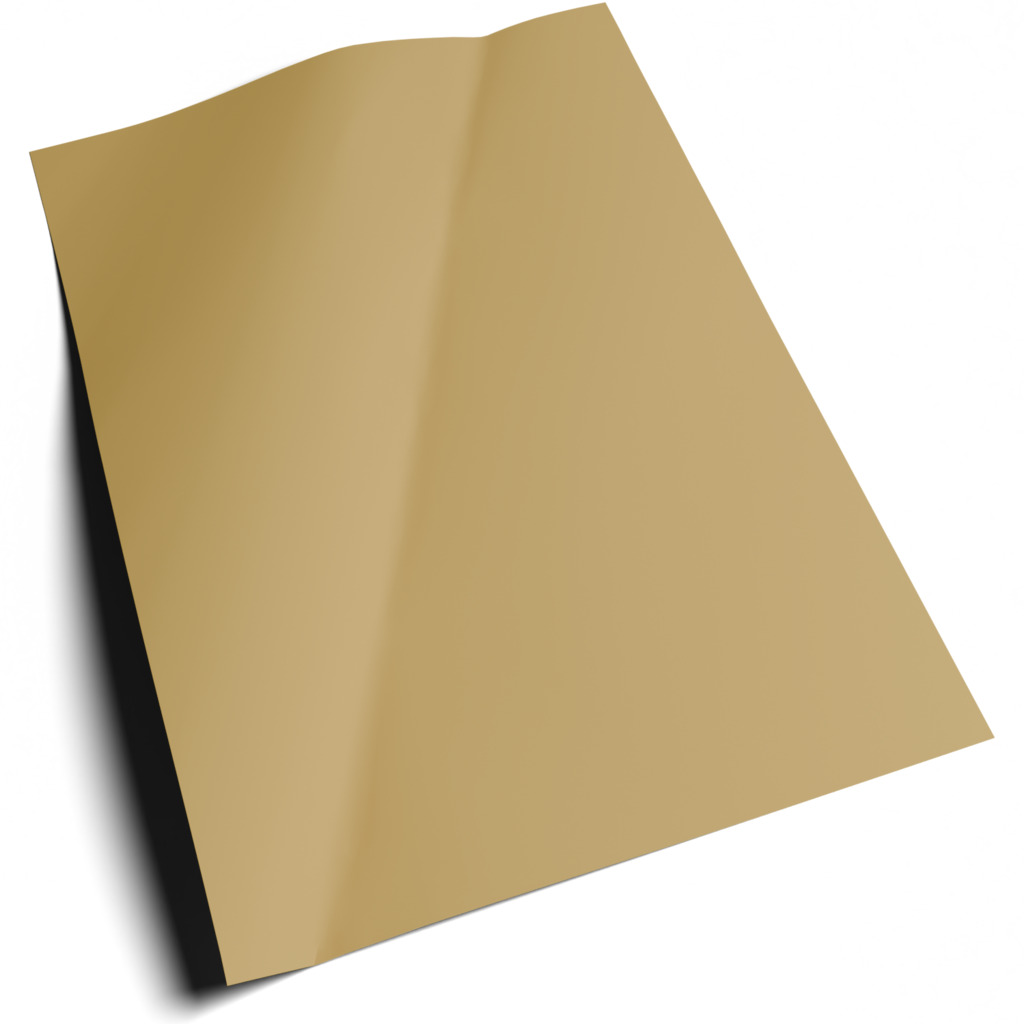}
    & \includegraphics[width=0.163\linewidth]{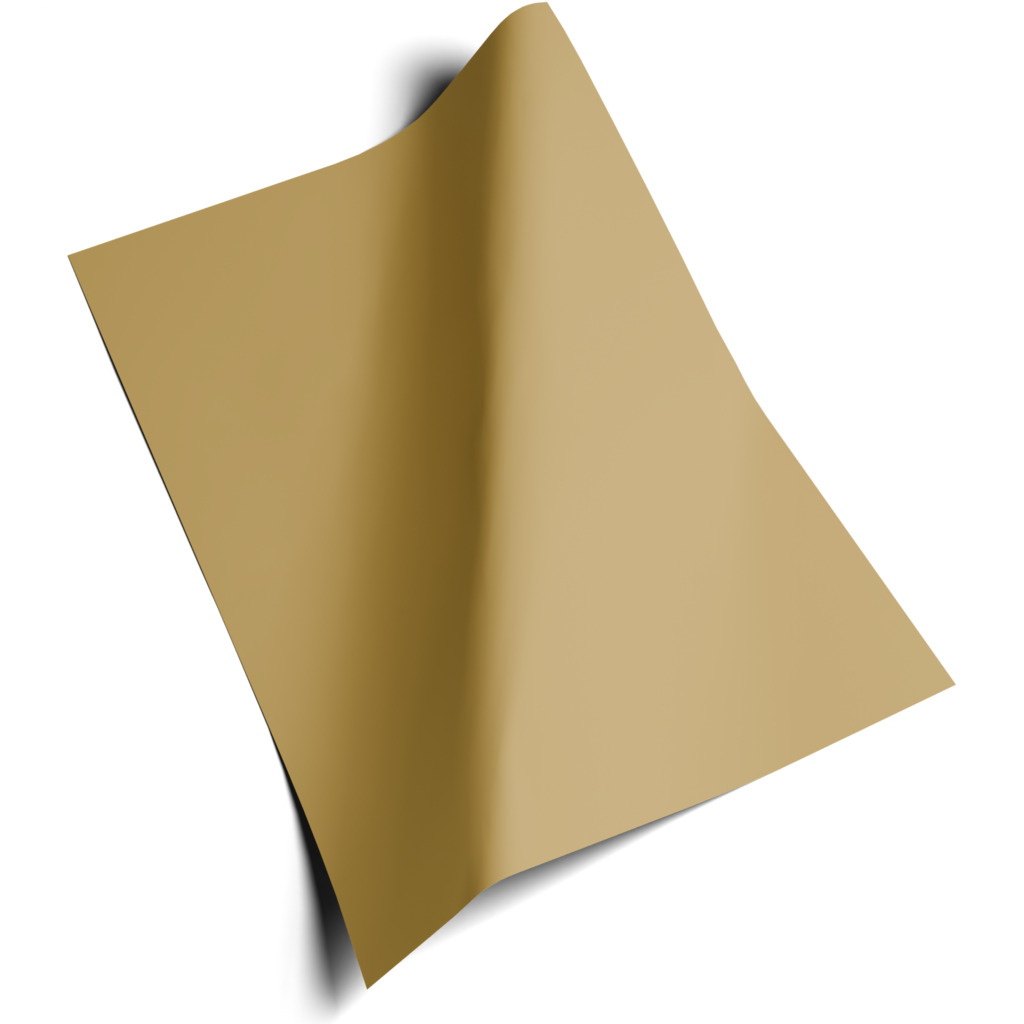}
    & \includegraphics[width=0.163\linewidth]{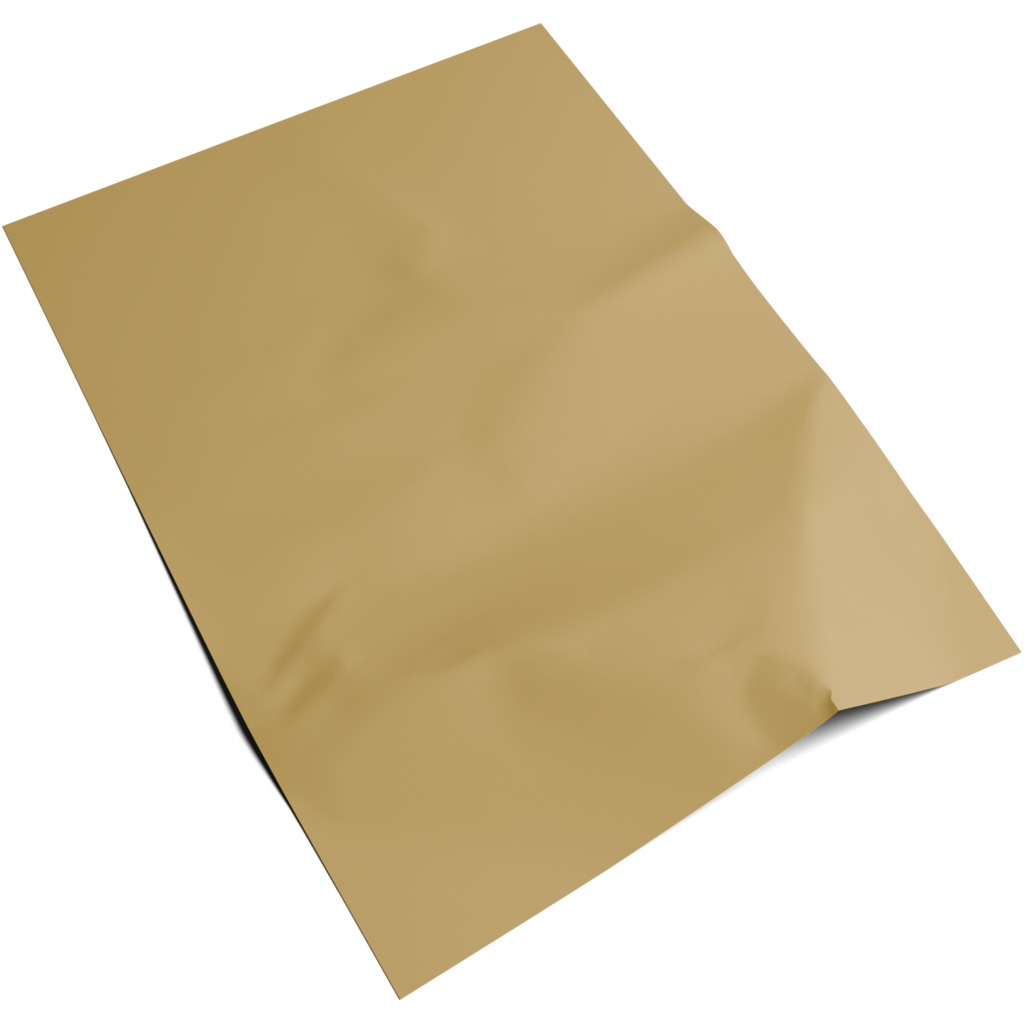}
    & \includegraphics[width=0.163\linewidth]{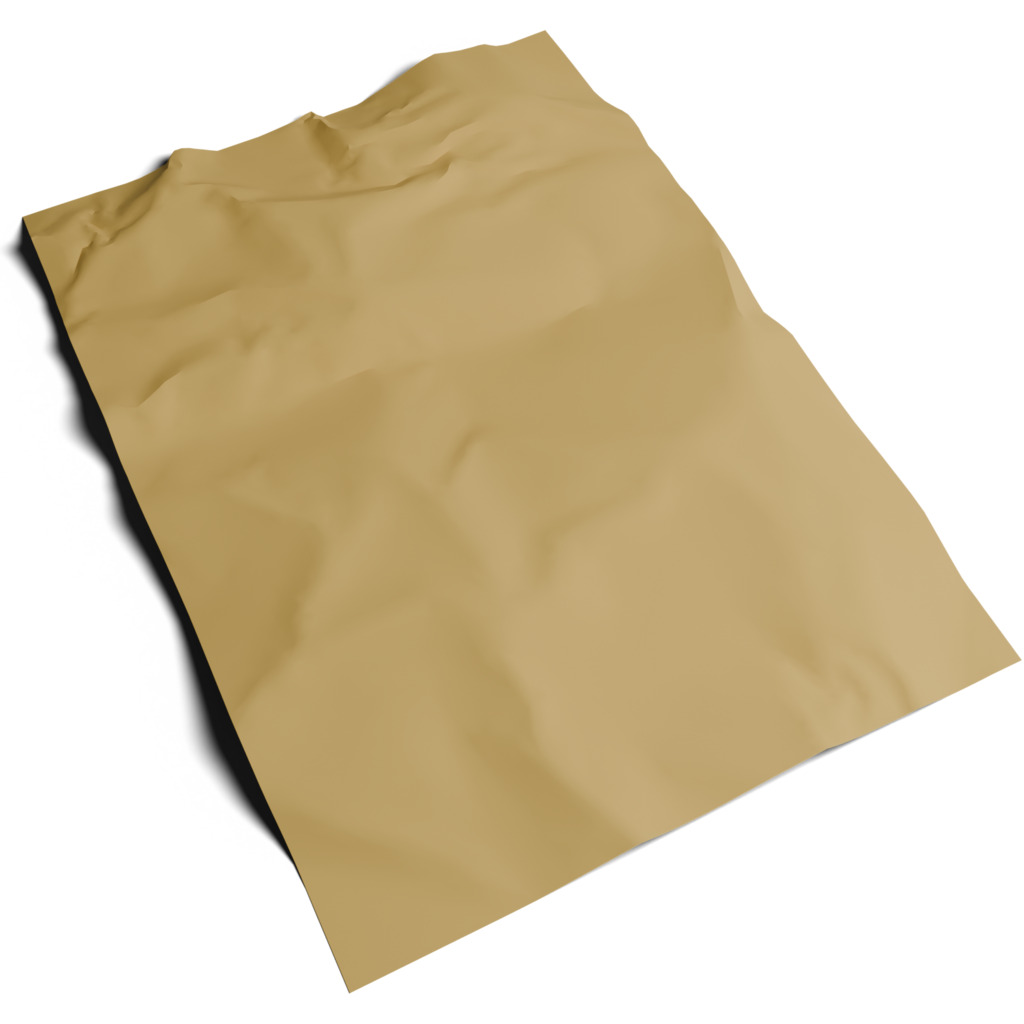}\\
    \scriptsize Pull &\scriptsize Fall on cylinder &\scriptsize Fold &\scriptsize Fall on gable &\scriptsize Fall on sphere &\scriptsize Fall on multiple
    \end{tabular}
    \caption{Meshes generated by our physics based simulation pipeline using the different scenarios.}
    \label{fig:meshes}
\end{figure}

The paper mesh, which represents the geometry of the document, is the most crucial component of each sample. It defines both the distortion in the document and its shading and cast shadows. To generate a dataset that is an order of magnitude larger than existing ones, we need a scalable approach to obtain these meshes. Prior methods based on 3D scanning~\cite{das.etal2019} or depth capture~\cite{verhoeven.etal2023} require manual work and are inherently unscalable.

To overcome this challenge, we rely on physical simulation. We utilize ArcSim~\cite{narain.etal2012, narain.etal2013}, a powerful adaptive simulator designed for sheets of deformable materials. While capable of simulating fabrics and plastics, it is particularly well-suited for paper. Paper exhibits a specific physical characteristic that makes it difficult to simulate: it does not stretch. ArcSim enforces this constraint, simulating paper while preventing unnatural stretching. In addition, regardless of its deformation, a sheet of paper mathematically forms a developable surface; ArcSim preserves this geometric property throughout the simulation.

To generate realistic meshes that accurately represent real-world document deformations, we design three distinct simulation scenarios. Each scenario is initialized with a flat, rectangular A4-sized mesh, a common format for documents. Examples of the resulting deformed document meshes are presented in \cref{fig:meshes}. 

\paragraph{Pull scenario.} The first scenario is straightforward and involves pulling on specific vertices of the original flat mesh. One or two vertices are randomly selected and displaced vertically to a random height. These control points can lie on the boundaries of the paper, mimicking a person picking up the document by its edge, or within its interior, which simulates pinching the paper. As only one or two vertices are displaced, this scenario predominantly produces smoothly curved documents, but it can still generate sharper features when lifting a central vertex. Meshes obtained with this scenario are presented in the first column of~\cref{fig:meshes}.

\begin{figure}[t]
    \centering
	\small
	\setlength{\tabcolsep}{0.5pt}
    \setlength{\fboxsep}{0pt}
    \begin{tabular}{cccccc}
    \includegraphics[width=0.163\linewidth]{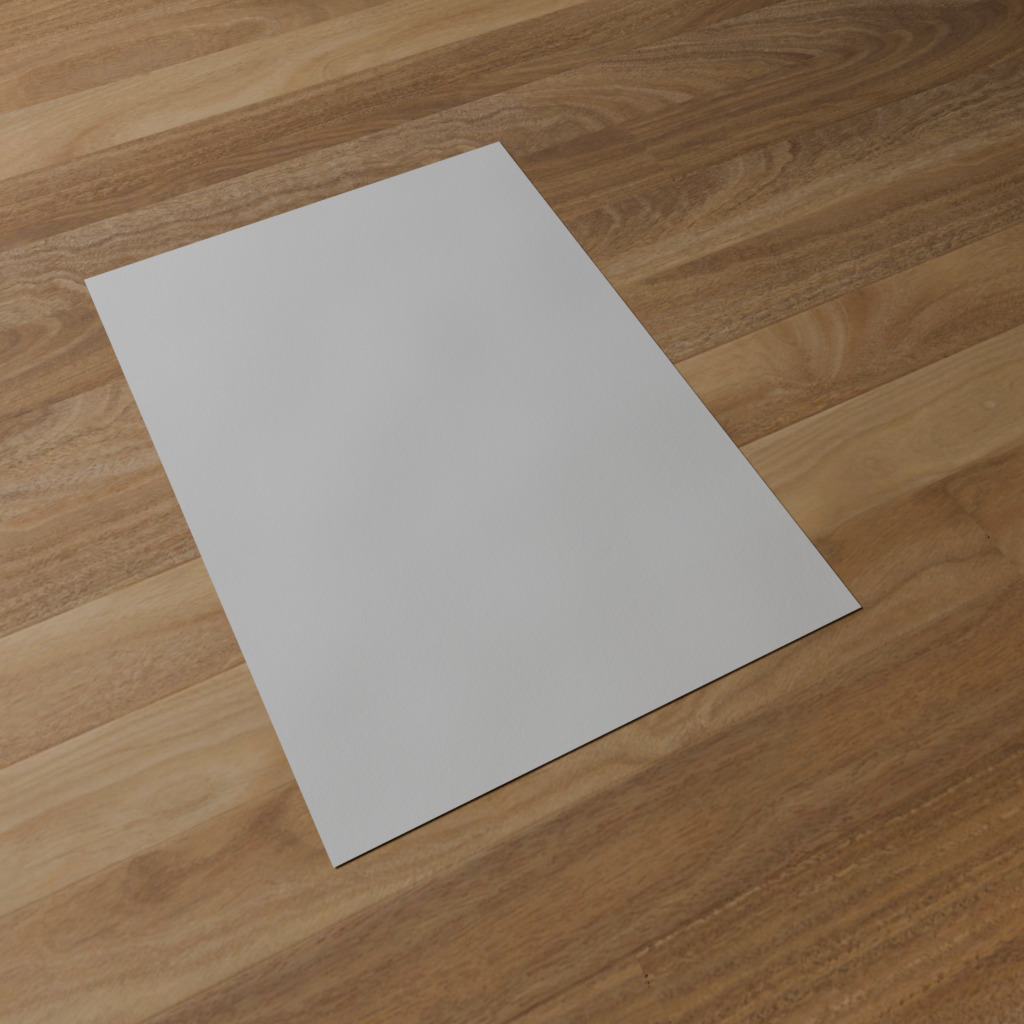}
    & \includegraphics[width=0.163\linewidth]{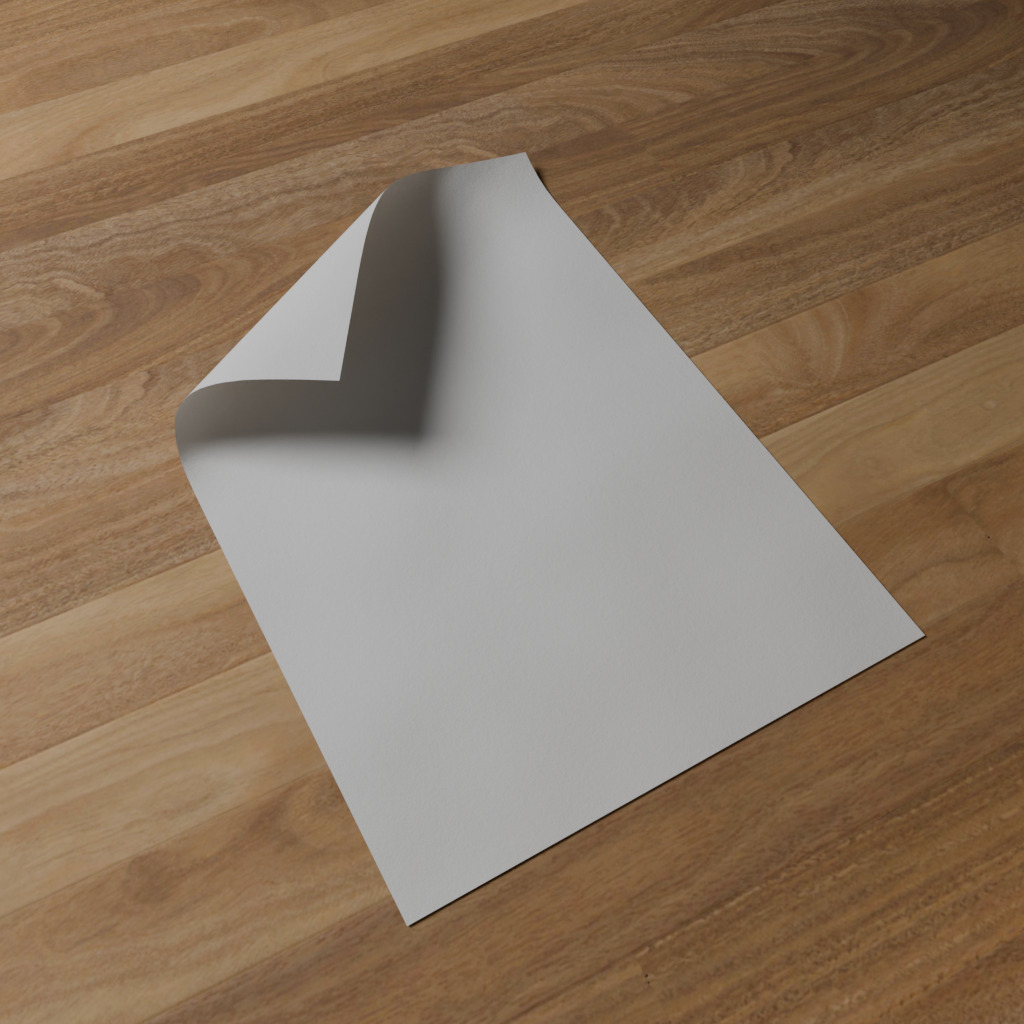}
    & \includegraphics[width=0.163\linewidth]{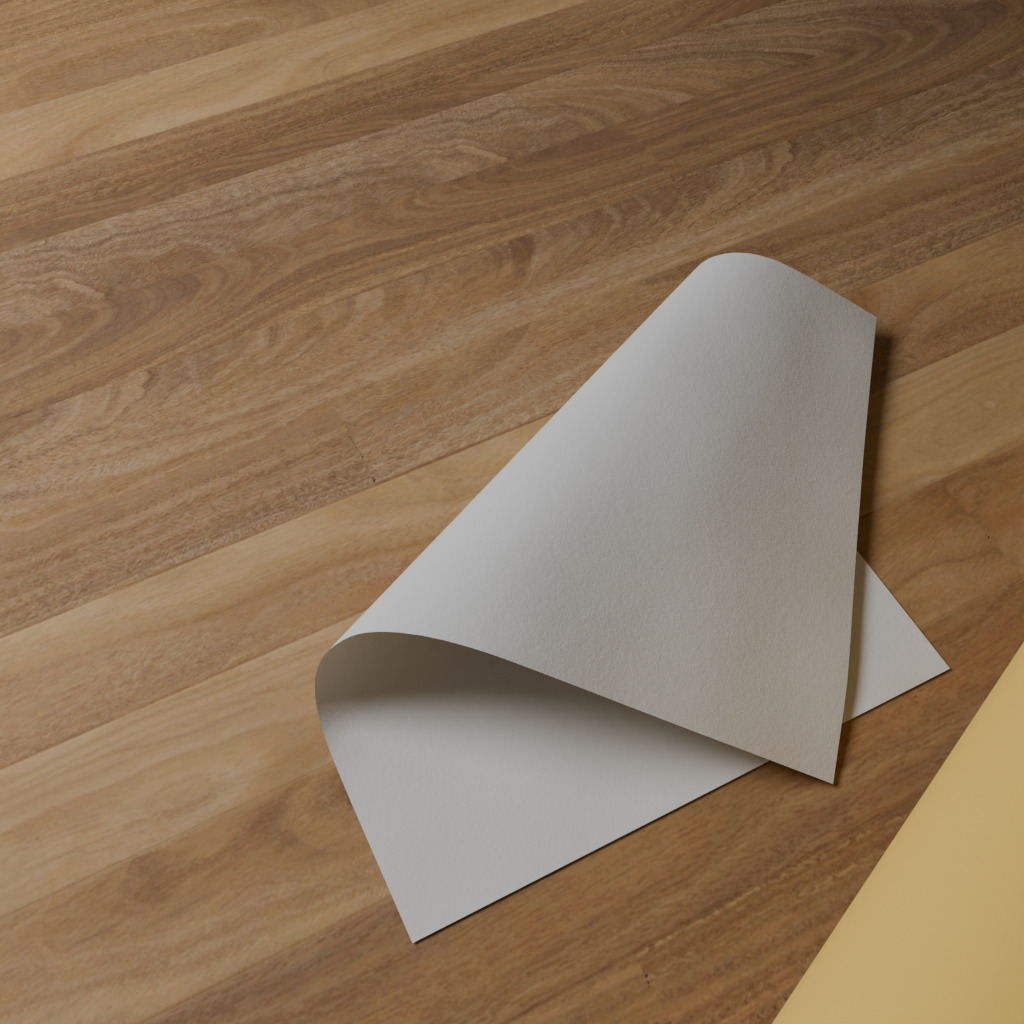}
    & \includegraphics[width=0.163\linewidth]{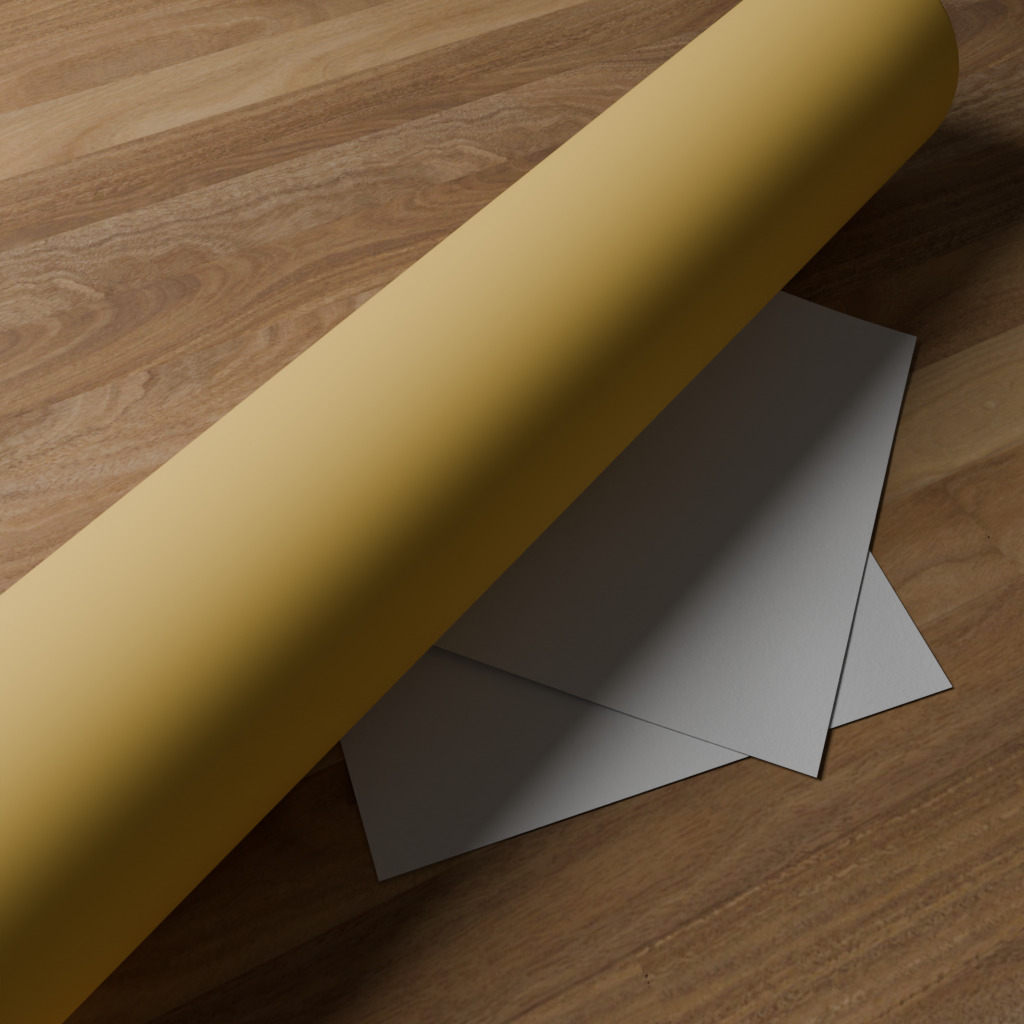}
    & \includegraphics[width=0.163\linewidth]{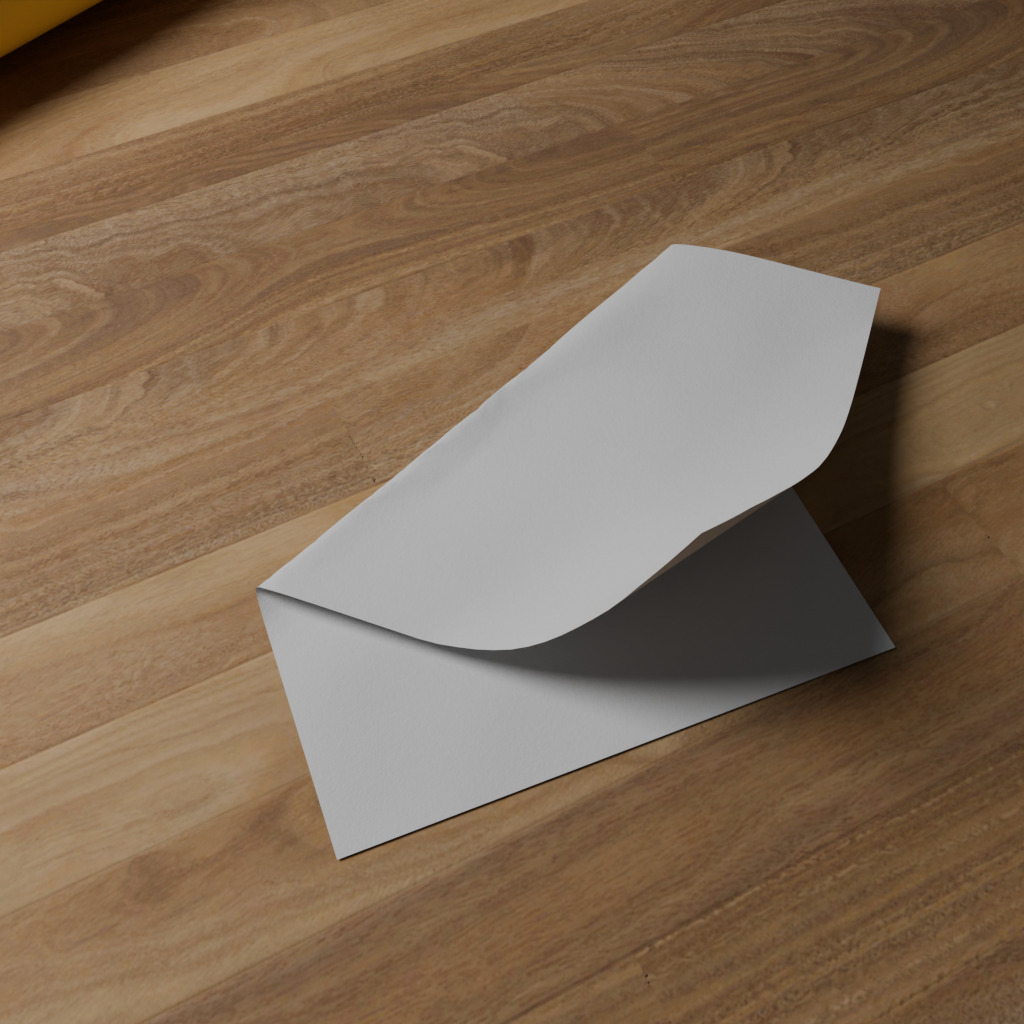}
    & \includegraphics[width=0.163\linewidth]{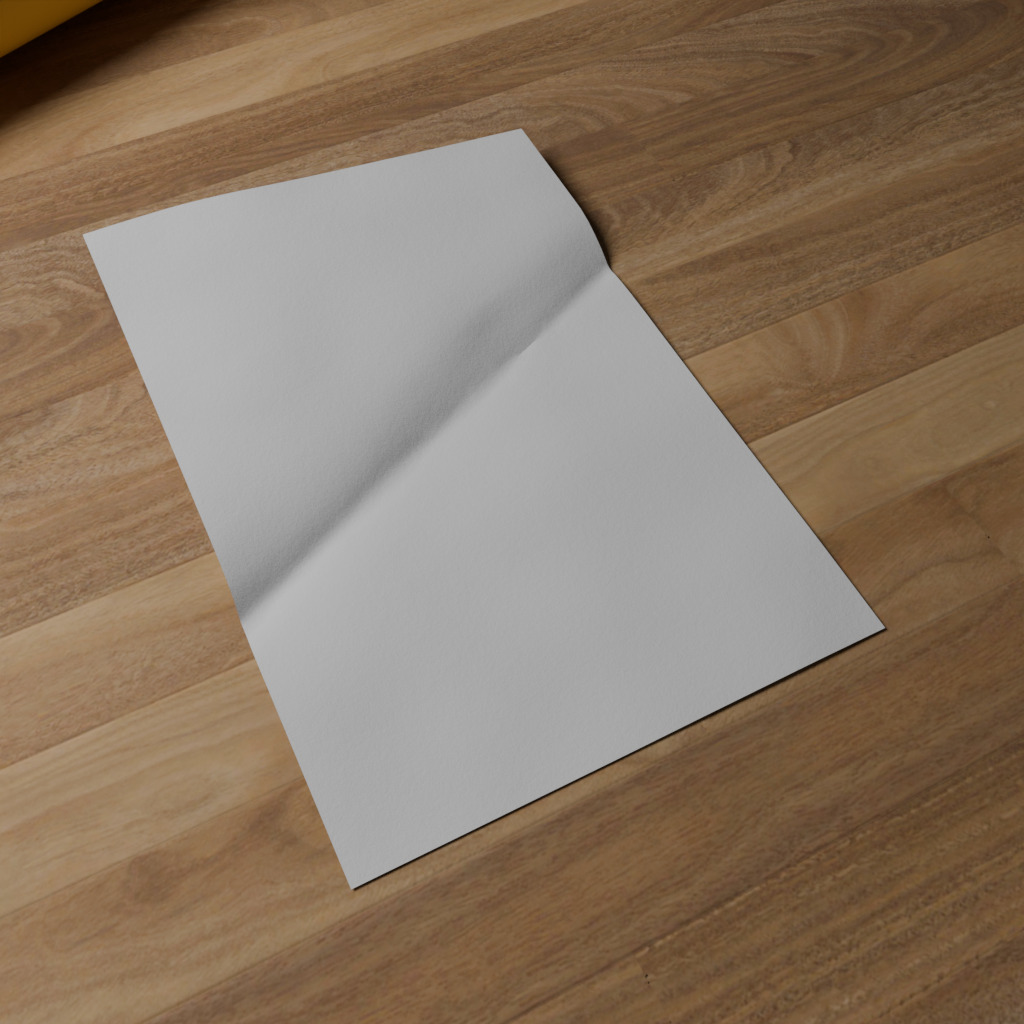}
    \end{tabular}
    \caption{The process of simulating a folded document. Starting from a flat mesh, a boundary vertex is pulled over another one. A roller simulates a hand flattening the crease, after which the mesh is unfolded. Note that the material textures shown here are for aesthetic purposes only.}
    \label{fig:fold}
\end{figure}

\paragraph{Fold scenario.} The second scenario is designed to simulate folded documents by mimicking the physical folding process. To that end, two boundary vertices are selected, and one is pulled over the other. Then, a rigid cylinder is rolled over the resulting mesh to flatten it and form a crease. Finally, the paper is unfolded by returning the displaced vertex back to its original position. This process, illustrated in~\cref{fig:fold}, creates convincing geometries with one fold. Examples of the resulting meshes are shown in the third column of~\cref{fig:meshes}. 

\begin{wrapfigure}[14]{r}{0.3\textwidth}
    \centering
	\small
    \includegraphics[width=\linewidth]{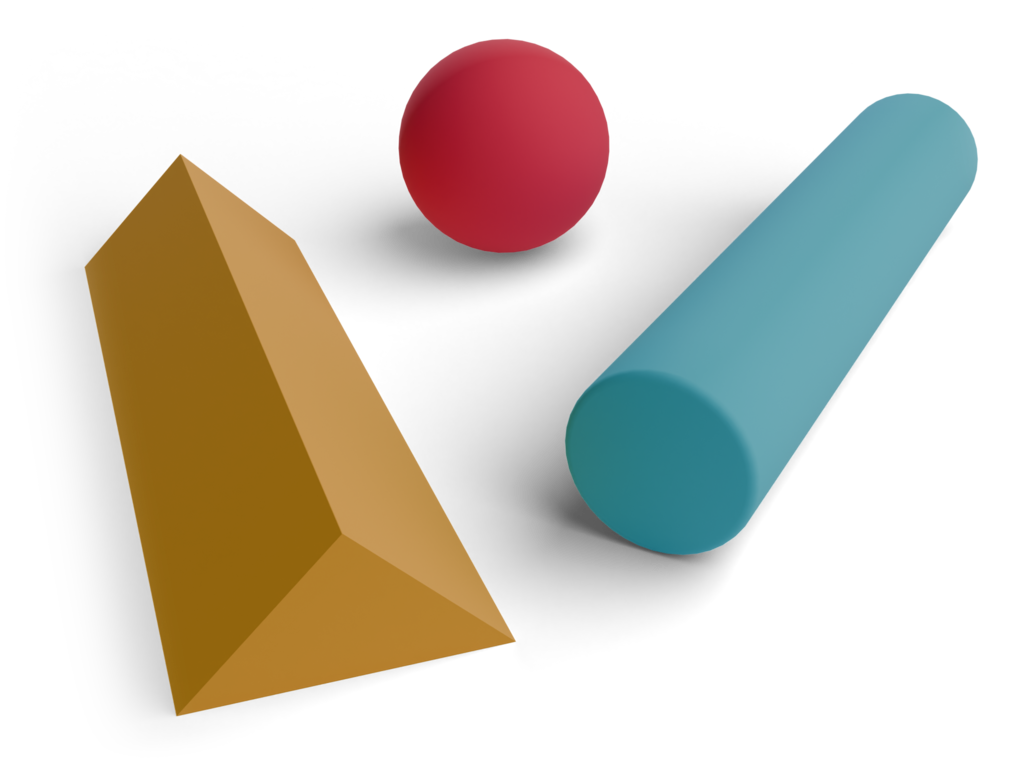}
    \caption{Primitives used for the fall scenario.}
    \label{fig:obstacles}
\end{wrapfigure}
\paragraph{Fall scenario.} The last scenario, while the simplest, is capable of generating the largest variety of outputs. In this setup, we simulate the document falling onto various 3D primitives under the influence of gravity. The outputs vary widely depending on the shape of the collider and the gravitational acceleration. We use three types of colliders: spheres, cylinders and gables (see~\cref{fig:obstacles}). Dropping the paper onto a cylinder produces a smoothly curved surface (\cref{fig:meshes},~second column). When falling onto a gable, the mesh contains much sharper, fold-like deformations (\cref{fig:meshes}, fourth column). Draping the document over a sphere forces the generation of complex wrinkles, as the sphere's non-zero Gaussian curvature geometrically conflicts with the developable nature of the paper (\cref{fig:meshes}, fifth column). Lastly, letting the paper fall onto multiple small spheres creates many creases, simulating a heavily crumpled document (\cref{fig:meshes}, last column). In all these simulations, the scale, position and orientation of the primitives are randomized. In the last two cases, we apply highly randomized gravitational acceleration to create various crease patterns.

\paragraph{} Together, these simulation scenarios enable the generation of curved (\textit{Pull}, \textit{Fall on cylinder}), folded (\textit{Fold}, \textit{Fall on gable}) and crumpled (\textit{Fall on sphere}, \textit{Fall on multiple spheres}) documents. For each of these six configurations, we generate nearly 50,000 unique meshes. Each mesh can be flipped to simulate viewing from the opposite side of the paper sheet. This simple operation produces vastly different appearance, ultimately resulting in a total of 584,708 distinct deformed document geometries.

\subsection{Rendering}

To generate high-quality samples for the final dataset, we import the warped paper meshes from the previous step into Blender and set up a unique scene configuration for each sample. The scene setup involves positioning the virtual camera, placing explicit light sources, defining a background surface for the paper to lie on and applying a physically based paper material paired with a document texture to the paper mesh. This scene is then rendered using Cycles, Blender's path-tracing engine~\cite{blender}. We opt for the more computationally expensive path tracing over rasterization to produce physically realistic images, which is especially important to accurately capture the self-shadowing and specular highlights of warped documents. We render the images and their corresponding ground truths at a resolution of $1024 \times 1440$ pixels with 128 render samples per pixel as a tradeoff between scalability and visual quality.

\paragraph{Camera.} For each sample, we select a camera position from a set of valid viewing angles. An angle is considered valid if the document is fully contained within the camera's field of view and all mesh face normals are consistently front-facing. This ensures that the entire document is visible and completely free of self-occlusion. The candidate angles are constrained to a moderate inclination range around the top-down view of the paper, closely reflecting real-world capture conditions for document images.

\begin{figure}[t]
    \centering
	\small
	\setlength{\tabcolsep}{0.5pt}
    \setlength{\fboxsep}{0pt}
    \begin{tabular}{cccc}
      \includegraphics[width=0.236\linewidth]{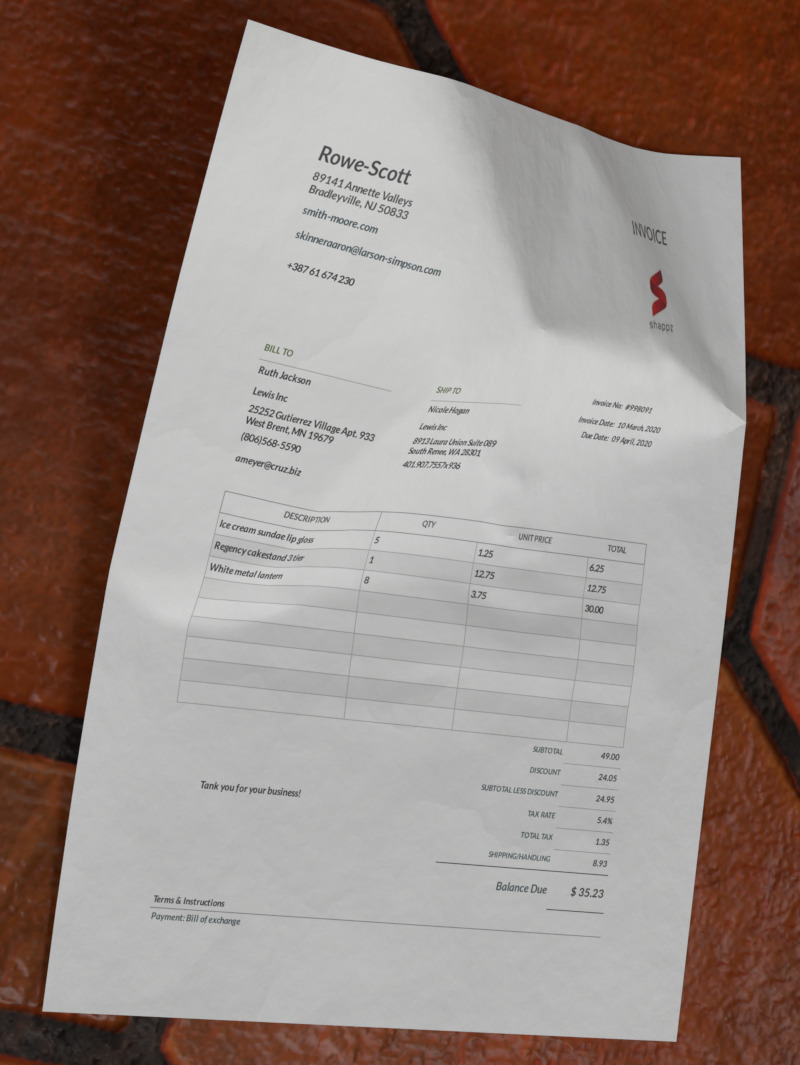}
    & \includegraphics[width=0.236\linewidth]{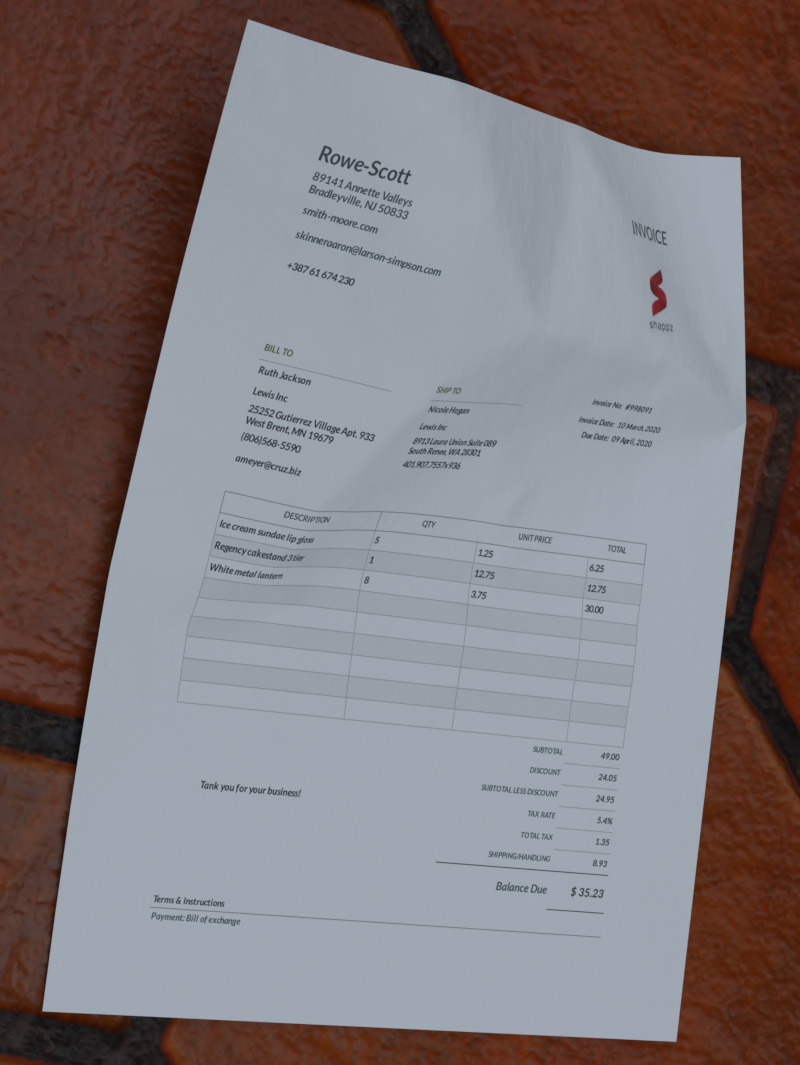}
    & \includegraphics[width=0.236\linewidth]{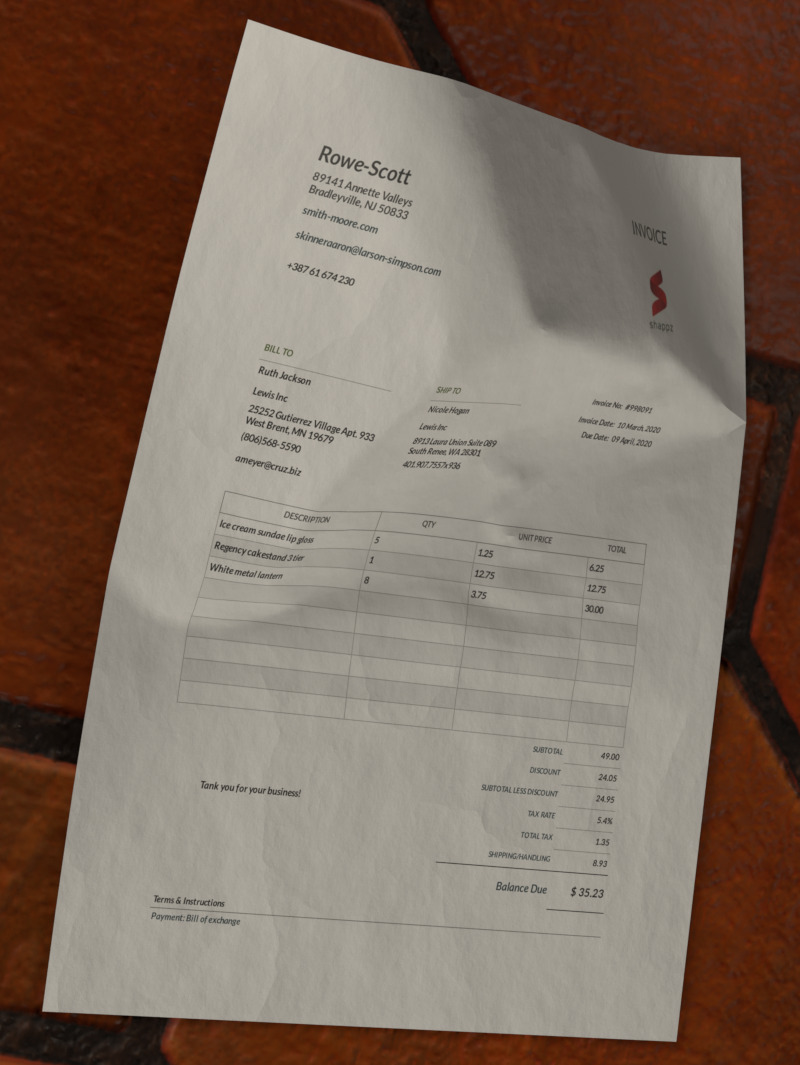}
    & \includegraphics[width=0.236\linewidth]{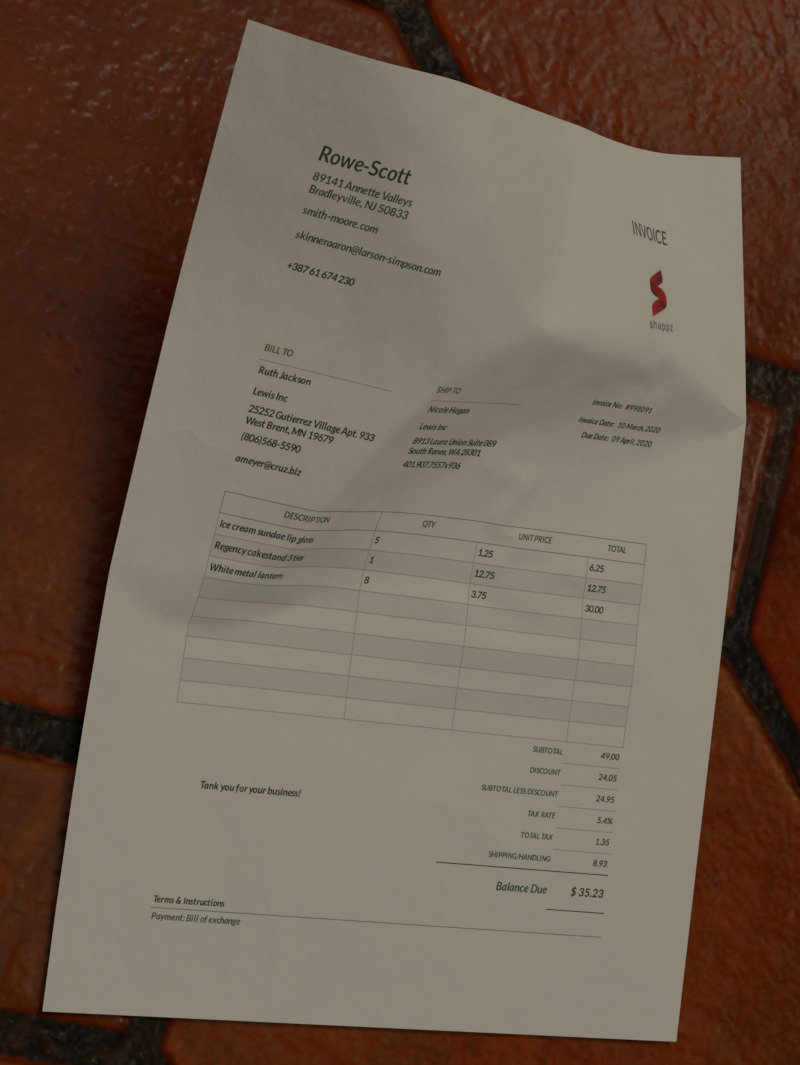}
    \end{tabular}
    \caption{Various lighting setups. From left to right: \textit{neutral 3-point lighting}, \textit{cool softbox lighting}, \textit{warm rim lighting} and \textit{warm natural lighting}.}
    \label{fig:light}
\end{figure}

\paragraph{Lighting.} We use four predefined light setups, with randomized parameters, to cover a wide range of real-world conditions. Each setup can be globally rotated around the mesh, and the color temperature of the lights can be adjusted. The four setups range from controlled studio environments to natural daylight (see \cref{fig:light}): \textit{3-point lighting} mimics a classic studio setup with key-, fill- and backlight, producing neutral illumination with soft, controlled shadows; \textit{Softbox lighting} utilizes several large area lights that cast very soft shadows, as is common in photography or indoor settings; \textit{Rim lighting} employs strong backlights on either side of the document, creating sharp edge highlights and hard shadows;  \textit{Natural lighting} uses strong directional sunlight to simulate daylight, replicating window-lit or outdoor environments. We use these explicit light sources rather than image-based lighting (i.e., illumination derived from HDR environment maps) to retain more control over the scene and to generate more pronounced specular highlights and shadows on the paper surface. 

\paragraph{Background material.} The background of the rendered image for each sample is a simple plane with a PBR (physically based rendering) material. The textures are sourced from the MatSynth dataset~\cite{vecchio.deschaintre2024}, which contains 5,789 high-quality PBR materials, including those captured by Deschaintre et al.~\cite{deschaintre.etal2018}, all under CC0 or CC-BY licenses. To maximize the diversity within the backgrounds of the \ours{} dataset, we uniformly sample materials and randomize their rotation, scale and spatial offset parameters. Furthermore, to guarantee that the document rests naturally on the background plane and casts physically accurate shadows, we run a simple rigid-body simulation before rendering.

\begin{figure}[t]
    \centering
	\small
	\setlength{\tabcolsep}{0.5pt}
    \setlength{\fboxsep}{0pt}
    \begin{tabular}{cccc}
      \includegraphics[width=0.236\linewidth]{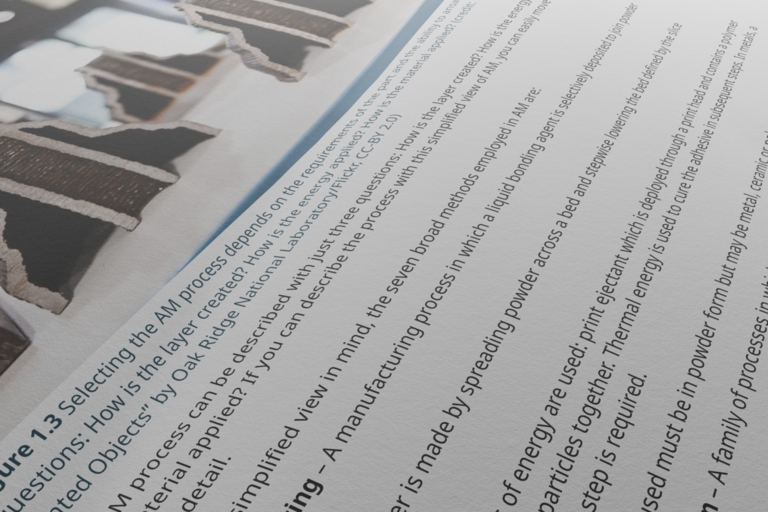}
    & \includegraphics[width=0.236\linewidth]{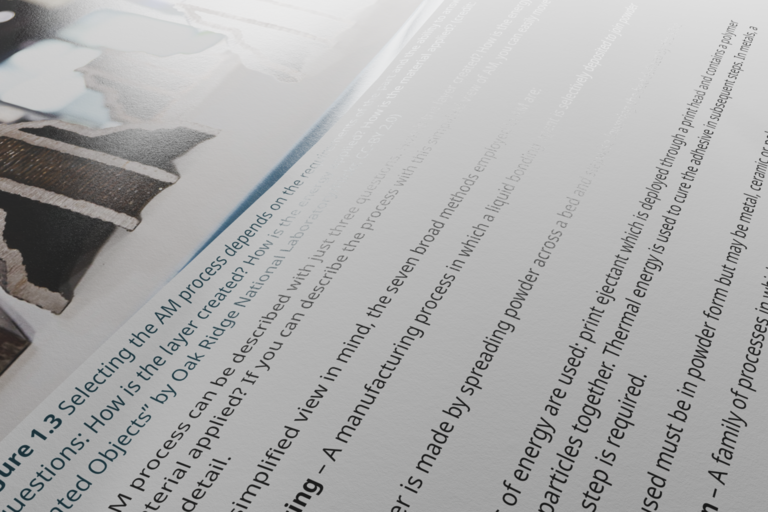}
    & \includegraphics[width=0.236\linewidth]{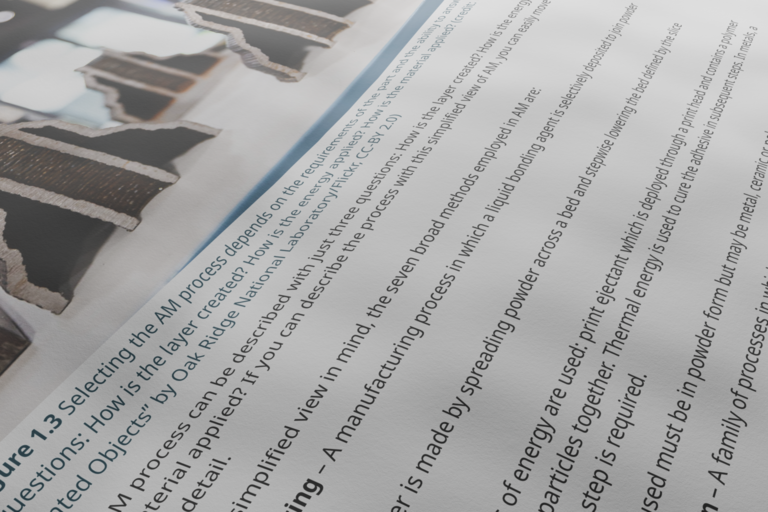}
    & \includegraphics[width=0.236\linewidth]{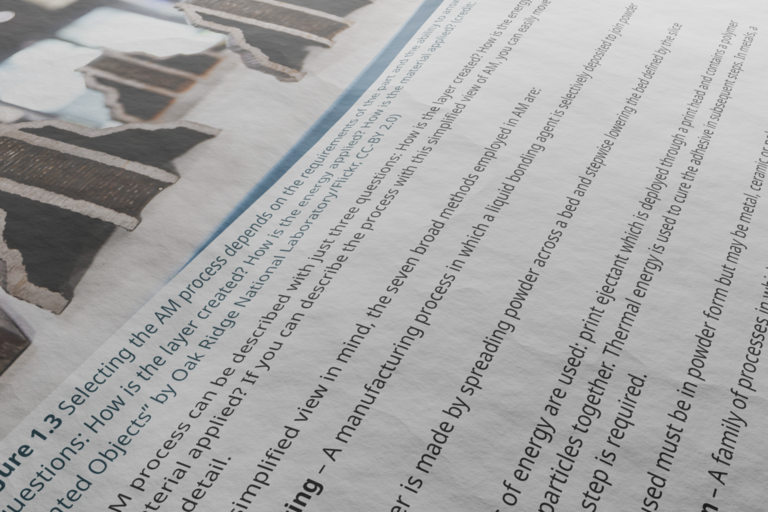}
    \end{tabular}
    \caption{Variation in the paper texture (zoomed view to highlight the material texture). From left to right: basic grainy paper texture, paper texture with glossy ink, slightly wavy paper texture, creased paper texture.}
    \label{fig:paper_material}
\end{figure}

\paragraph{Procedural paper material.} To make the rendered images as realistic as possible and mitigate the synthetic-to-real domain gap, we develop a highly parameterized procedural paper material using Blender's shader node system~\cite{blender}. The surface imperfections are driven by a composite height map, which is subsequently converted into a normal map. This height map is created by combining three procedural components: a high-frequency Perlin noise \cite{perlin1985} to simulate the microscopic fibers of the paper with a grainy texture, a low-frequency Perlin noise to model the macroscopic waving of the paper and a Worley noise \cite{worley1996} to generate micro-creases. In addition, we simulate glossy ink by modulating the surface roughness based on the document image texture, which also provides the albedo. The scale and strength of each component can be adjusted, making the paper material highly versatile. Examples of the resulting paper textures are presented in \cref{fig:paper_material}.

\begin{figure}[t]
    \centering
    \includegraphics[width=0.98\linewidth]{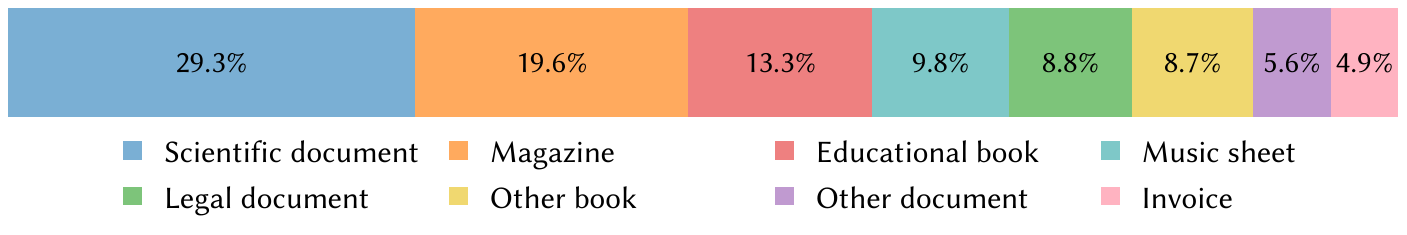}
    \caption{Proportion of each type of document in our \ours{} dataset.}
    \label{fig:document_type_count}
\end{figure}

\paragraph{Documents.} To make the \ours{} dataset as comprehensive as possible, we render samples using a highly diverse set of document textures. In total, we collect 510,903 portrait-format document pages distributed across eight categories (see \cref{fig:document_type_count}). We gather documents from various sources, all under permissive licenses to guarantee broad usability. Scientific papers are drawn from various arXiv~\cite{arxiv} categories and are filtered for CC-BY licensing. Educational materials are sourced from OpenStax~\cite{openstax}, while general books spanning multiple languages are obtained from the Directory of Open Access Books (DOAB)~\cite{doab}. We also incorporate sheet music from IMSLP~\cite{imslp}, enterprise documents from RealKIE~\cite{RealKIE} and procedurally generated invoices from Inv3D~\cite{hertlein.etal2023}. Additional documents, including math problems, diagrams and charts, are extracted from the CoSyn-400K dataset~\cite{cosyn-400k}. Finally, due to the scarcity of freely available magazine pages, we follow the approach of Verhoeven et al.~\cite{verhoeven.etal2023} and use a text-to-image model to generate these. Specifically, we first create a templated prompt that we fill with magazine-specific attributes, refine it with a large language model (Gemini 3 Flash~\cite{gemini}) and generate the final image using FLUX.2-klein-9B~\cite{flux2}. While the text in these synthetic images may be illegible, the layout and structure of the documents closely match those of real magazines.

\begin{figure}[t]
    \centering
	\small
	\setlength{\tabcolsep}{0.5pt}
	\renewcommand{\arraystretch}{0.0}  
    \setlength{\fboxsep}{0pt}
    \begin{tabular}{cccccc}
    \includegraphics[width=0.163\linewidth]{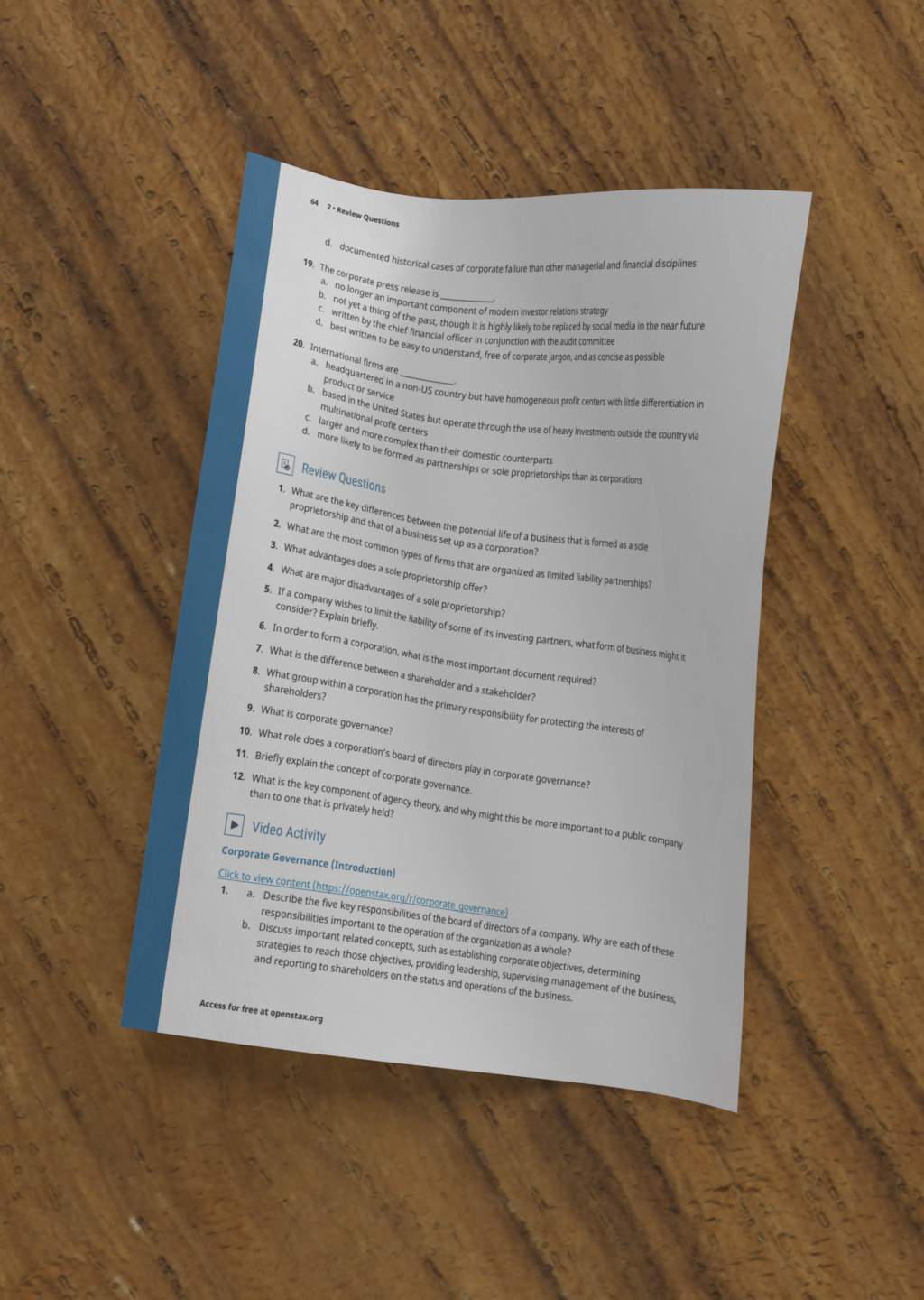}
    & \includegraphics[width=0.163\linewidth]{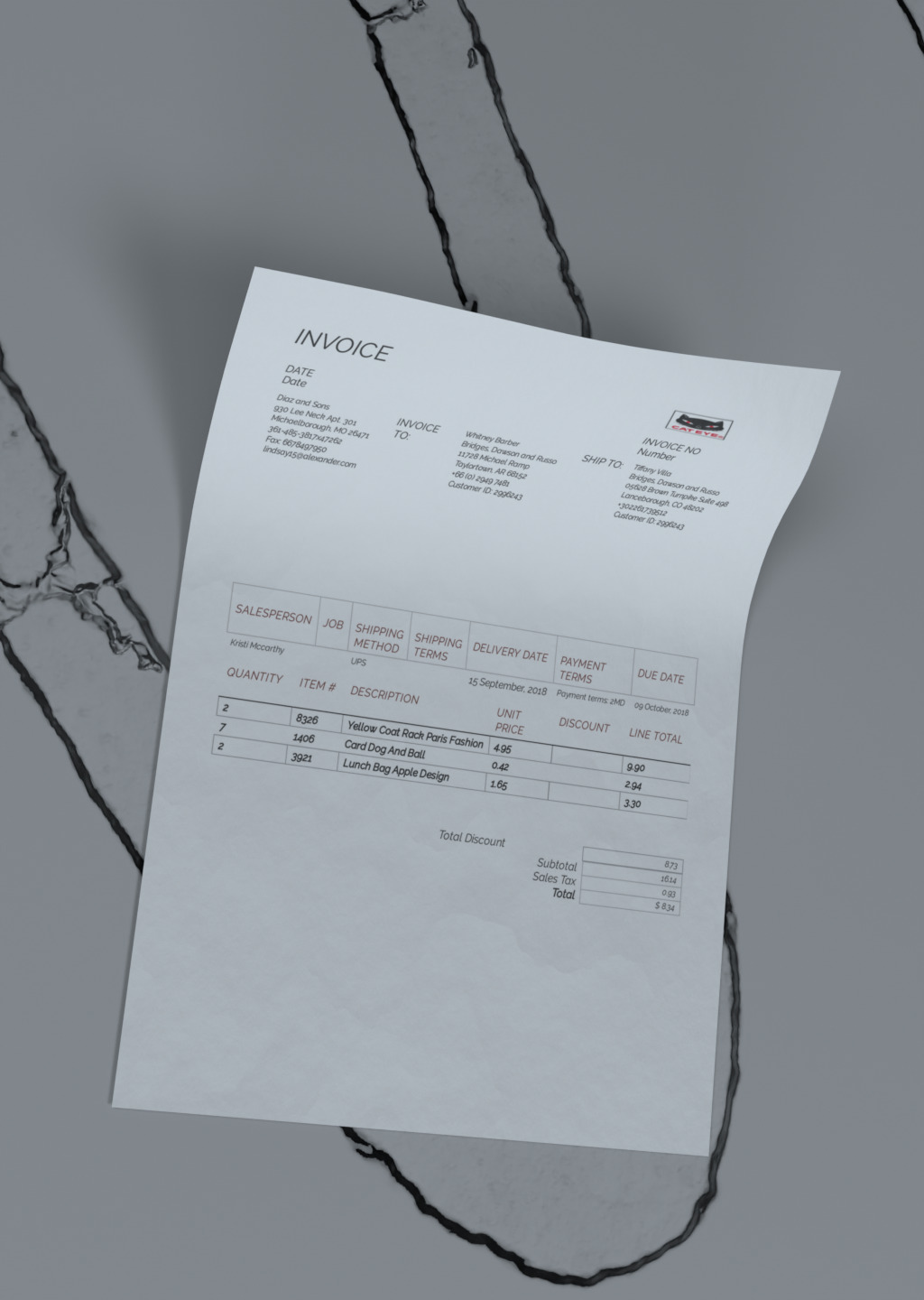}
    & \includegraphics[width=0.163\linewidth]{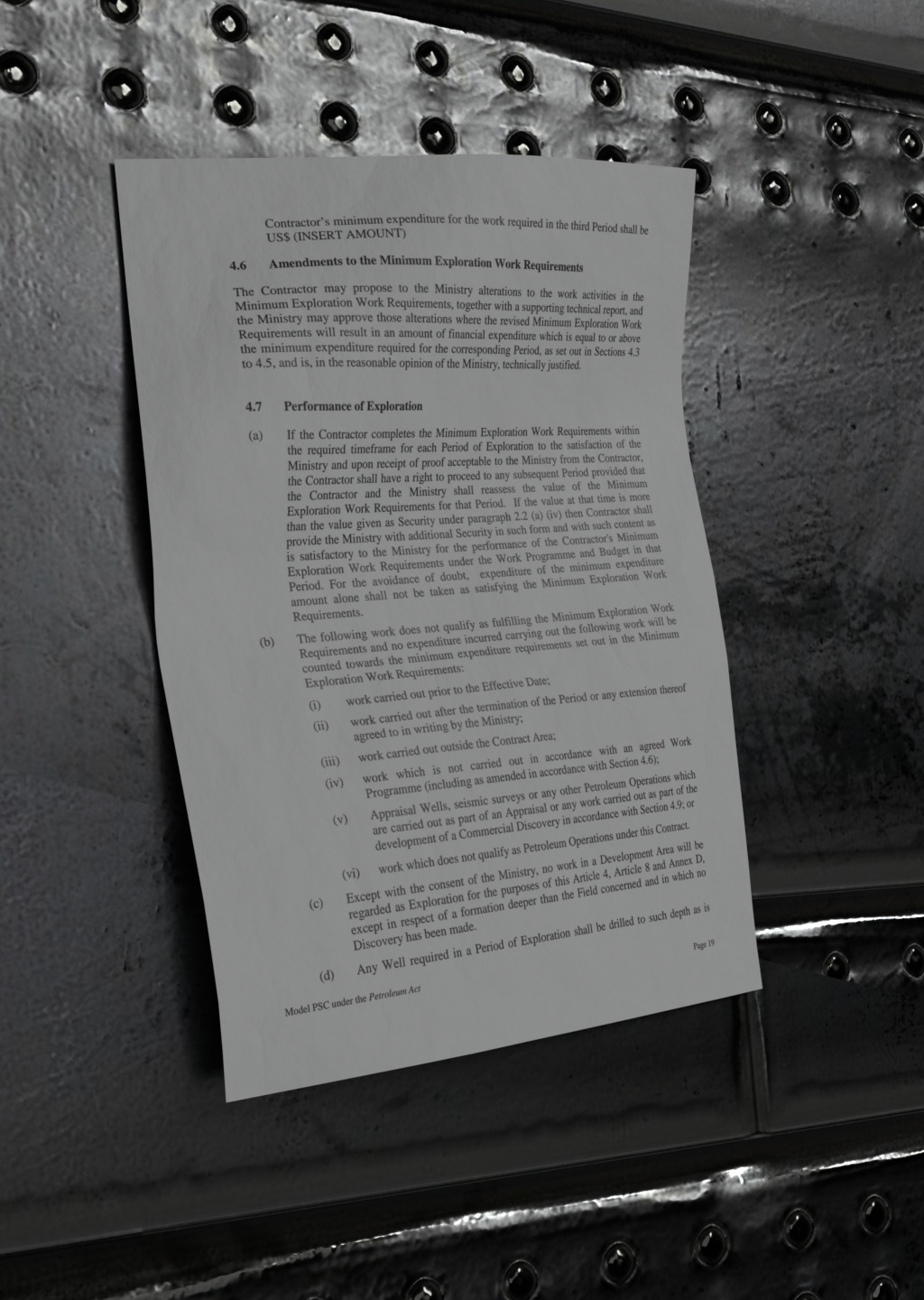}
    & \includegraphics[width=0.163\linewidth]{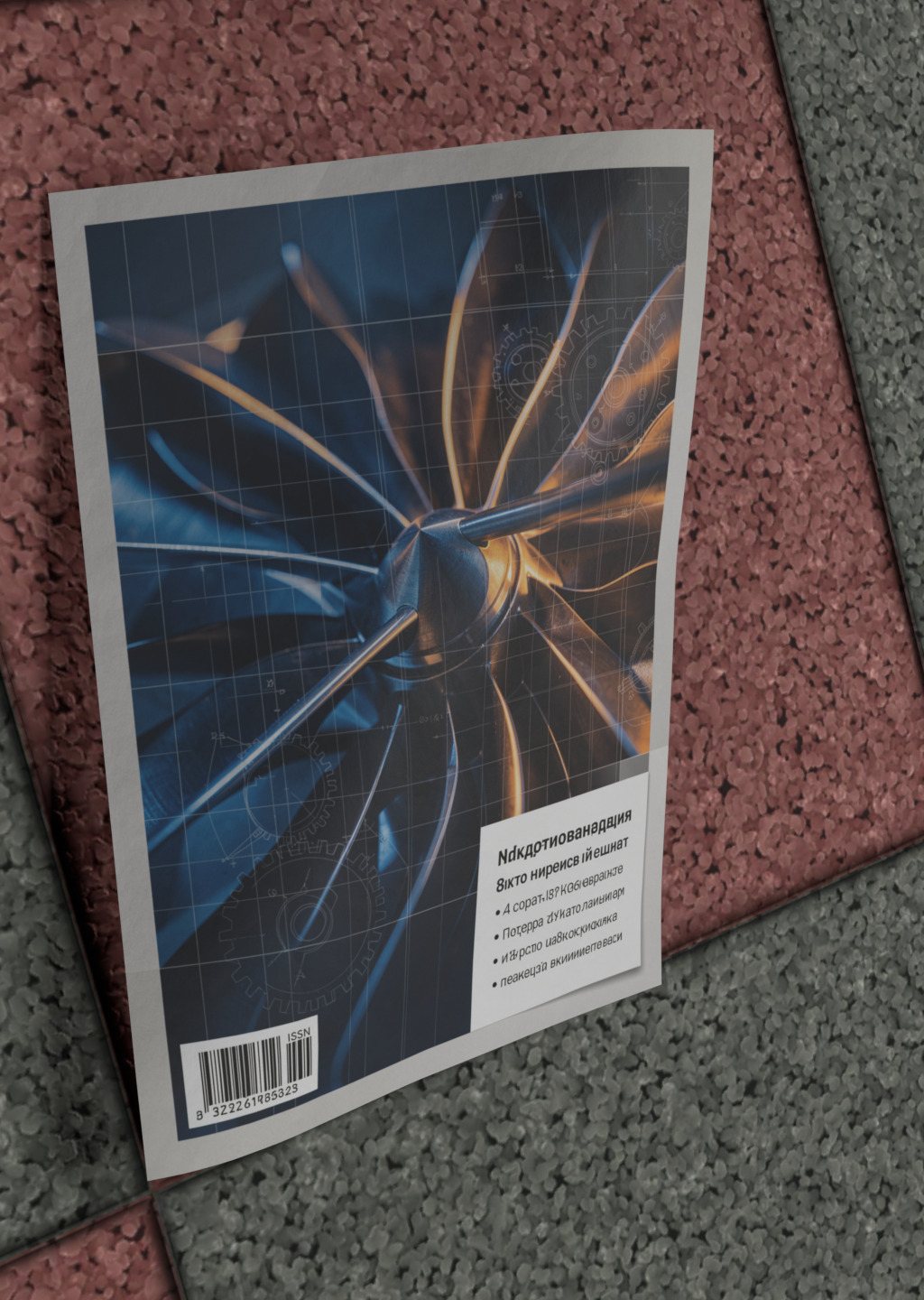}
    & \includegraphics[width=0.163\linewidth]{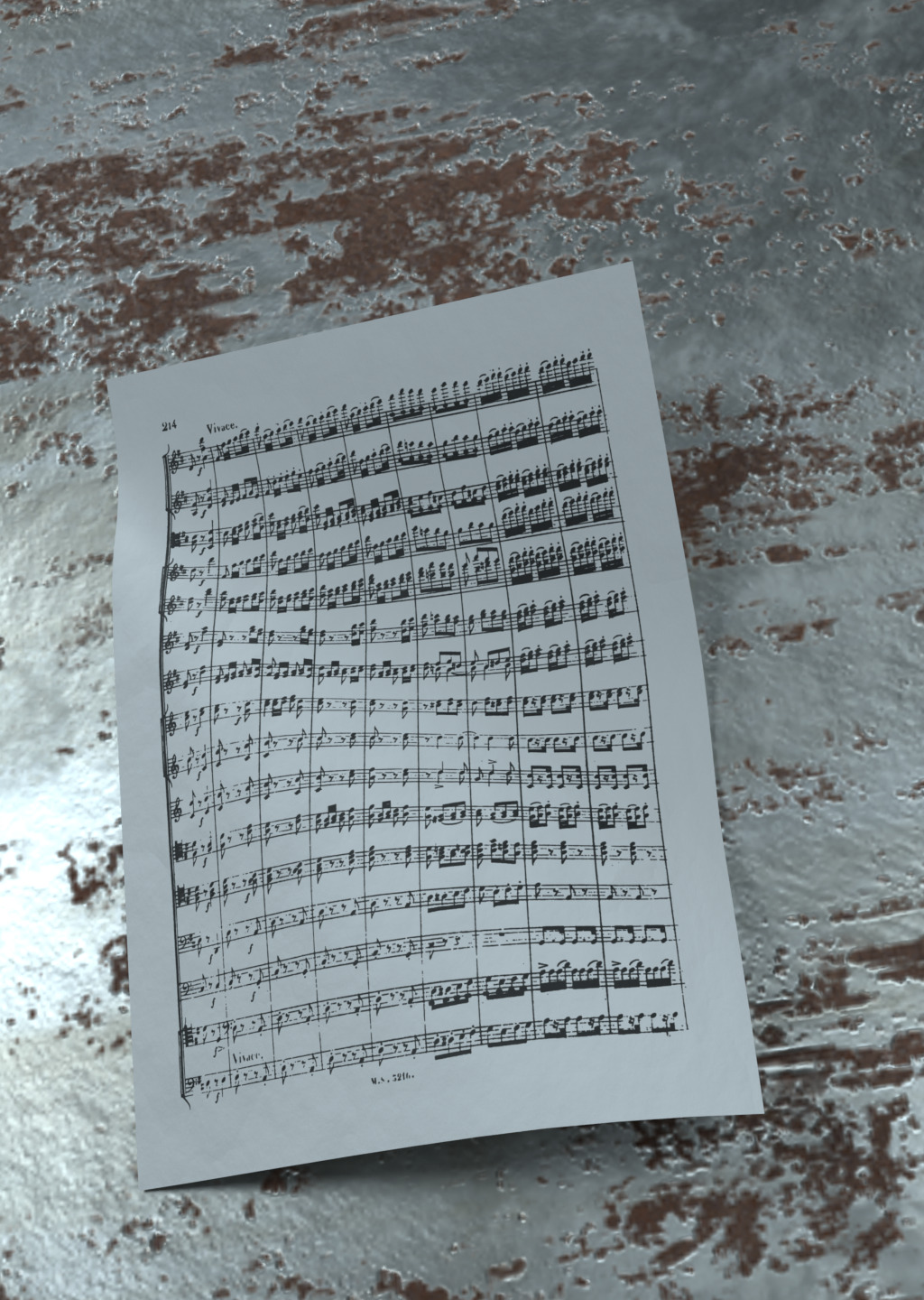}
    & \includegraphics[width=0.163\linewidth]{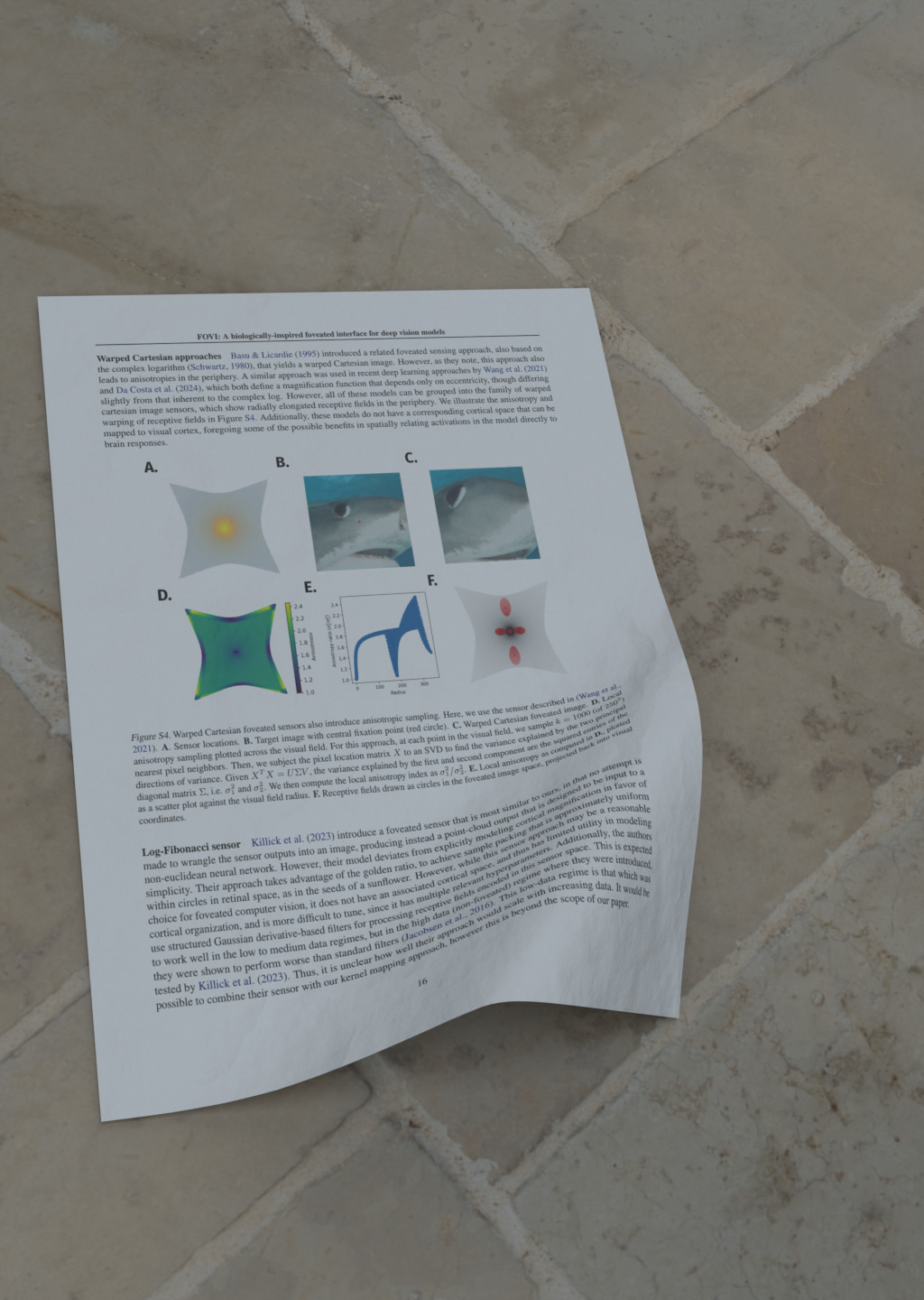}\\[1pt]
    \includegraphics[width=0.163\linewidth]{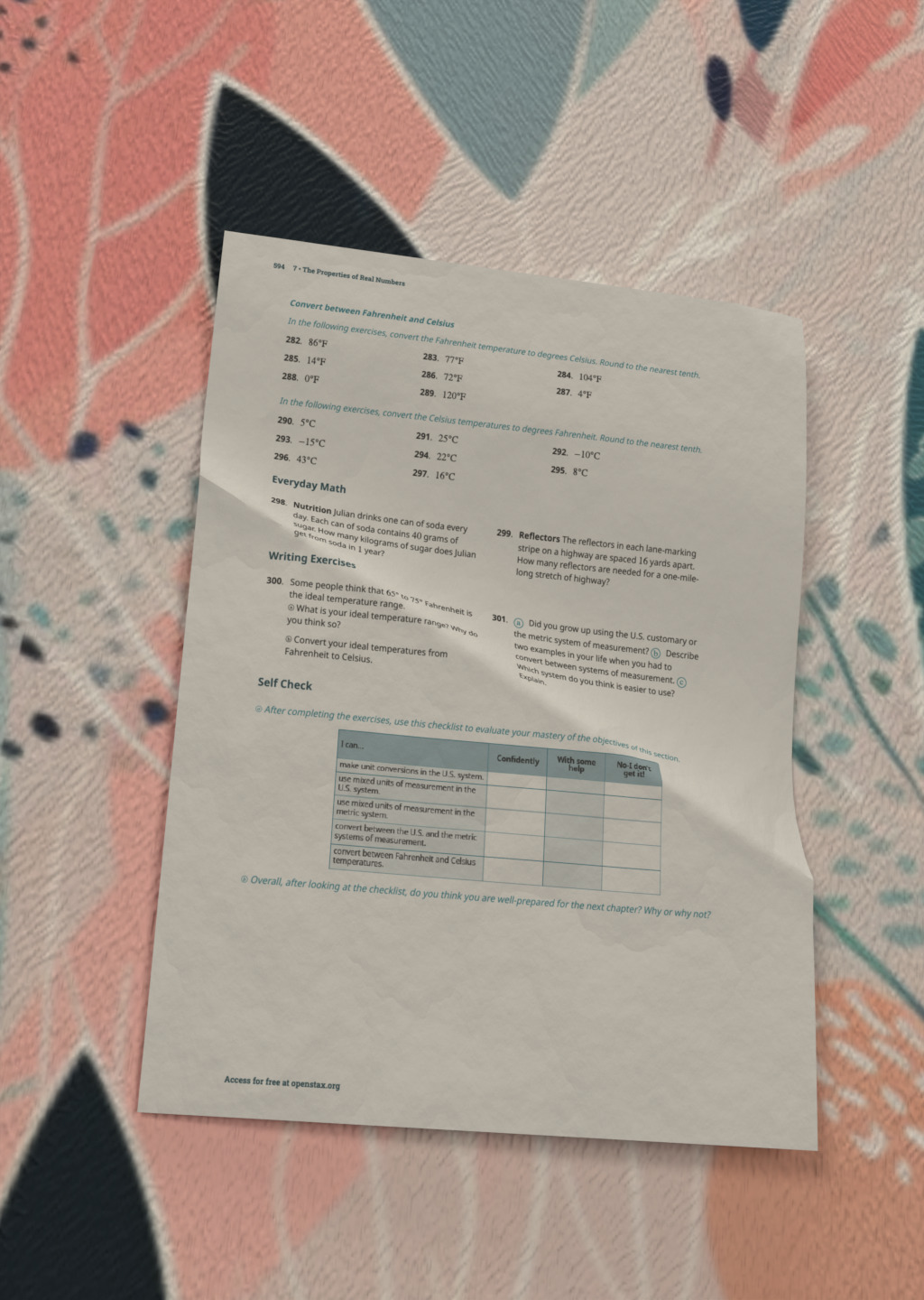}
    & \includegraphics[width=0.163\linewidth]{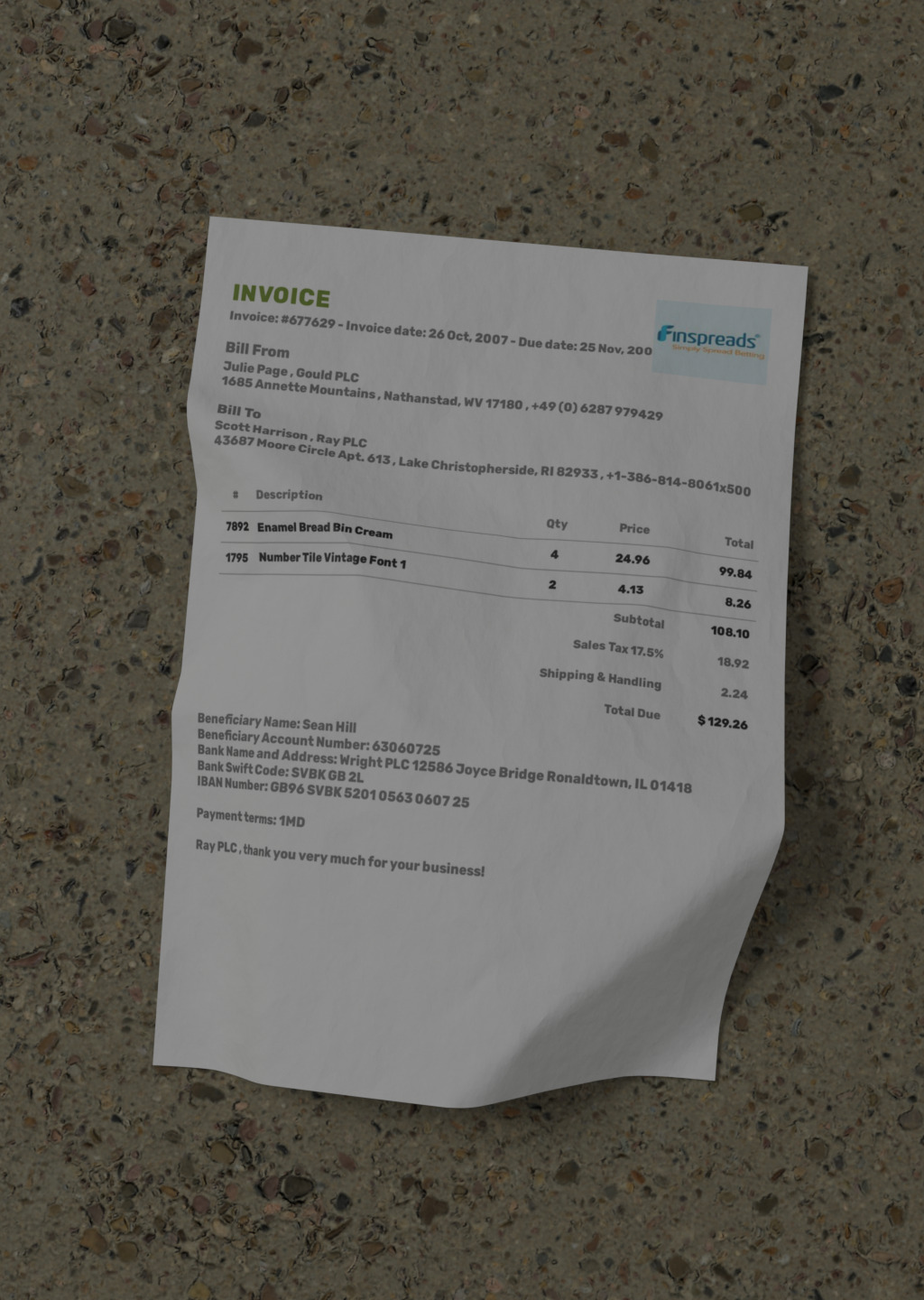}
    & \includegraphics[width=0.163\linewidth]{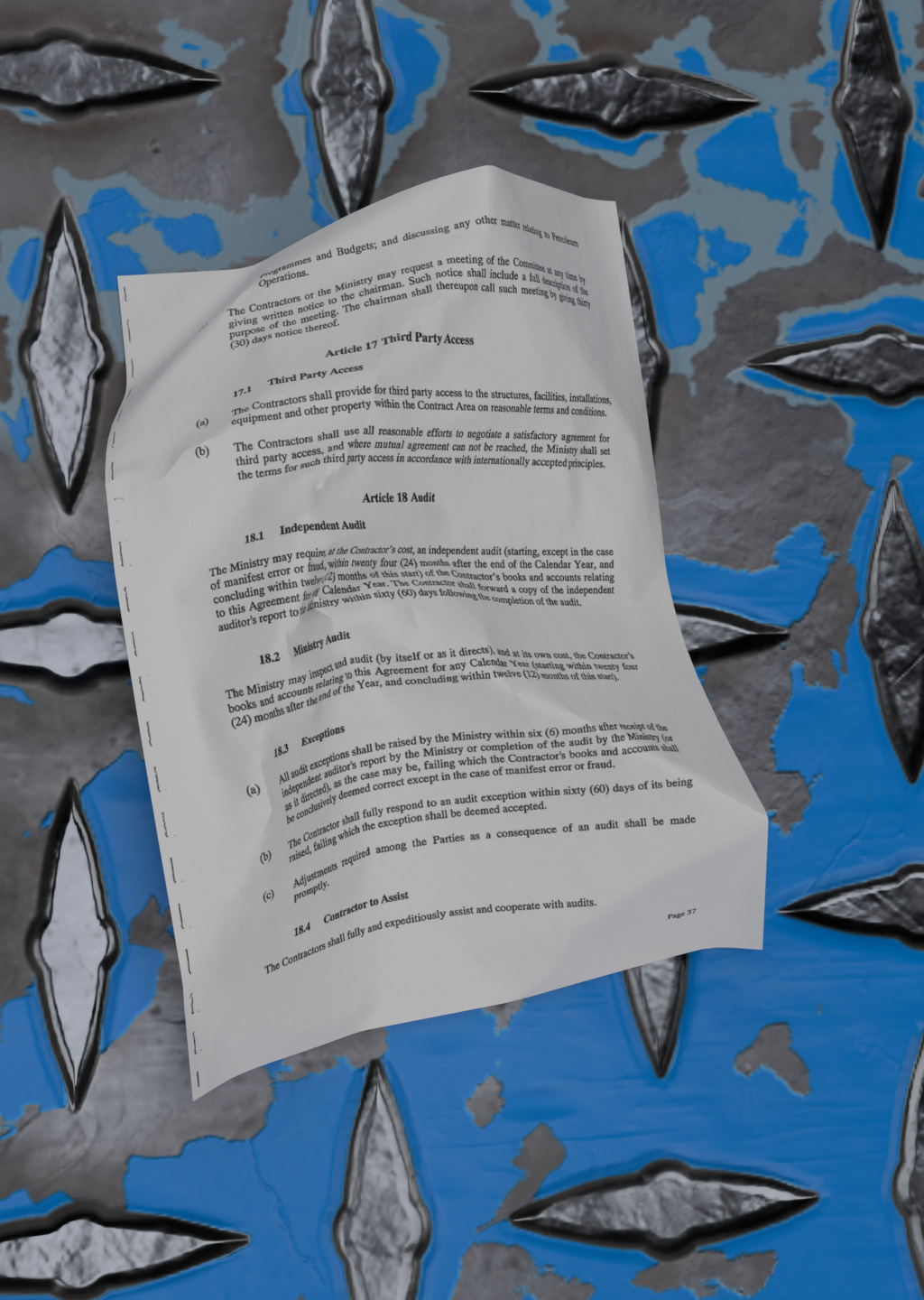}
    & \includegraphics[width=0.163\linewidth]{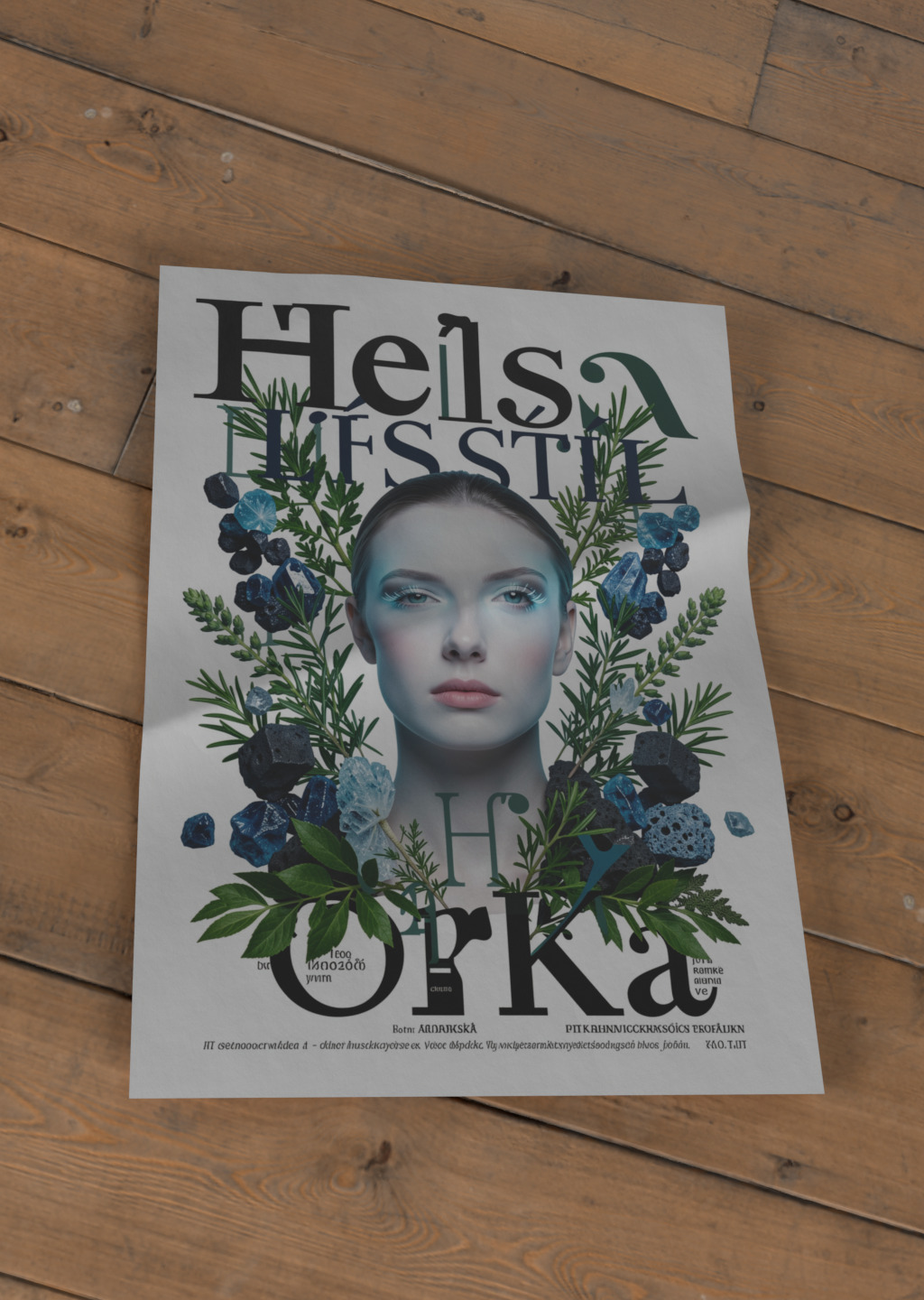}
    & \includegraphics[width=0.163\linewidth]{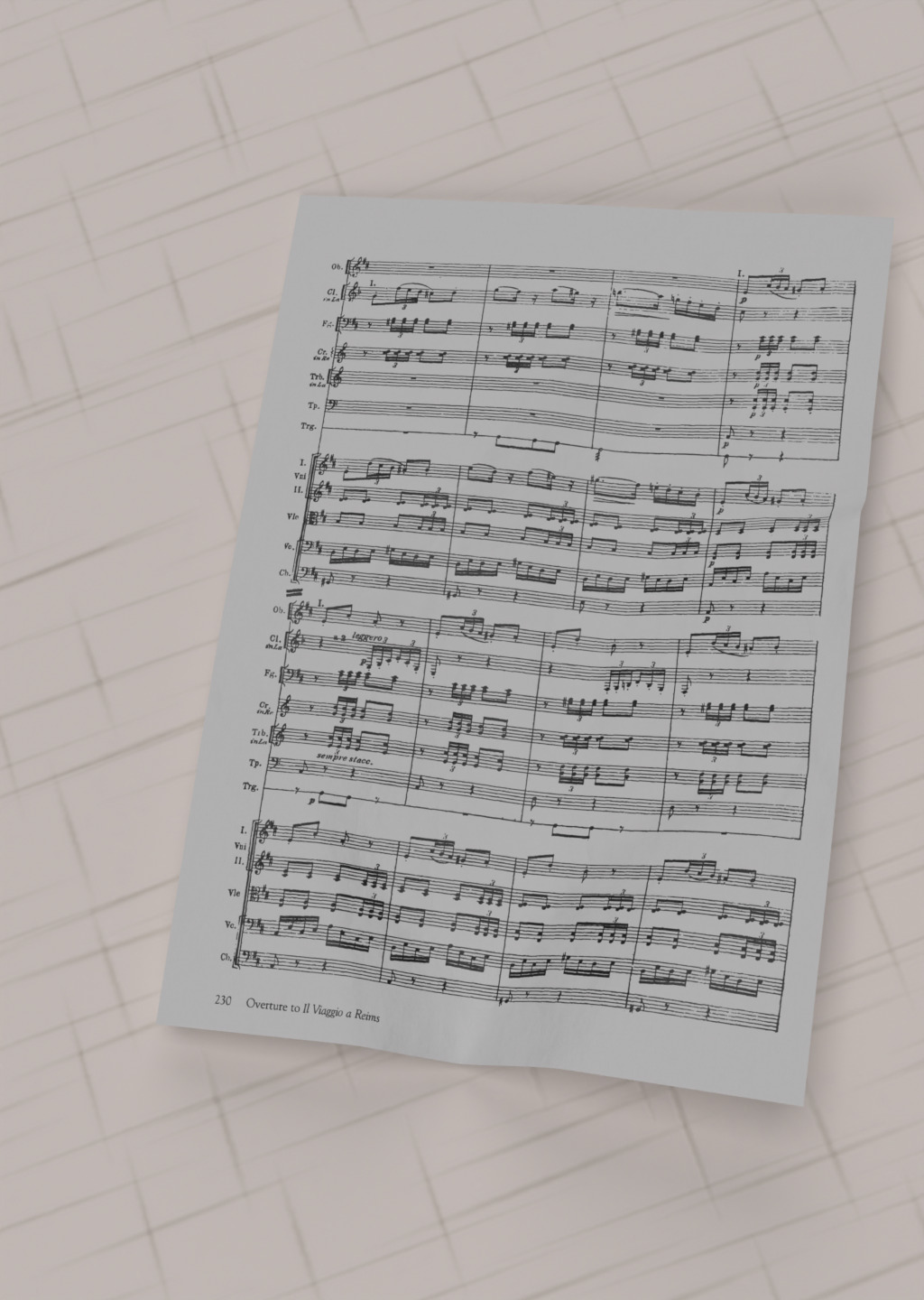}
    & \includegraphics[width=0.163\linewidth]{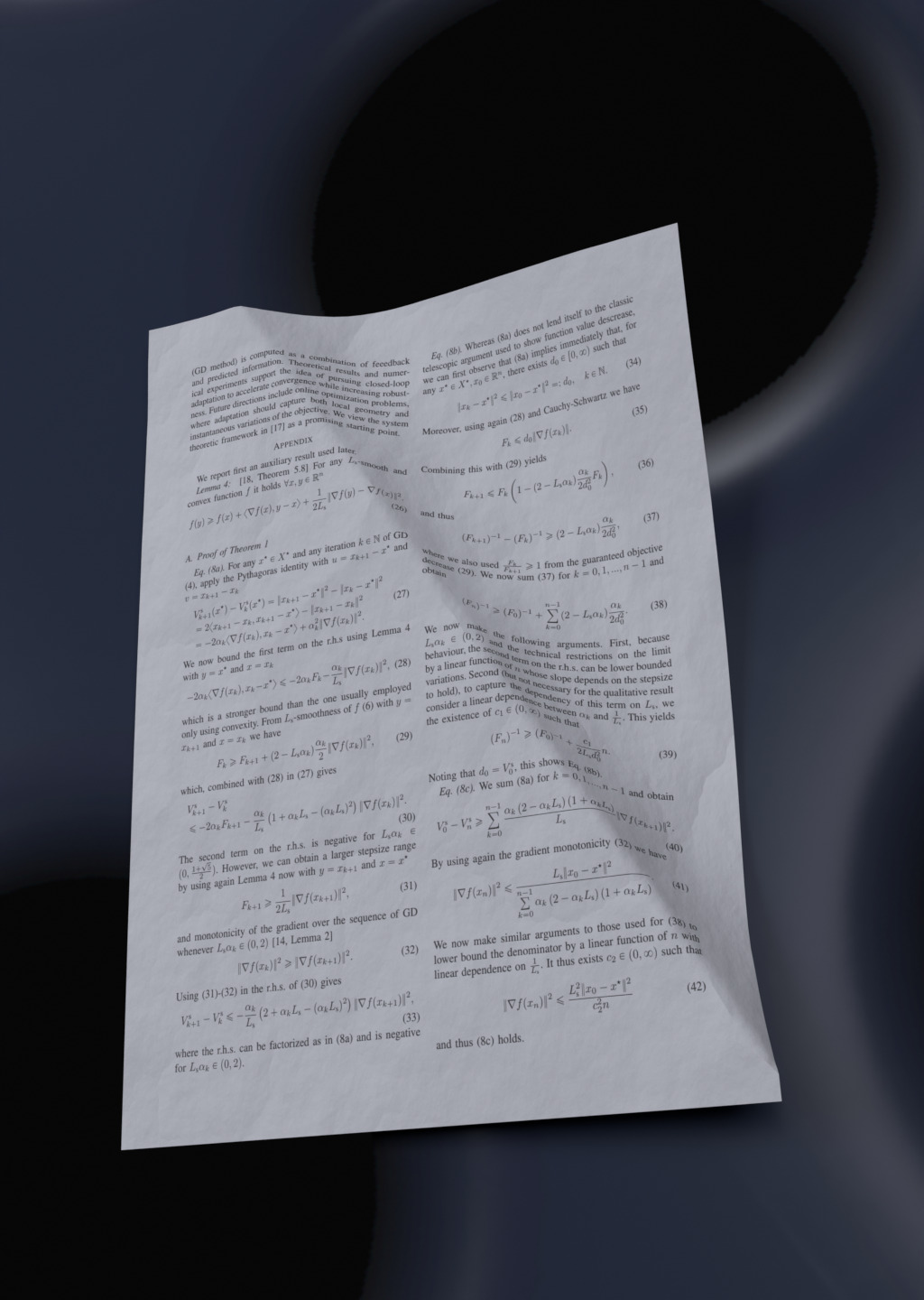}\\[1pt]
    \includegraphics[width=0.163\linewidth]{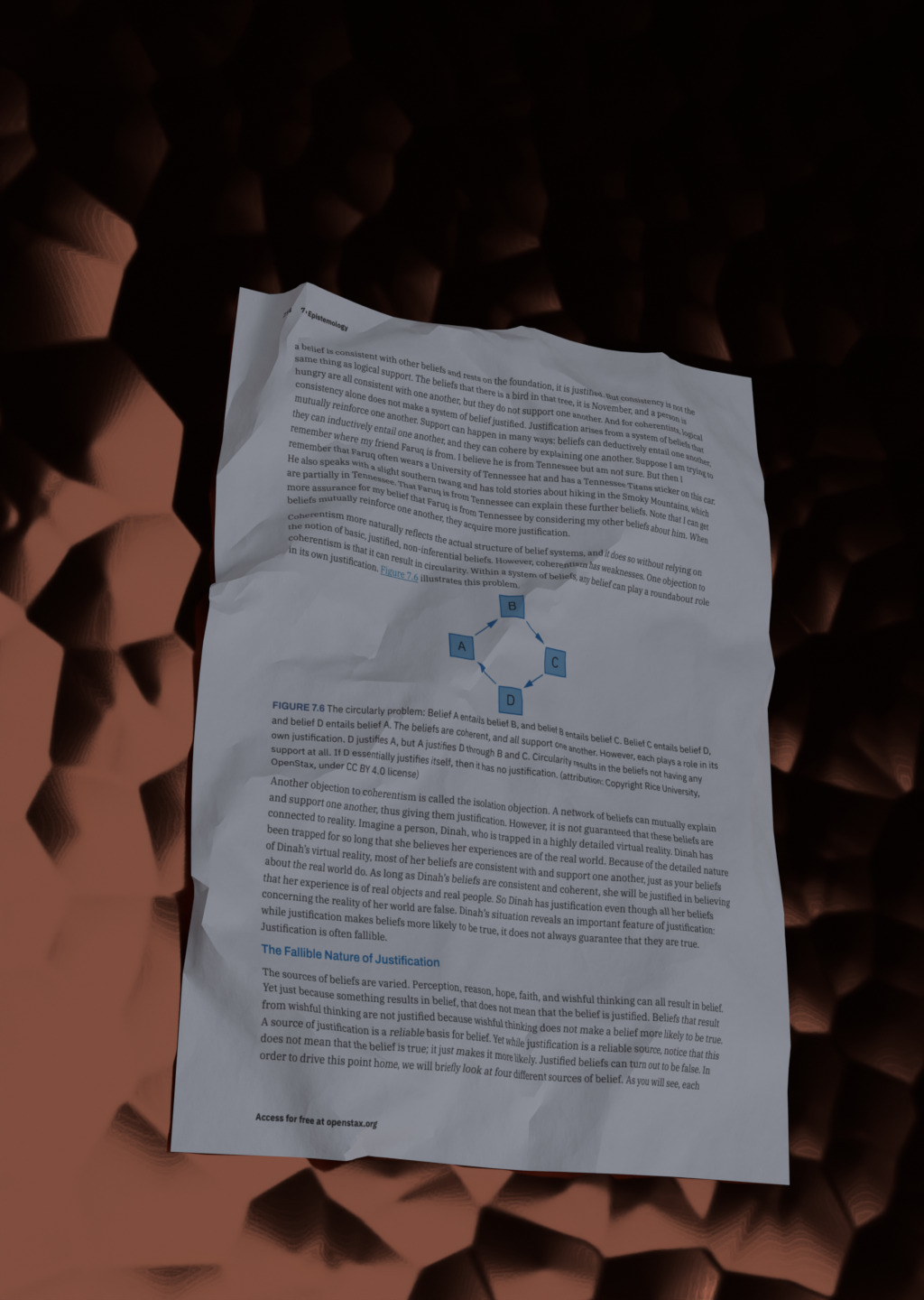}
    & \includegraphics[width=0.163\linewidth]{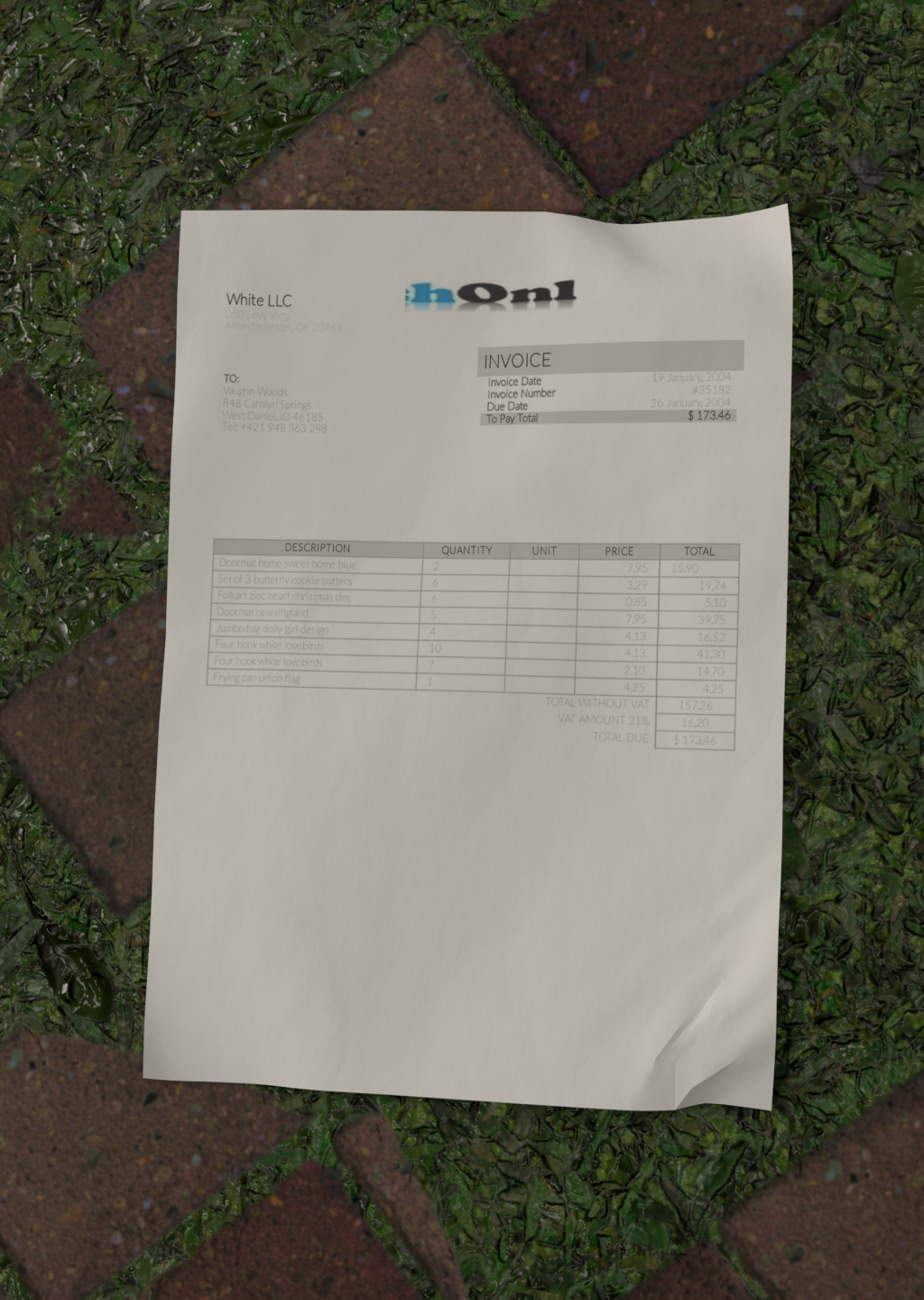}
    & \includegraphics[width=0.163\linewidth]{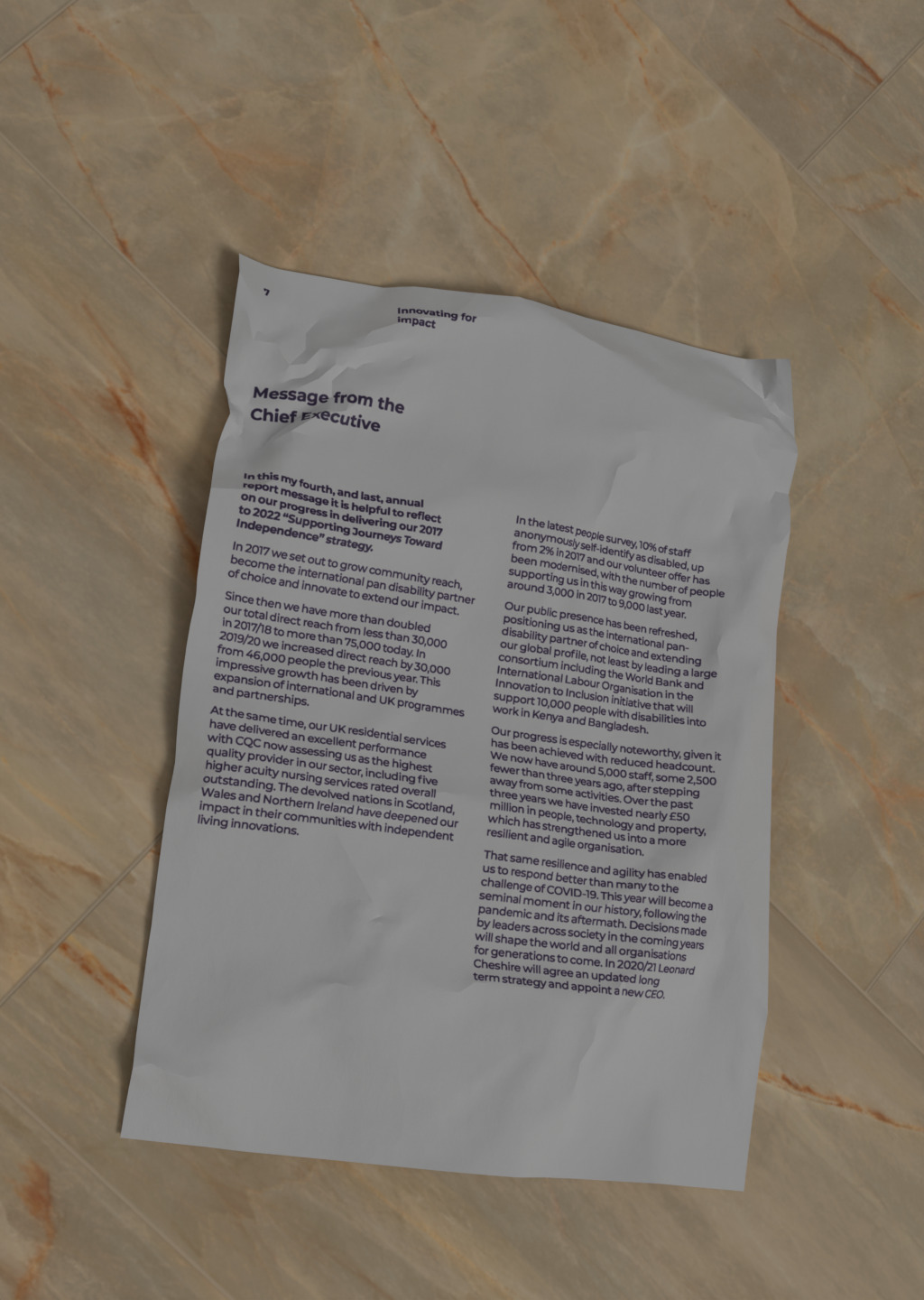}
    & \includegraphics[width=0.163\linewidth]{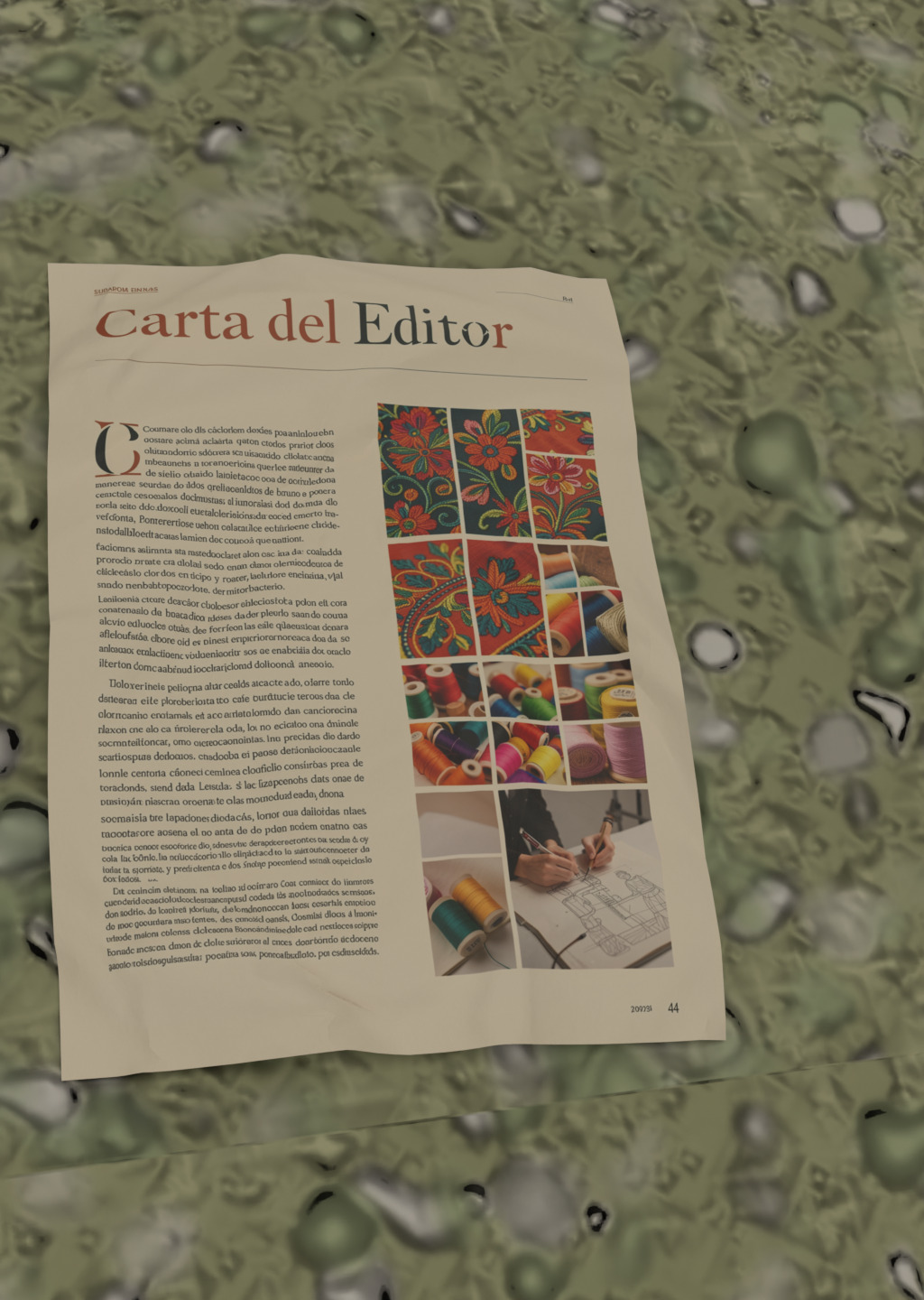}
    & \includegraphics[width=0.163\linewidth]{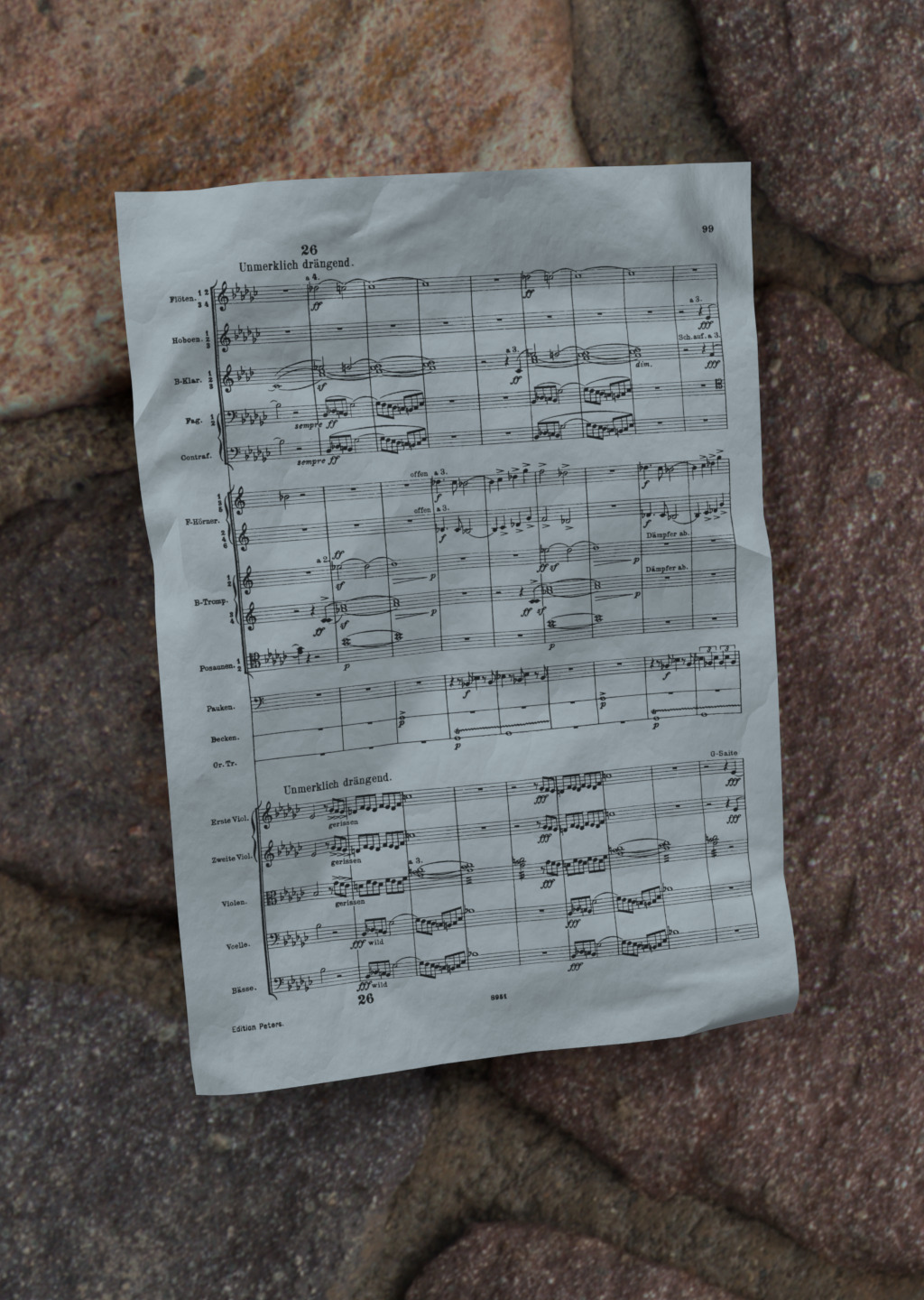}
    & \includegraphics[width=0.163\linewidth]{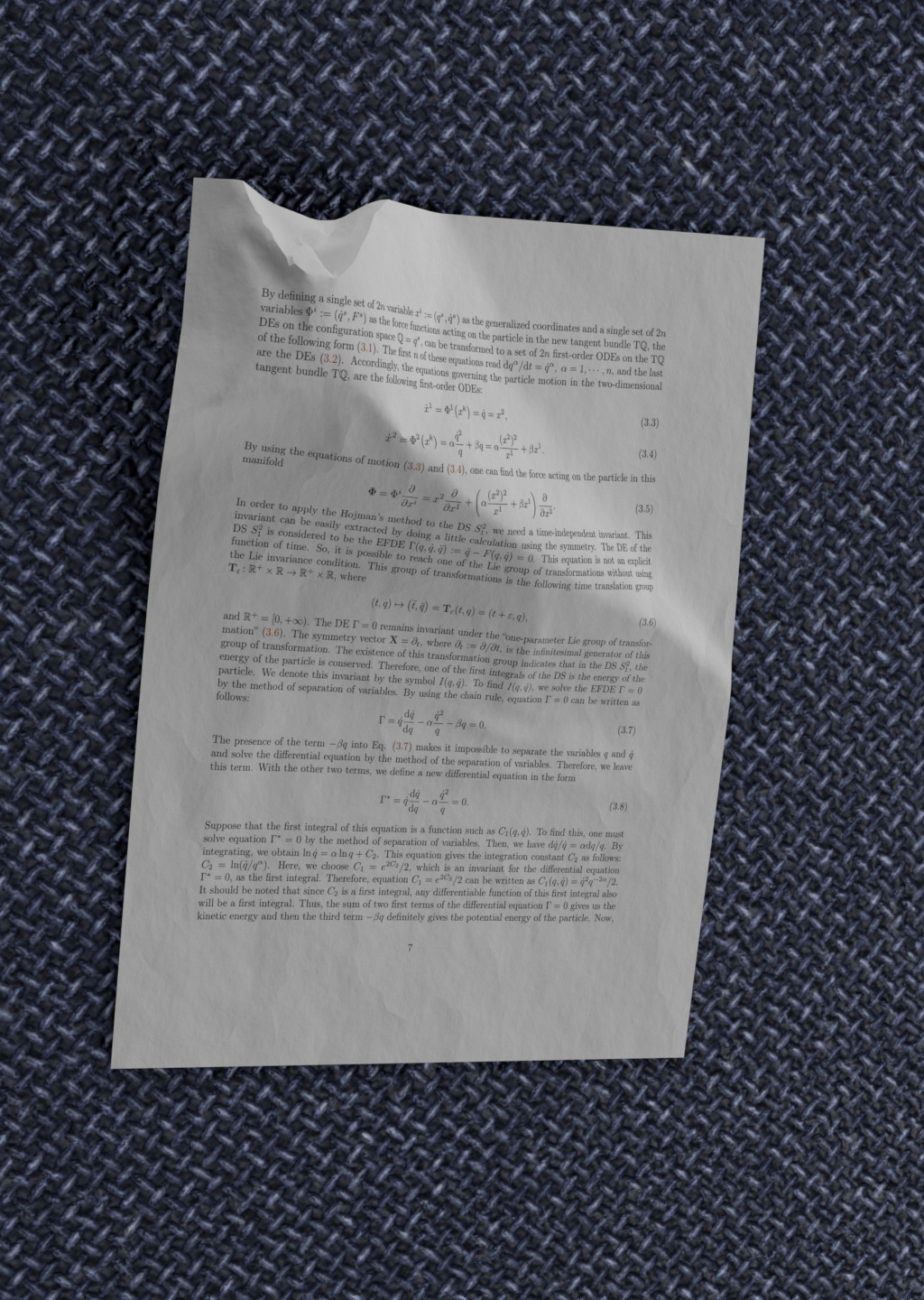}
    \end{tabular}
    \caption{Samples from our \ours{} dataset. \ours{} is composed of various document types (from left to right: book pages, invoices, legal documents, magazines, music sheets and scientific papers), with various deformations and lighting setups.}
    \label{fig:samples}
\end{figure}

\paragraph{Additional renders.} Alongside the main rendered image, we generate a comprehensive set of ground-truth annotations for each sample. To cleanly isolate the document, we first assign a black emissive material to the background plane. We capture the shading map by rendering the scene with identical parameters but removing the document's texture. The remaining annotations are generated by applying an unlit, emissive shader to the paper mesh and successively rendering the unshaded document albedo, surface normals, 3D world coordinates and UV maps. A sample from the dataset with all its annotations is presented in~\cref{fig:dataset_sample}. Modern networks are commonly trained to predict a dense backward mapping from the distorted input to the flat document. This mapping can be easily obtained by inverting the provided UV map.

\paragraph{}
Each of the 584,708 generated meshes is rendered twice. Some meshes are discarded because they create self-occlusions across all sampled viewpoints. Each document texture is reused no more than three times. Through this highly scalable pipeline, we generate the \ours{} dataset, comprising 1,000,000 training samples, 100,000 validation samples and over 38,000 test samples. A diverse selection of training images is presented in~\cref{fig:samples}.

The simulation and rendering scripts, all assets used to create \ours{} and the dataset itself are publicly available, allowing anyone to easily use and extend our work.

\let\originalfbox\fbox
\renewcommand\fbox{\fcolorbox{white}{white}}
\begin{table}[t]
    \centering
    \caption{Quantitative comparison of the unwarping performance of various methods on the DocUNet benchmark. The first row reports the metrics for the original warped documents. The last row presents our lightweight baseline model trained exclusively on our \ours{} dataset. \best{Gold}, \second{silver} and \third{bronze} backgrounds indicate the best, second-best and third-best scores, respectively.}
    \resizebox{0.60\linewidth}{!}{%
    \begin{tabular}{l r r r r r}
        \toprule
        Method & MS-SSIM $\uparrow$ & LD $`\downarrow$ & AD $\downarrow$ & CER $\downarrow$ & ED $\downarrow$ \\
        \midrule
        Warped image & \other{0.247} & \other{20.53} & \other{1.006} & \other{0.517} & \other{2029} \\\hdashline[0.5pt/1pt]
        DewarpNet \cite{das.etal2019} & \other{0.472} & \other{8.38} & \other{0.395} & \other{0.216} & \other{828} \\
        DisplacementFlow \cite{xie.etal2020} & \other{0.432} & \other{7.62} & \other{0.395} & \other{0.291} & \other{1207} \\
        DDControlPoints \cite{xie.etal2021} & \other{0.473} & \other{8.93} & \other{0.423} & \other{0.272} & \other{1102} \\
        DocTr \cite{feng.etal2021} & \other{0.509} & \other{7.78} & \other{0.366} & \other{0.180} & \other{713} \\
        PieceWise \cite{das.etal2021} & \other{0.490} & \other{8.65} & \other{0.430} & \other{0.247} & \other{977} \\
        FDRNet \cite{xue.etal2022} & \other{0.543} & \other{8.08} & \other{0.396} & \other{0.215} & \other{876} \\
        RDGR \cite{jiang.etal2022} & \other{0.495} & \other{8.50} & \other{0.432} & \other{0.170} & \other{725} \\
        Marior \cite{zhang.etal2022} & \other{0.476} & \other{7.37} & \other{0.404} & \other{0.198} & \other{788} \\
        PaperEdge \cite{ma.etal2022} & \other{0.472} & \other{7.98} & \other{0.367} & \other{0.189} & \other{751} \\
        DocGeoNet \cite{feng.etal2022} & \other{0.504} & \other{7.70} & \other{0.378} & \other{0.182} & \other{704} \\
        UVDoc \cite{verhoeven.etal2023} & \third{0.544} & \other{6.83} & \other{0.315} & \other{0.171} & \other{704} \\
        DocRES \cite{zhang.etal2024} & \other{0.464} & \other{9.40} & \other{0.470} & \other{0.231} & \other{890} \\
        DocScanner \cite{feng.etal2025} & \other{0.518} & \other{7.41} & \other{0.333} & \second{0.165} & \second{633} \\
        DvD \cite{zhang.etal2025} & \second{0.548} & \third{6.60} & \third{0.280} & \other{0.171} & \other{643} \\
        AADD \cite{wang.etal2025a} & \other{0.542} & \second{6.26} & \second{0.277} & \third{0.166} & \third{642} \\
        \hdashline[0.5pt/1pt]
        Ours & \best{0.558} & \best{5.98} & \best{0.252} & \best{0.161} & \best{628} \\
        \bottomrule
    \end{tabular}
    }
    \label{tab:results}
\end{table}
\let\fbox\originalfbox

\section{Experiments}
\label{sec:experiment}

To showcase the practical utility of \ours{}, we train a lightweight baseline model that jointly tackles document unwarping and illumination correction. The network is adapted from the one presented in UVDoc~\cite{verhoeven.etal2023}. We employ the same architecture, simply swapping the 3D grid prediction head with a shading prediction module. The shading head uses a U-Net style decoder that concatenates encoder features via skip connections at each resolution stage to obtain higher precision results, especially along document boundaries. As a result, our network outputs both a coarse unwarping map and a shading image, which can be applied independently to perform document unwarping and illumination correction. We use this architecture to serve as a baseline and evaluate the benefits of training a model on our \ours{} dataset over a combination of Doc3D and UVDoc. In addition, the lightweight architecture ensures that the model is fast at inference time for both document unwarping and illumination correction. 

\paragraph{Training details.} Similar to UVDoc~\cite{verhoeven.etal2023}, the predicted backward mapping is a coarse 45$\times$31 unwarping grid. The input of the model is a 512$\times$720 image of a warped document, tightly cropped around the document boundaries. We optimize the network using AdamW~\cite{kingma.ba2015, loshchilov.hutter2018} with a batch size of 16. The learning rate is initially set to 0.0001 with a weight decay of 0.0001, and evolves according to a cosine scheduler. We apply an $L_1$ loss to the coarse backward mapping, the shading map and the reconstructed image. The weight of the backward mapping loss is set to double that of the reconstruction and shading losses. We train the model for 38 epochs on \ours{} alone, using eight NVIDIA GeForce RTX 4090 GPUs, with each epoch taking approximately 1.5 hours.

\paragraph{Evaluation.} We assess our model's performance on the standard DocUNet benchmark~\cite{ma.etal2018}, comparing it against current state-of-the-art approaches. We evaluate these methods across multiple metrics. Image similarity is measured using multi-scale structural similarity (MS-SSIM), local distortion (LD) and aligned distortion (AD), while optical character recognition (OCR) accuracy is evaluated via the character error rate (CER) and edit distance (ED) metrics. Details about these metrics are provided in the supplementary material.

\paragraph{Results.} Quantitative results on the DocUNet benchmark are presented in~\cref{tab:results}. Note that the results of other SOTA methods were computed based on the results provided by their authors, or using the code they made available. Works that released neither their code nor their results on the DocUNet benchmark, such as Uni-DocDiff, are therefore not included in the comparison. As shown, our simple baseline network trained solely on our \ours{} dataset outperforms all state-of-the-art methods across the visual (MS-SSIM, LD and AD) and OCR (ED and CER) metrics, with improvements ranging from 1.8\% to 9\%. We provide qualitative visual comparisons of the unwarping against current state-of-the-art models in \cref{fig:visual_comparison} and in the supplementary material.

We also compare the results of our method after both document unwarping and illumination correction on the DocUNet benchmark to approaches capable of performing both tasks~\cite{feng.etal2021, zhang.etal2024}. Quantitative results are presented in~\cref{tab:results_illum}. Our simple baseline network outperforms all competing methods on the visual metrics (MS-SSIM and PSNR). While it does not achieve the best performance on the OCR metrics, it produces the most visually convincing results, better preserving the original colors of the document, as visible in the qualitative visual comparisons presented in the supplementary material.

\begin{figure}[t]
    \centering
	\small
	\setlength{\tabcolsep}{0.5pt}
	\renewcommand{\arraystretch}{0.0}  
    \setlength{\fboxsep}{0pt}
    \begin{tabular}{cccccc}
      \includegraphics[width=0.163\linewidth]{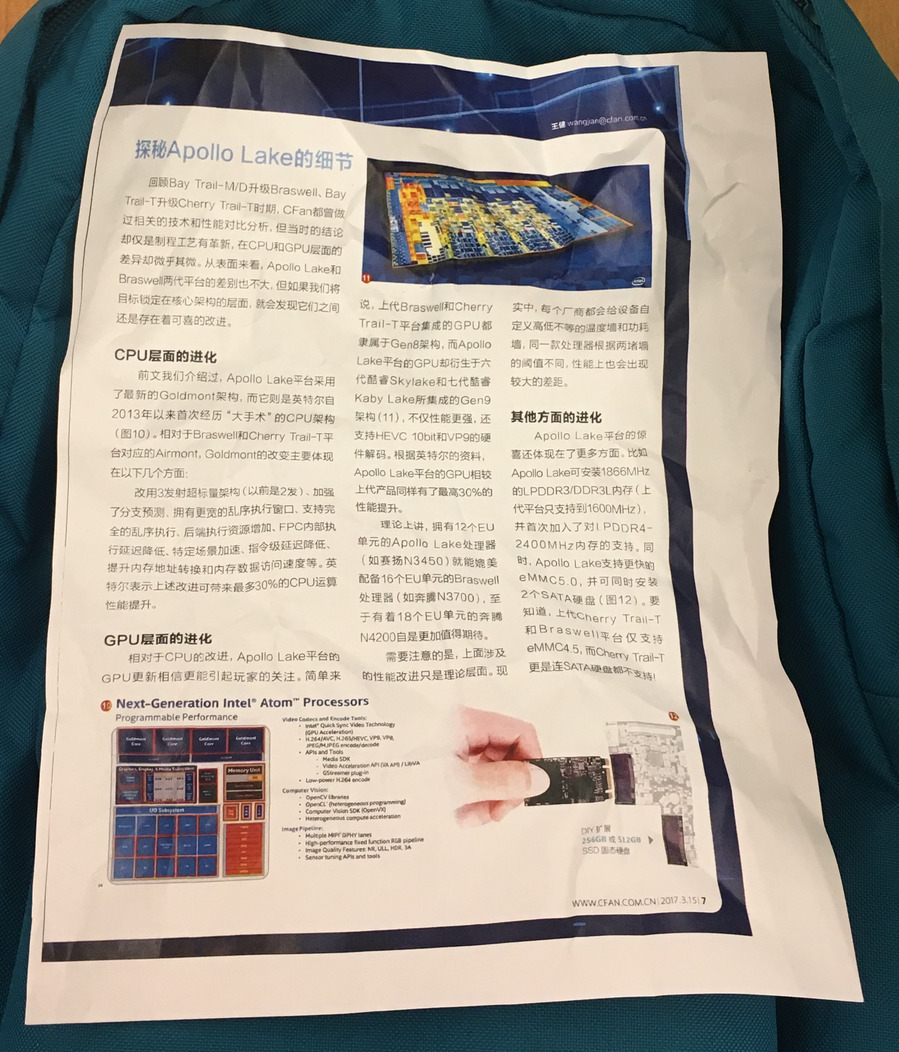}
    & \includegraphics[width=0.163\linewidth]{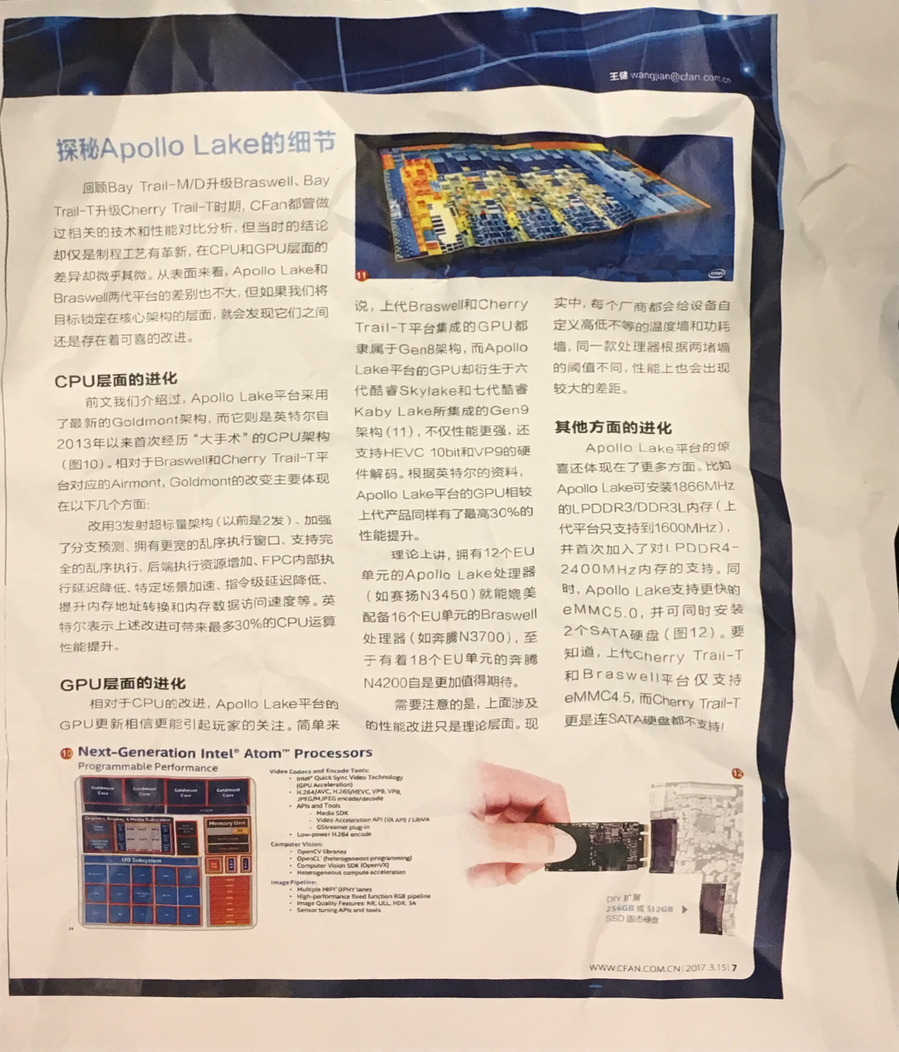}
    & \includegraphics[width=0.163\linewidth]{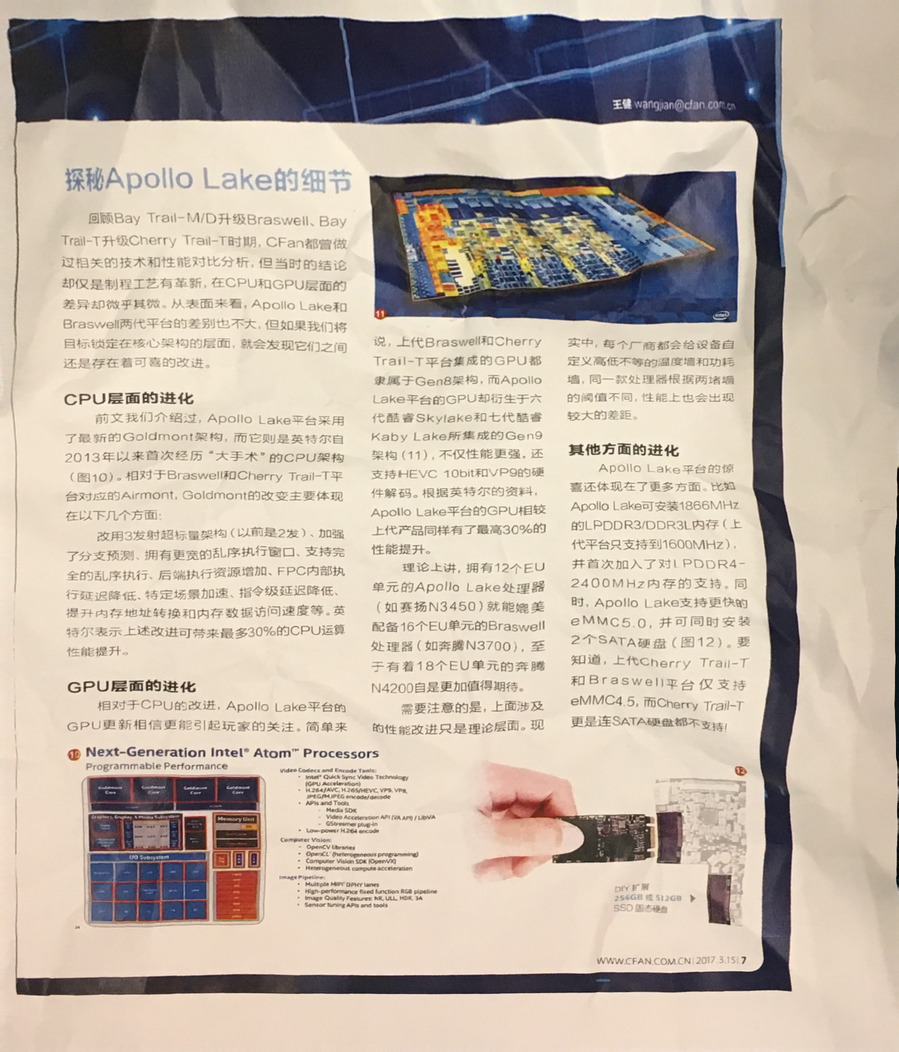}
    & \includegraphics[width=0.163\linewidth]{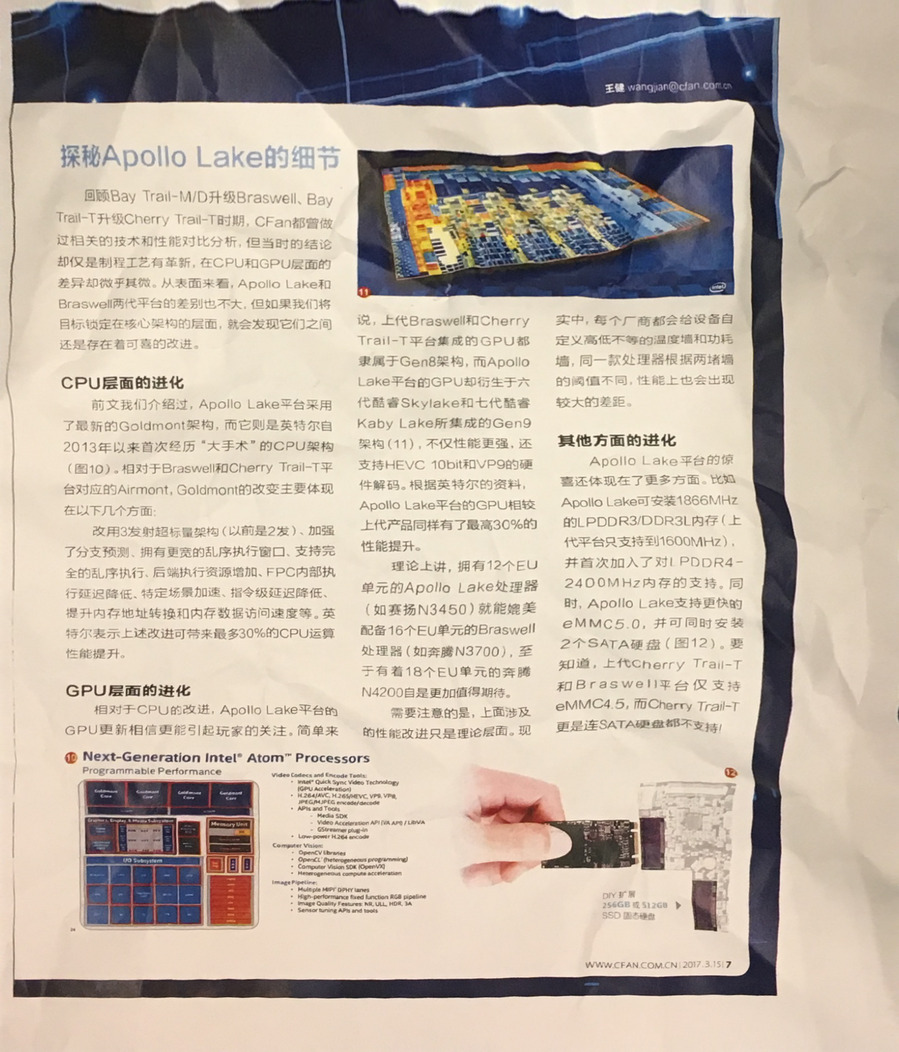}
    & \includegraphics[width=0.163\linewidth]{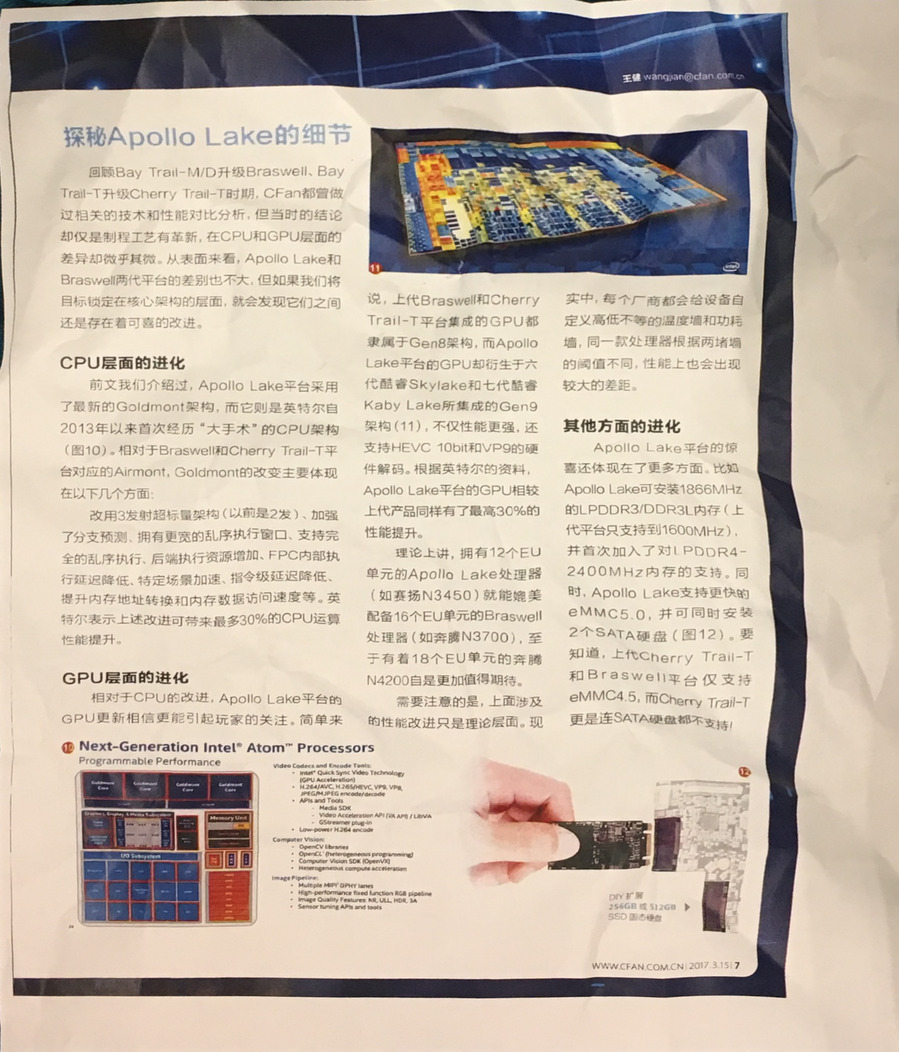}
    & \includegraphics[width=0.163\linewidth]{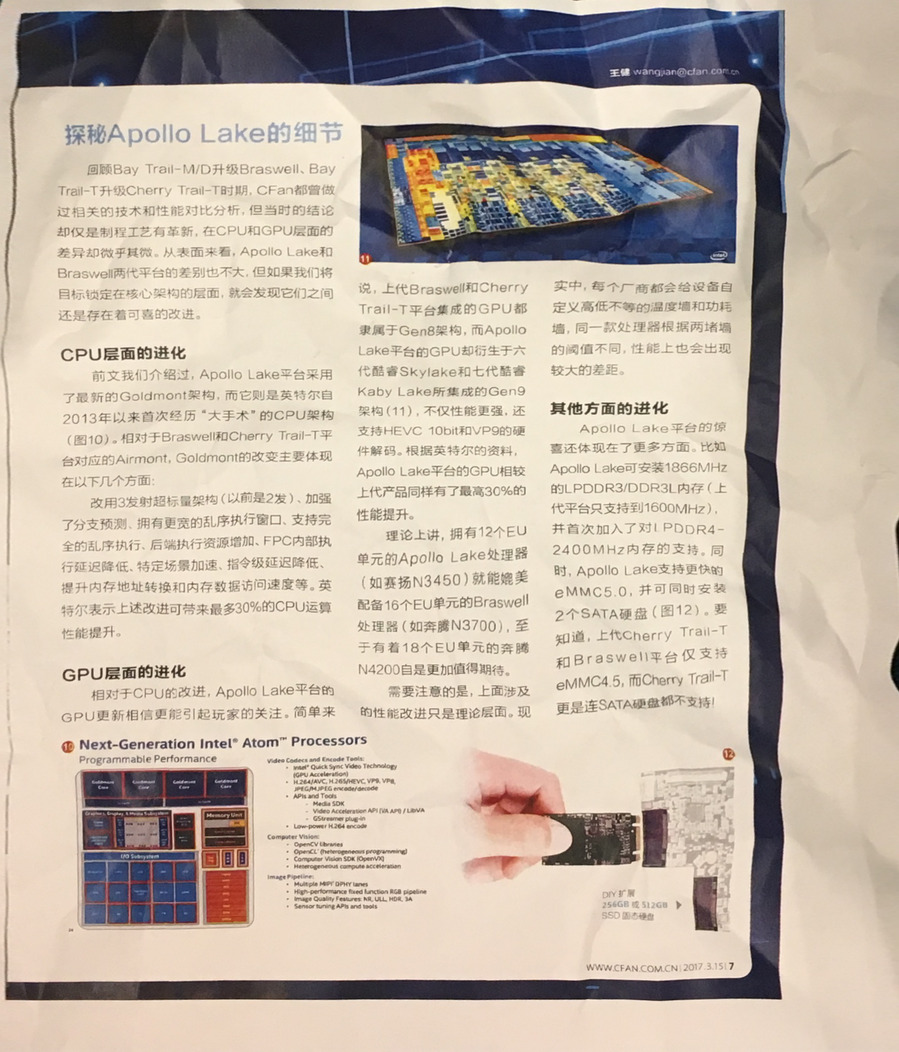}\\[1pt]
      \includegraphics[width=0.163\linewidth]{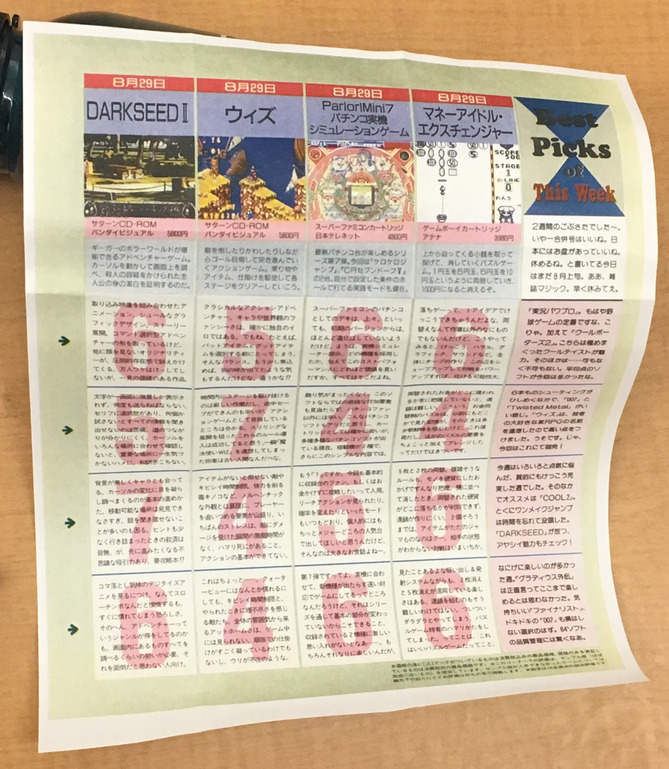}
    & \includegraphics[width=0.163\linewidth]{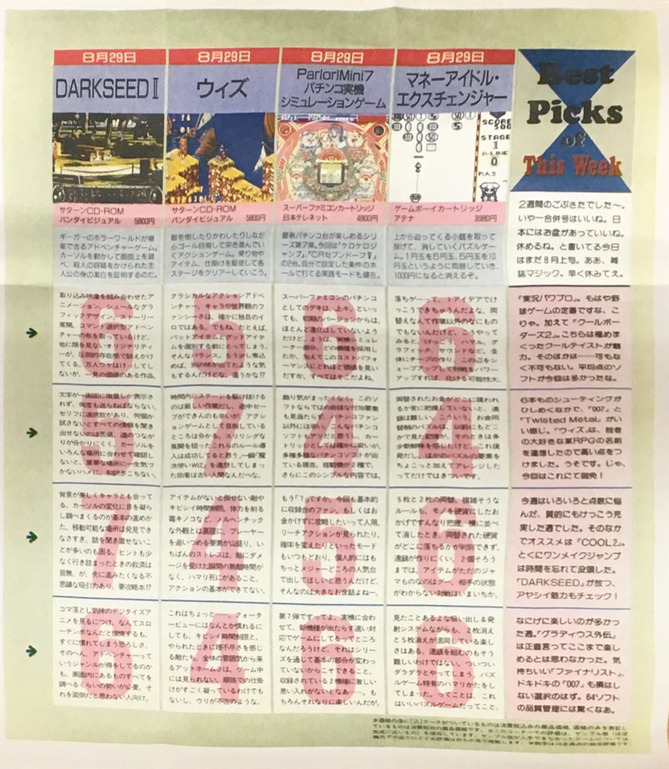}
    & \includegraphics[width=0.163\linewidth]{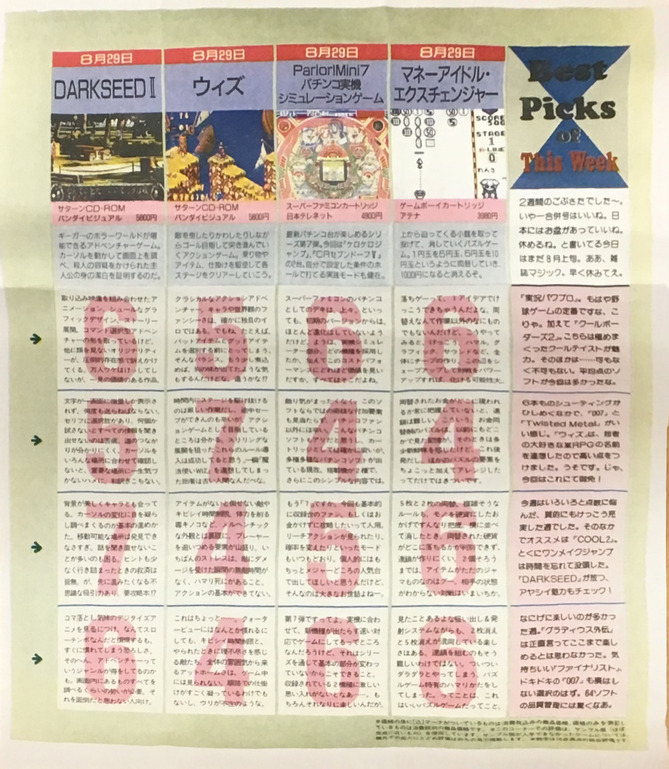}
    & \includegraphics[width=0.163\linewidth]{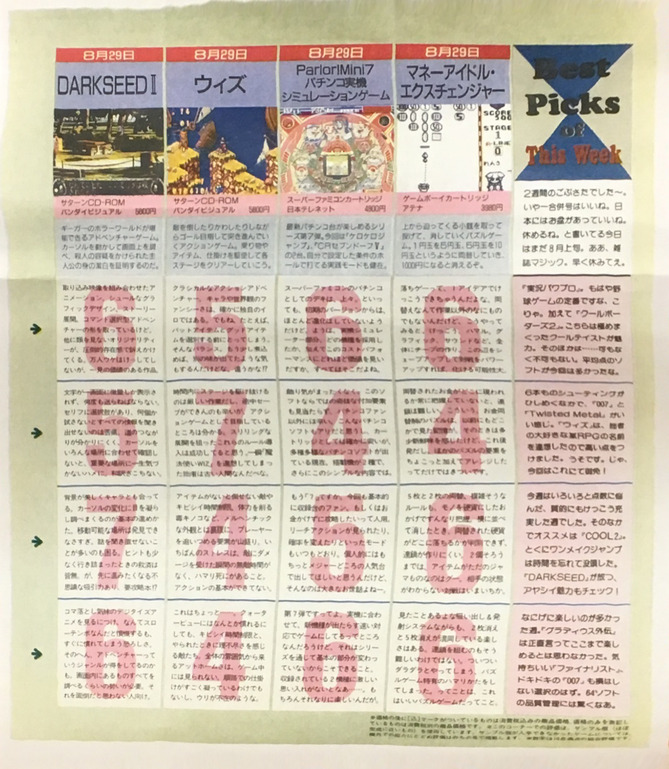}
    & \includegraphics[width=0.163\linewidth]{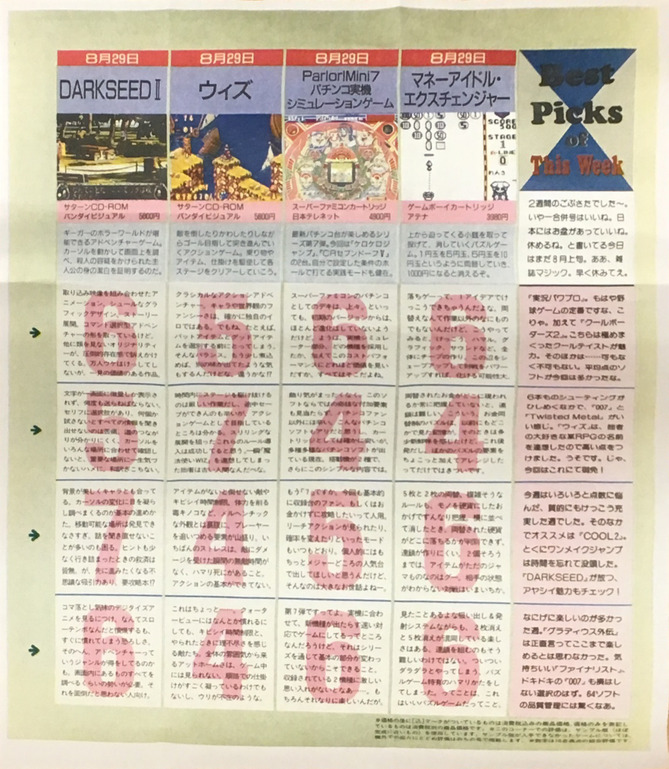}
    & \includegraphics[width=0.163\linewidth]{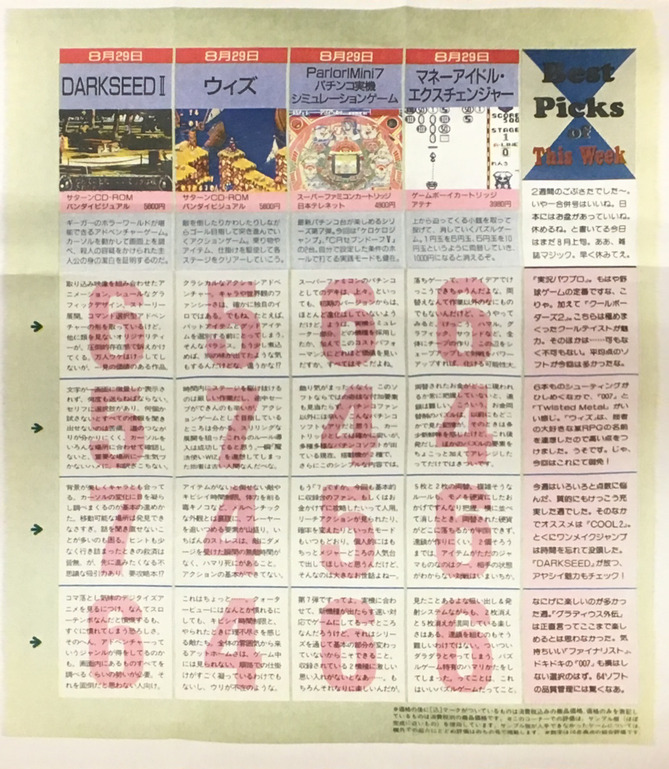}\\[1pt]
      \includegraphics[width=0.163\linewidth]{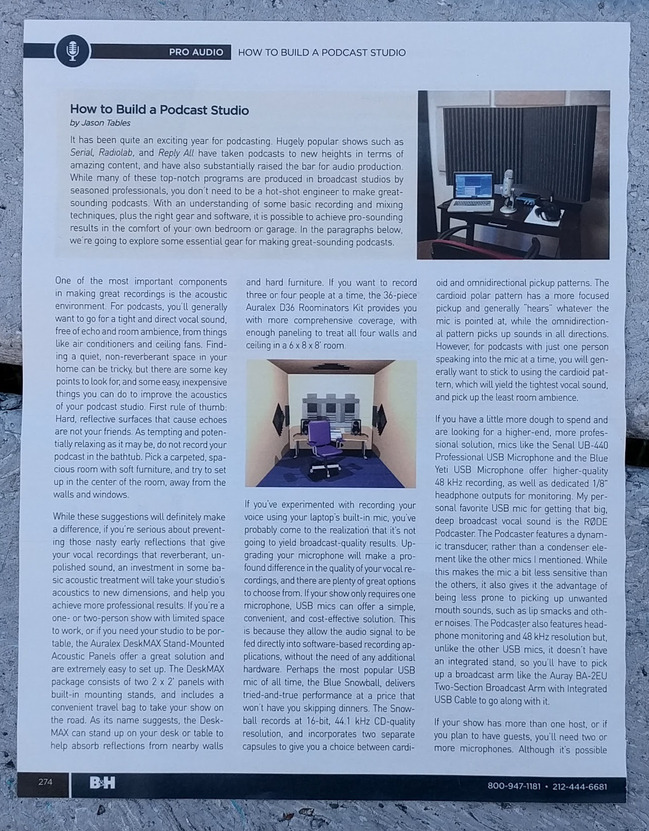}
    & \includegraphics[width=0.163\linewidth]{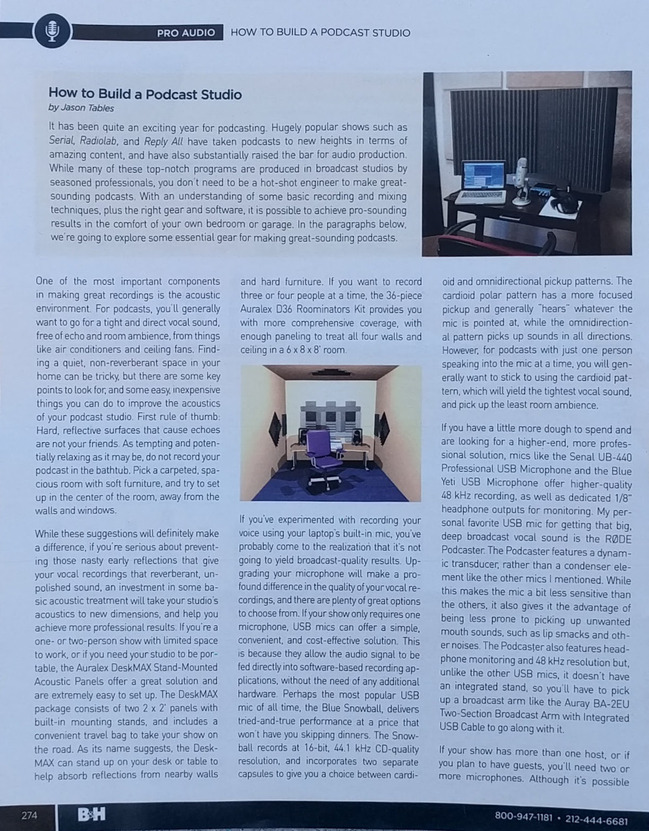}
    & \includegraphics[width=0.163\linewidth]{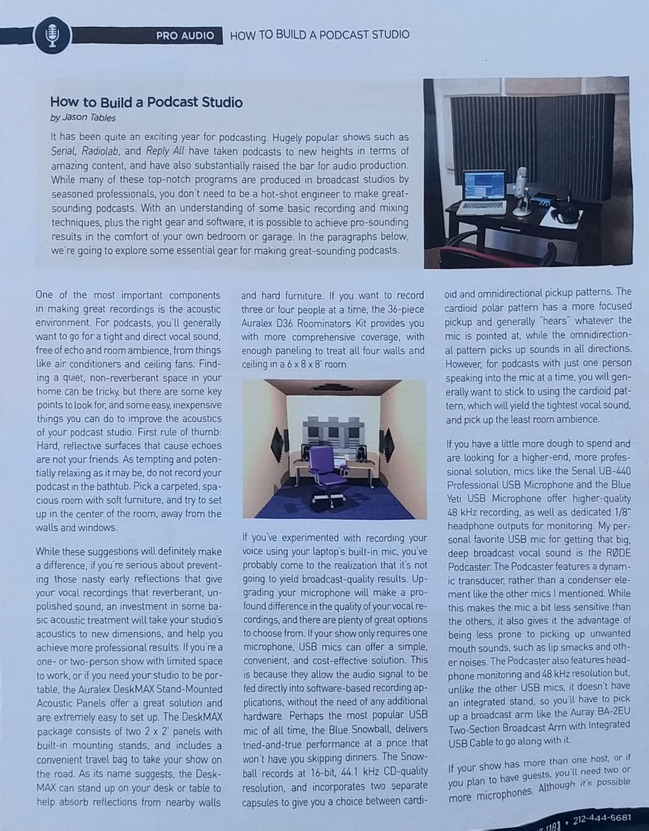}
    & \includegraphics[width=0.163\linewidth]{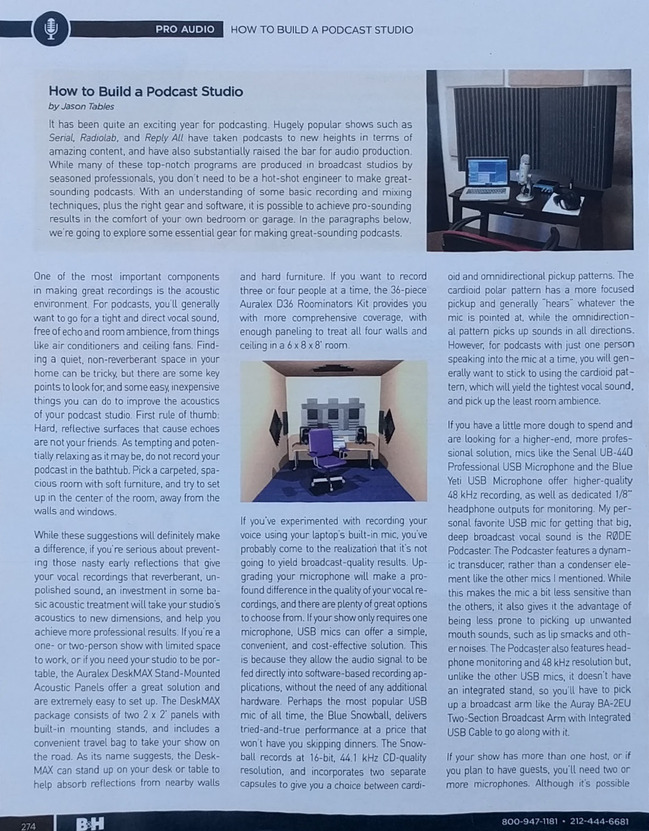}
    & \includegraphics[width=0.163\linewidth]{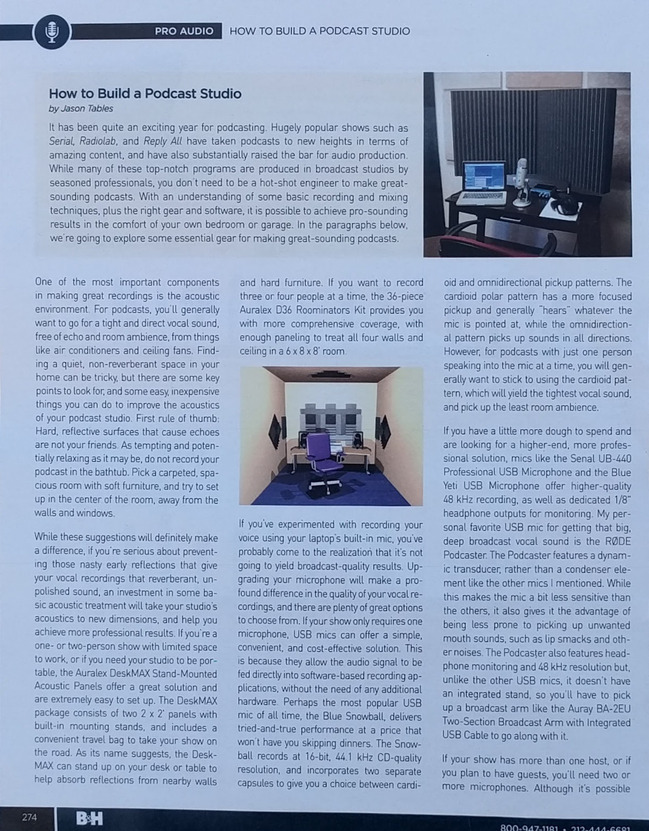}
    & \includegraphics[width=0.163\linewidth]{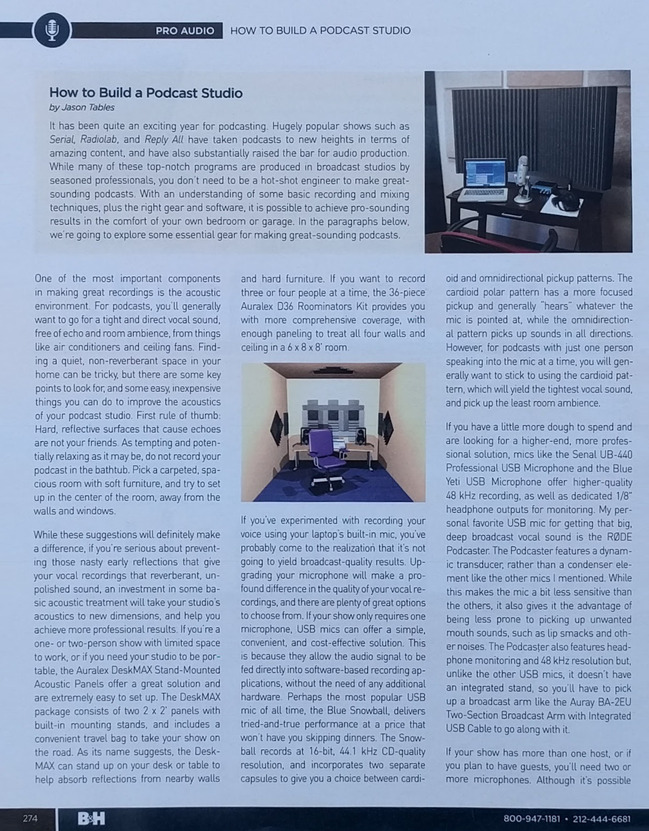}\\[5pt]
    \scriptsize Input &\scriptsize UVDoc \cite{verhoeven.etal2023} &\scriptsize DocScanner \cite{feng.etal2025} &\scriptsize DvD \cite{zhang.etal2025} &\scriptsize AADD \cite{wang.etal2025a} &\scriptsize Ours
    \end{tabular}
    \caption{Qualitative comparisons of unwarping-only results produced by various document unwarping methods on the DocUNet benchmark.}
    \label{fig:visual_comparison}
\end{figure}

\let\originalfbox\fbox
\renewcommand\fbox{\fcolorbox{white}{white}}
\begin{table}[t]
    \centering
    \caption{Quantitative comparison on both the unwarping and illumination correction tasks of various methods on the DocUNet benchmark. The first row reports the metrics for the original warped documents. The last row presents our lightweight baseline model trained exclusively on our \ours{} dataset. \best{Gold} and\second{silver} backgrounds indicate the best and second-best scores, respectively.}
    \resizebox{0.47\linewidth}{!}{%
    \begin{tabular}{l r r r r}
        \toprule
        Method & MS-SSIM $\uparrow$ & PSNR $\uparrow$ & CER $\downarrow$ & ED $\downarrow$ \\
        \midrule
        Warped Image & \other{0.247} & \other{7.92} & \other{0.517} & \other{2029} \\\hdashline[0.5pt/1pt]
        DocTr \cite{feng.etal2021} & \second{0.496} & \other{10.79} & \best{0.126} & \best{491} \\
        DocRES \cite{zhang.etal2024} & \other{0.473} & \second{11.68} & \other{0.173} & \other{637} \\
        \hdashline[0.5pt/1pt]
        Ours & \best{0.584} & \best{12.68} & \second{0.149} & \second{601} \\
        \bottomrule
    \end{tabular}
    }
    \label{tab:results_illum}
\end{table}
\let\fbox\originalfbox

\let\originalfbox\fbox
\renewcommand\fbox{\fcolorbox{white}{white}}
\begin{table}[t]
    \centering
    \caption{Quantitative comparison of the unwarping performance of variants of our methods on the DocUNet benchmark. The \original{gray row} is our original experiment.
    }
    \begin{tabular}{r c c r r r r r}
        \toprule
        \multicolumn{3}{l}{Method} & MS-SSIM $\uparrow$ & LD $\downarrow$ & AD $\downarrow$ & CER $\downarrow$ & ED $\downarrow$ \\
        \midrule
        \#samples & Input & Output & & & & & \\
        100k& $512 \times 720$ & $31 \times 45$ & \other{0.532} & \other{6.89} & \other{0.323} & \other{0.194} & \other{704} \\
        500k& $512 \times 720$ & $31 \times 45$ & \other{0.546} & \other{6.52} & \other{0.278} & \other{0.168} & \other{654} \\
        \rowcolor{original}
        1M & $512 \times 720$ &$ 31 \times 45$ & \original{0.558} & \original{5.98} & \original{0.252} & \original{0.161} & \original{628} \\
        1M & $1024 \times 1440$ & $121 \times 177$ & \other{0.570} & \other{5.51} & \other{0.219} & \other{0.164} & \other{651}\\
        \bottomrule
    \end{tabular}
    \label{tab:ablation}
\end{table}
\let\fbox\originalfbox

\subsection{Ablation studies}

\paragraph{Training with fewer samples.}
We trained our baseline model on subsets of 100K and 500K samples from our \ours{} dataset (see \cref{tab:ablation}). These models were trained for fewer epochs due to time constraints. Even when trained with fewer samples, our simple model achieves similar performance to most SOTA methods, demonstrating the high quality of the dataset. Our model trained on the full dataset still performs best, highlighting the value of \ours{}'s scale.

\paragraph{Training with higher resolution.}
We intentionally use a lightweight baseline model to prove that our \ours{} dataset yields SOTA results without architectural tricks. To demonstrate that our high-resolution, large-scale data can push performance further, we modified this simple model by increasing the input/output resolution. As shown in \cref{tab:ablation} (last row), this increased capacity yields massive gains in visual metrics, with the AD score improving by more than 20\% over SOTA.

\section{Discussion}
\label{sec:discussion}

We presented \ours{}, a large-scale synthetic dataset for document unwarping and illumination correction. Comprising 1,000,000 training samples, \ours{} represents the first dataset of this size for these tasks. High-quality samples are obtained through the combination of physically simulated geometries and high-resolution path-traced rendering, mitigating the sim-to-real gap. We demonstrate a practical application of this dataset by training a lightweight baseline model for document unwarping and illumination correction. 
Our dataset is available at https://igl.ethz.ch/projects/SyntheticDoc/ and the code used to generate it at https://github.com/tanguymagne/SyntheticDoc.

\paragraph{Limitations.} Due to the scale and high-quality rendering, the final dataset is very large in size. The training set alone, containing only the rendered images, albedos, shadow maps and UV maps, is over \unit[5]{TB}. This makes model training computationally demanding, causing us to opt for a relatively lightweight architecture. We believe that with less constrained hardware capacities, a very large model could truly benefit from the size of this dataset to push performance even further. 
Furthermore, since we focused on the most common real-world capture scenarios, the current dataset does not cover extreme viewpoints or diverse camera types. However, our setup can be easily extended to cover such cases.

\paragraph{Future work.} Built entirely upon highly scalable steps, our generation pipeline can be readily reproduced to further expand the dataset with more document textures or new physical simulation scenarios for the deformed document meshes, including more complex fold scenarios (explicit Z-fold or tri-fold simulations) or other paper formats. In addition, the procedural nature of our approach makes it relatively straightforward to extract additional ground-truth annotations such as OCR transcripts or document layouts, extending the utility of \ours{} to other downstream document analysis tasks. Furthermore, our generation pipeline could be used to create a multi-view dataset for document processing.

\paragraph{Ethical considerations.} When gathering the assets required to produce our dataset, we ensured that they were all available under permissive licenses. 
While utilizing text-to-image models to generate magazine pages can raise concerns regarding data provenance and copyright infringement, we mitigate this issue by employing strictly generic prompts that do not attempt to replicate specific real-world magazines.

Our data generation pipeline is computationally efficient. Simulating a mesh takes at most 10 minutes on a CPU-only machine, while rendering each sample takes approximately 5 seconds on a machine equipped with a consumer-grade RTX 4090 GPU. In addition, the cluster we use to generate our dataset and train our model is carbon neutral and powered entirely by renewable energy, minimizing the environmental footprint of our work.

\section*{Acknowledgements}

We thank the anonymous reviewers for their insightful feedback and constructive suggestions. We are also grateful to Danielle Luterbacher for her help in managing the hardware required to create and store a dataset of this size.

\clearpage  


%
%
\bibliographystyle{splncs04}
\bibliography{references}

@misc{arxiv,
  author  = {{arXiv}},
  title   = {{arXiv}},
  year    = {2026},
  url     = {https://arxiv.org/},
  urldate = {2026-02-28}
}

@misc{blender,
  author  = {{Blender Foundation}},
  title   = {Blender 4.5},
  year    = {2026},
  url     = {https://www.blender.org/},
  urldate = {2026-02-28}
}

@article{brown.etal2006,
  author   = {Brown, M.S. and Tsoi, Y.-C.},
  journal  = {IEEE Transactions on Image Processing},
  title    = {Geometric and shading correction for images of printed materials using boundary},
  year     = {2006},
  volume   = {15},
  number   = {6},
  pages    = {1544-1554},
  doi      = {10.1109/TIP.2006.871082}
}

@inproceedings{brown.seales2001,
  author    = {Brown, M.S. and Seales, W.B.},
  booktitle = {Proceedings Eighth IEEE International Conference on Computer Vision. ICCV 2001},
  title     = {Document restoration using 3D shape: a general deskewing algorithm for arbitrarily warped documents},
  year      = {2001},
  volume    = {2},
  number    = {},
  pages     = {367-374 vol.2},
  doi       = {10.1109/ICCV.2001.937649}
}

@article{brown.seales2004,
  author   = {Brown, M.S. and Seales, W.B.},
  journal  = {IEEE Transactions on Pattern Analysis and Machine Intelligence},
  title    = {Image restoration of arbitrarily warped documents},
  year     = {2004},
  volume   = {26},
  number   = {10},
  pages    = {1295-1306},
  doi      = {10.1109/TPAMI.2004.87}
}

@inproceedings{cosyn-400k,
  title     = {Scaling Text-Rich Image Understanding via Code-Guided Synthetic Multimodal Data Generation},
  author    = {Yang, Yue  and
               Patel, Ajay  and
               Deitke, Matt  and
               Gupta, Tanmay  and
               Weihs, Luca  and
               Head, Andrew  and
               Yatskar, Mark  and
               Callison-Burch, Chris  and
               Krishna, Ranjay  and
               Kembhavi, Aniruddha  and
               Clark, Christopher},
  editor    = {Che, Wanxiang  and
               Nabende, Joyce  and
               Shutova, Ekaterina  and
               Pilehvar, Mohammad Taher},
  booktitle = {Proceedings of the 63rd Annual Meeting of the Association for Computational Linguistics (Volume 1: Long Papers)},
  month     = jul,
  year      = {2025},
  address   = {Vienna, Austria},
  publisher = {Association for Computational Linguistics},
  doi       = {10.18653/v1/2025.acl-long.855},
  pages     = {17486--17505},
  isbn      = {979-8-89176-251-0}
}

@inproceedings{das.etal2019,
  author    = {Das, Sagnik and Ma, Ke and Shu, Zhixin and Samaras, Dimitris and Shilkrot, Roy},
  title     = {DewarpNet: Single-Image Document Unwarping With Stacked 3D and 2D Regression Networks},
  booktitle = {Proceedings of the IEEE/CVF International Conference on Computer Vision (ICCV)},
  month     = {October},
  year      = {2019}
}

@inproceedings{das.etal2020,
  title     = {Intrinsic Decomposition of Document Images In-the-Wild},
  author    = {Das, Sagnik and Sial, Hassan and Ma, Ke and Baldrich, Ram{\'o}n and Vanrell, Maria and Samaras, Dimitris},
  booktitle = {Proceedings of the 31st British Machine Vision Conference (BMVC)},
  year      = {2020},
  doi       = {10.5244/C.34.188}
}

@inproceedings{das.etal2021,
  author    = {Das, Sagnik and Singh, Kunwar Yashraj and Wu, Jon and Bas, Erhan and Mahadevan, Vijay and Bhotika, Rahul and Samaras, Dimitris},
  title     = {End-to-End Piece-Wise Unwarping of Document Images},
  booktitle = {Proceedings of the IEEE/CVF International Conference on Computer Vision (ICCV)},
  month     = {October},
  year      = {2021},
  pages     = {4268-4277}
}

@article{deschaintre.etal2018,
  author     = {Deschaintre, Valentin and Aittala, Miika and Durand, Fredo and Drettakis, George and Bousseau, Adrien},
  title      = {Single-image SVBRDF capture with a rendering-aware deep network},
  year       = {2018},
  issue_date = {August 2018},
  publisher  = {Association for Computing Machinery},
  address    = {New York, NY, USA},
  volume     = {37},
  number     = {4},
  issn       = {0730-0301},
  doi        = {10.1145/3197517.3201378},
  journal    = {ACM Trans. Graph.},
  month      = jul,
  articleno  = {128},
  numpages   = {15}
}

@misc{doab,
  author  = {{DOAB}},
  title   = {Directory of Open Access Books},
  year    = {2026},
  url     = {https://www.doabooks.org/},
  urldate = {2026-02-28}
}

@inproceedings{feng.etal2021,
  author    = {Feng, Hao and Wang, Yuechen and Zhou, Wengang and Deng, Jiajun and Li, Houqiang},
  title     = {DocTr: Document Image Transformer for Geometric Unwarping and Illumination Correction},
  year      = {2021},
  isbn      = {9781450386517},
  publisher = {Association for Computing Machinery},
  address   = {New York, NY, USA},
  doi       = {10.1145/3474085.3475388},
  booktitle = {Proceedings of the 29th ACM International Conference on Multimedia},
  pages     = {273–281},
  numpages  = {9},
  location  = {Virtual Event, China},
  series    = {MM '21}
}

@inproceedings{feng.etal2022,
  author    = {Feng, Hao
               and Zhou, Wengang
               and Deng, Jiajun
               and Wang, Yuechen
               and Li, Houqiang},
  editor    = {Avidan, Shai
               and Brostow, Gabriel
               and Ciss{\'e}, Moustapha
               and Farinella, Giovanni Maria
               and Hassner, Tal},
  title     = {Geometric Representation Learning for Document Image Rectification},
  booktitle = {Computer Vision -- ECCV 2022},
  year      = {2022},
  publisher = {Springer Nature Switzerland},
  address   = {Cham},
  pages     = {475--492},
  isbn      = {978-3-031-19836-6}
}

@article{feng.etal2024,
  author   = {Feng, Hao and Liu, Shaokai and Deng, Jiajun and Zhou, Wengang and Li, Houqiang},
  journal  = {IEEE Transactions on Multimedia},
  title    = {Deep Unrestricted Document Image Rectification},
  year     = {2024},
  volume   = {26},
  number   = {},
  pages    = {6142-6154},
  doi      = {10.1109/TMM.2023.3347094}
}

@misc{flux2,
  author  = {{Black Forest Labs}},
  title   = {Flux.2 Klein 9B},
  year    = {2026},
  url     = {https://huggingface.co/black-forest-labs/FLUX.2-klein-9B},
  urldate = {2026-02-28}
}

@misc{gemini,
  author  = {{Google}},
  title   = {Gemini 3 Flash Preview},
  year    = {2026},
  url     = {https://ai.google.dev/gemini-api/docs/models/gemini-3-flash-preview},
  urldate = {2026-02-28}
}

@article{hertlein.etal2023,
  author     = {Hertlein, Felix and Naumann, Alexander and Philipp, Patrick},
  title      = {Inv3D: a high-resolution 3D invoice dataset for template-guided single-image document unwarping},
  year       = {2023},
  issue_date = {Sep 2023},
  publisher  = {Springer-Verlag},
  address    = {Berlin, Heidelberg},
  volume     = {26},
  number     = {3},
  issn       = {1433-2833},
  doi        = {10.1007/s10032-023-00434-x},
  journal    = {Int. J. Doc. Anal. Recognit.},
  month      = apr,
  pages      = {175–186},
  numpages   = {12}
}

@misc{imslp,
  author  = {{IMSLP}},
  title   = {International Music Score Library Project (IMSLP) / Petrucci Music Library},
  year    = {2026},
  url     = {https://imslp.org/},
  urldate = {2026-02-28}
}

@inproceedings{jiang.etal2022,
  author    = {Jiang, Xiangwei and Long, Rujiao and Xue, Nan and Yang, Zhibo and Yao, Cong and Xia, Gui-Song},
  title     = {Revisiting Document Image Dewarping by Grid Regularization},
  booktitle = {Proceedings of the IEEE/CVF Conference on Computer Vision and Pattern Recognition (CVPR)},
  month     = {June},
  year      = {2022},
  pages     = {4543-4552}
}

@article{koo.etal2009,
  author   = {Koo, Hyung Il and Kim, Jinho and Cho, Nam Ik},
  journal  = {IEEE Transactions on Image Processing},
  title    = {Composition of a Dewarped and Enhanced Document Image From Two View Images},
  year     = {2009},
  volume   = {18},
  number   = {7},
  pages    = {1551-1562},
  doi      = {10.1109/TIP.2009.2019301}
}

@inproceedings{kumari.das2025,
  author    = {Kumari, Pooja and Das, Sukhendu},
  title     = {Document Image Rectification using Stable Diffusion Transformer},
  booktitle = {Proceedings of the IEEE/CVF Conference on Computer Vision and Pattern Recognition (CVPR) Workshops},
  month     = {June},
  year      = {2025},
  pages     = {3426-3435}
}

@article{li.etal2019,
  author     = {Li, Xiaoyu and Zhang, Bo and Liao, Jing and Sander, Pedro V.},
  title      = {Document rectification and illumination correction using a patch-based CNN},
  year       = {2019},
  issue_date = {December 2019},
  publisher  = {Association for Computing Machinery},
  address    = {New York, NY, USA},
  volume     = {38},
  number     = {6},
  issn       = {0730-0301},
  doi        = {10.1145/3355089.3356563},
  journal    = {ACM Trans. Graph.},
  month      = nov,
  articleno  = {168},
  numpages   = {11}
}

@inproceedings{li.etal2023,
  author    = {Li, Heng and Wu, Xiangping and Chen, Qingcai and Xiang, Qianjin},
  title     = {Foreground and Text-lines Aware Document Image Rectification},
  booktitle = {Proceedings of the IEEE/CVF International Conference on Computer Vision (ICCV)},
  month     = {October},
  year      = {2023},
  pages     = {19574-19583}
}

@inproceedings{li.etal2023b,
  author    = {Li, Zinuo and Chen, Xuhang and Pun, Chi-Man and Cun, Xiaodong},
  title     = {High-Resolution Document Shadow Removal via A Large-Scale Real-World Dataset and A Frequency-Aware Shadow Erasing Net},
  booktitle = {Proceedings of the IEEE/CVF International Conference on Computer Vision (ICCV)},
  month     = {October},
  year      = {2023},
  pages     = {12449-12458}
}

@article{liang.etal2008,
  author   = {Liang, Jian and DeMenthon, Daniel and Doermann, David},
  journal  = {IEEE Transactions on Pattern Analysis and Machine Intelligence},
  title    = {Geometric Rectification of Camera-Captured Document Images},
  year     = {2008},
  volume   = {30},
  number   = {4},
  pages    = {591-605},
  doi      = {10.1109/TPAMI.2007.70724}
}

@inproceedings{lin.etal2020,
  author    = {Lin, Yun-Hsuan and Chen, Wen-Chin and Chuang, Yung-Yu},
  title     = {BEDSR-Net: A Deep Shadow Removal Network From a Single Document Image},
  booktitle = {Proceedings of the IEEE/CVF Conference on Computer Vision and Pattern Recognition (CVPR)},
  month     = {June},
  year      = {2020}
}

@misc{liu.etal2026,
  title         = {BookNet: Book Image Rectification via Cross-Page Attention Network},
  author        = {Shaokai Liu and Hao Feng and Bozhi Luan and Min Hou and Jiajun Deng and Wengang Zhou},
  year          = {2026},
  eprint        = {2601.21938},
  archiveprefix = {arXiv},
  primaryclass  = {cs.CV},
  url           = {https://arxiv.org/abs/2601.21938}
}

@inproceedings{ma.etal2018,
  author    = {Ma, Ke and Shu, Zhixin and Bai, Xue and Wang, Jue and Samaras, Dimitris},
  title     = {DocUNet: Document Image Unwarping via a Stacked U-Net},
  booktitle = {Proceedings of the IEEE Conference on Computer Vision and Pattern Recognition (CVPR)},
  month     = {June},
  year      = {2018}
}

@inproceedings{ma.etal2022,
  author    = {Ma, Ke and Das, Sagnik and Shu, Zhixin and Samaras, Dimitris},
  title     = {Learning From Documents in the Wild to Improve Document Unwarping},
  year      = {2022},
  isbn      = {9781450393379},
  publisher = {Association for Computing Machinery},
  address   = {New York, NY, USA},
  doi       = {10.1145/3528233.3530756},
  booktitle = {ACM SIGGRAPH 2022 Conference Proceedings},
  articleno = {34},
  numpages  = {9},
  location  = {Vancouver, BC, Canada},
  series    = {SIGGRAPH '22}
}

@inproceedings{markovitz.etal2020,
  author    = {Markovitz, Amir and Lavi, Inbal and Perel, Or and Mazor, Shai and Litman, Roee},
  title     = {Can You Read Me Now? Content Aware Rectification Using Angle Supervision},
  year      = {2020},
  isbn      = {978-3-030-58609-6},
  publisher = {Springer-Verlag},
  address   = {Berlin, Heidelberg},
  doi       = {10.1007/978-3-030-58610-2_13},
  booktitle = {Computer Vision – ECCV 2020: 16th European Conference, Glasgow, UK, August 23–28, 2020, Proceedings, Part XII},
  pages     = {208–223},
  numpages  = {16},
  location  = {Glasgow, United Kingdom}
}

@inproceedings{meng.etal2014,
  author    = {Meng, Gaofeng and Wang, Ying and Qu, Shenquan and Xiang, Shiming and Pan, Chunhong},
  booktitle = {2014 IEEE Conference on Computer Vision and Pattern Recognition},
  title     = {Active Flattening of Curved Document Images via Two Structured Beams},
  year      = {2014},
  volume    = {},
  number    = {},
  pages     = {3890-3897},
  doi       = {10.1109/CVPR.2014.497}
}

@inproceedings{meng.etal2018,
  author    = {Meng, Gaofeng and Su, Yuanqi and Wu, Ying and Xiang, Shiming and Pan, Chunhong},
  title     = {Exploiting Vector Fields for Geometric Rectification of Distorted Document Images},
  booktitle = {Proceedings of the European Conference on Computer Vision (ECCV)},
  month     = {September},
  year      = {2018}
}

@article{narain.etal2012,
  author     = {Narain, Rahul and Samii, Armin and O'Brien, James F.},
  title      = {Adaptive anisotropic remeshing for cloth simulation},
  year       = {2012},
  issue_date = {November 2012},
  publisher  = {Association for Computing Machinery},
  address    = {New York, NY, USA},
  volume     = {31},
  number     = {6},
  issn       = {0730-0301},
  doi        = {10.1145/2366145.2366171},
  journal    = {ACM Trans. Graph.},
  month      = nov,
  articleno  = {152},
  numpages   = {10}
}

@article{narain.etal2013,
  author     = {Narain, Rahul and Pfaff, Tobias and O'Brien, James F.},
  title      = {Folding and crumpling adaptive sheets},
  year       = {2013},
  issue_date = {July 2013},
  publisher  = {Association for Computing Machinery},
  address    = {New York, NY, USA},
  volume     = {32},
  number     = {4},
  issn       = {0730-0301},
  doi        = {10.1145/2461912.2462010},
  journal    = {ACM Trans. Graph.},
  month      = jul,
  articleno  = {51},
  numpages   = {8}
}

@misc{openstax,
  author  = {{OpenStax}},
  title   = {{OpenStax}},
  year    = {2026},
  url     = {https://openstax.org/},
  urldate = {2026-02-28}
}

@inproceedings{perlin1985,
  author    = {Perlin, Ken},
  title     = {An image synthesizer},
  year      = {1985},
  isbn      = {0897911660},
  publisher = {Association for Computing Machinery},
  address   = {New York, NY, USA},
  doi       = {10.1145/325334.325247},
  booktitle = {Proceedings of the 12th Annual Conference on Computer Graphics and Interactive Techniques},
  pages     = {287–296},
  numpages  = {10},
  series    = {SIGGRAPH '85}
}

@inproceedings{quan.etal2024,
  author    = {Quan, Jiahao and Wang, Hailing and Wu, Chunwei and Cao, Guitao},
  booktitle = {2024 IEEE International Conference on Systems, Man, and Cybernetics (SMC)},
  title     = {DLE: Document Illumination Correction with Dynamic Light Estimation},
  year      = {2024},
  volume    = {},
  number    = {},
  pages     = {3701-3707},
  doi       = {10.1109/SMC54092.2024.10831684}
}

@inproceedings{radford.etal2021,
  title     = {Learning Transferable Visual Models From Natural Language Supervision},
  author    = {Radford, Alec and Kim, Jong Wook and Hallacy, Chris and Ramesh, Aditya and Goh, Gabriel and Agarwal, Sandhini and Sastry, Girish and Askell, Amanda and Mishkin, Pamela and Clark, Jack and Krueger, Gretchen and Sutskever, Ilya},
  booktitle = {Proceedings of the 38th International Conference on Machine Learning},
  pages     = {8748--8763},
  year      = {2021},
  editor    = {Meila, Marina and Zhang, Tong},
  volume    = {139},
  series    = {Proceedings of Machine Learning Research},
  month     = {18--24 Jul},
  publisher = {PMLR},
  url       = {https://proceedings.mlr.press/v139/radford21a.html}
}

@misc{RealKIE,
  title         = {RealKIE: Five Novel Datasets for Enterprise Key Information Extraction},
  author        = {Benjamin Townsend and Madison May and Katherine Mackowiak and Christopher Wells},
  year          = {2025},
  eprint        = {2403.20101},
  archiveprefix = {arXiv},
  primaryclass  = {cs.CL},
  url           = {https://arxiv.org/abs/2403.20101}
}

@inproceedings{tang.etal2024,
  author    = {Tang, Hao and Guo, Junyuan and Wang, Teng and Yu, Yanwei and Wang, Chao},
  booktitle = {ICASSP 2024 - 2024 IEEE International Conference on Acoustics, Speech and Signal Processing (ICASSP)},
  title     = {Efficient Joint Rectification of Photometric and Geometric Distortions in Document Images},
  year      = {2024},
  volume    = {},
  number    = {},
  pages     = {3690-3694},
  doi       = {10.1109/ICASSP48485.2024.10447446}
}

@inproceedings{tian.narasimhan2011,
  author    = {Tian, Yuandong and Narasimhan, Srinivasa G.},
  booktitle = {CVPR 2011},
  title     = {Rectification and 3D reconstruction of curved document images},
  year      = {2011},
  volume    = {},
  number    = {},
  pages     = {377-384},
  doi       = {10.1109/CVPR.2011.5995540}
}

@inproceedings{tsoi.brown2007,
  author    = {Tsoi, Yau-Chat and Brown, Michael S.},
  booktitle = {2007 IEEE Conference on Computer Vision and Pattern Recognition},
  title     = {Multi-View Document Rectification using Boundary},
  year      = {2007},
  volume    = {},
  number    = {},
  pages     = {1-8},
  doi       = {10.1109/CVPR.2007.383251}
}

@inproceedings{ulges.etal2004,
  author    = {Ulges, Adrian and Lampert, Christoph H. and Breuel, Thomas},
  title     = {Document capture using stereo vision},
  year      = {2004},
  isbn      = {1581139381},
  publisher = {Association for Computing Machinery},
  address   = {New York, NY, USA},
  doi       = {10.1145/1030397.1030434},
  booktitle = {Proceedings of the 2004 ACM Symposium on Document Engineering},
  pages     = {198–200},
  numpages  = {3},
  location  = {Milwaukee, Wisconsin, USA},
  series    = {DocEng '04}
}

@inproceedings{vecchio.deschaintre2024,
  author    = {Vecchio, Giuseppe and Deschaintre, Valentin},
  title     = {MatSynth: A Modern PBR Materials Dataset},
  booktitle = {Proceedings of the IEEE/CVF Conference on Computer Vision and Pattern Recognition (CVPR)},
  month     = {June},
  year      = {2024},
  pages     = {22109-22118}
}

@inproceedings{verhoeven.etal2023,
  author    = {Verhoeven, Floor and Magne, Tanguy and Sorkine-Hornung, Olga},
  title     = {UVDoc: Neural Grid-based Document Unwarping},
  year      = {2023},
  isbn      = {9798400703157},
  publisher = {Association for Computing Machinery},
  address   = {New York, NY, USA},
  doi       = {10.1145/3610548.3618174},
  booktitle = {SIGGRAPH Asia 2023 Conference Papers},
  articleno = {110},
  numpages  = {11},
  location  = {Sydney, NSW, Australia},
  series    = {SA '23}
}

@inproceedings{wang.etal2022,
  author    = {Wang, Yonghui and Zhou, Wengang and Lu, Zhenbo and Li, Houqiang},
  title     = {UDoc-GAN: Unpaired Document Illumination Correction with Background Light Prior},
  year      = {2022},
  isbn      = {9781450392037},
  publisher = {Association for Computing Machinery},
  address   = {New York, NY, USA},
  doi       = {10.1145/3503161.3547916},
  booktitle = {Proceedings of the 30th ACM International Conference on Multimedia},
  pages     = {5074–5082},
  numpages  = {9},
  location  = {Lisboa, Portugal},
  series    = {MM '22}
}

@article{wang.etal2024,
  title   = {DocNLC: A Document Image Enhancement Framework with Normalized and Latent Contrastive Representation for Multiple Degradations},
  volume  = {38},
  url     = {https://ojs.aaai.org/index.php/AAAI/article/view/28366},
  doi     = {10.1609/aaai.v38i6.28366},
  number  = {6},
  journal = {Proceedings of the AAAI Conference on Artificial Intelligence},
  author  = {Wang, Ruilu and Xue, Yang and Jin, Lianwen},
  year    = {2024},
  month   = {Mar.},
  pages   = {5563-5571}
}

@inproceedings{worley1996,
  author    = {Worley, Steven},
  title     = {A cellular texture basis function},
  year      = {1996},
  isbn      = {0897917464},
  publisher = {Association for Computing Machinery},
  address   = {New York, NY, USA},
  doi       = {10.1145/237170.237267},
  booktitle = {Proceedings of the 23rd Annual Conference on Computer Graphics and Interactive Techniques},
  pages     = {291–294},
  numpages  = {4},
  series    = {SIGGRAPH '96}
}

@inproceedings{xie.etal2020,
  author    = {Xie, Guo-Wang
               and Yin, Fei
               and Zhang, Xu-Yao
               and Liu, Cheng-Lin},
  editor    = {Bai, Xiang
               and Karatzas, Dimosthenis
               and Lopresti, Daniel},
  title     = {Dewarping Document Image by Displacement Flow Estimation with Fully Convolutional Network},
  booktitle = {Document Analysis Systems},
  year      = {2020},
  publisher = {Springer International Publishing},
  address   = {Cham},
  pages     = {131--144},
  isbn      = {978-3-030-57058-3}
}

@inproceedings{xie.etal2021,
  author    = {Xie, Guo-Wang
               and Yin, Fei
               and Zhang, Xu-Yao
               and Liu, Cheng-Lin},
  editor    = {Llad{\'o}s, Josep
               and Lopresti, Daniel
               and Uchida, Seiichi},
  title     = {Document Dewarping with Control Points},
  booktitle = {Document Analysis and Recognition -- ICDAR 2021},
  year      = {2021},
  publisher = {Springer International Publishing},
  address   = {Cham},
  pages     = {466--480},
  isbn      = {978-3-030-86549-8}
}

@inproceedings{xue.etal2022,
  author    = {Xue, Chuhui and Tian, Zichen and Zhan, Fangneng and Lu, Shijian and Bai, Song},
  title     = {Fourier Document Restoration for Robust Document Dewarping and Recognition},
  booktitle = {Proceedings of the IEEE/CVF Conference on Computer Vision and Pattern Recognition (CVPR)},
  month     = {June},
  year      = {2022},
  pages     = {4573-4582}
}

@article{you.etal2018,
  author   = {You, Shaodi and Matsushita, Yasuyuki and Sinha, Sudipta and Bou, Yusuke and Ikeuchi, Katsushi},
  journal  = {IEEE Transactions on Pattern Analysis and Machine Intelligence},
  title    = {Multiview Rectification of Folded Documents},
  year     = {2018},
  volume   = {40},
  number   = {2},
  pages    = {505-511},
  doi      = {10.1109/TPAMI.2017.2675980}
}

@inproceedings{zhang.etal2007,
  author    = {Zhang, Li and Yip, Andy M. and Tan, Chew Lim},
  title     = {Photometric and geometric restoration of document images using inpainting and shape-from-shading},
  year      = {2007},
  isbn      = {9781577353232},
  publisher = {AAAI Press},
  booktitle = {Proceedings of the 22nd National Conference on Artificial Intelligence - Volume 2},
  pages     = {1121–1126},
  numpages  = {6},
  location  = {Vancouver, British Columbia, Canada},
  series    = {AAAI'07}
}

@article{zhang.etal2008,
  author   = {Zhang, Li and Zhang, Yu and Tan, Chew},
  journal  = {IEEE Transactions on Pattern Analysis and Machine Intelligence},
  title    = {An Improved Physically-Based Method for Geometric Restoration of Distorted Document Images},
  year     = {2008},
  volume   = {30},
  number   = {4},
  pages    = {728-734},
  doi      = {10.1109/TPAMI.2007.70831}
}

@inproceedings{zhang.etal2022,
  author    = {Zhang, Jiaxin and Luo, Canjie and Jin, Lianwen and Guo, Fengjun and Ding, Kai},
  title     = {Marior: Margin Removal and Iterative Content Rectification for Document Dewarping in the Wild},
  year      = {2022},
  isbn      = {9781450392037},
  publisher = {Association for Computing Machinery},
  address   = {New York, NY, USA},
  doi       = {10.1145/3503161.3548214},
  booktitle = {Proceedings of the 30th ACM International Conference on Multimedia},
  pages     = {2805–2815},
  numpages  = {11},
  location  = {Lisboa, Portugal},
  series    = {MM '22}
}

@inproceedings{zhang.etal2024,
  author    = {Zhang, Jiaxin and Peng, Dezhi and Liu, Chongyu and Zhang, Peirong and Jin, Lianwen},
  title     = {DocRes: A Generalist Model Toward Unifying Document Image Restoration Tasks},
  booktitle = {Proceedings of the IEEE/CVF Conference on Computer Vision and Pattern Recognition (CVPR)},
  month     = {June},
  year      = {2024},
  pages     = {15654-15664}
}

@article{zhang.etal2024b,
  author   = {Zhang, Jiaxin and Liang, Lingyu and Ding, Kai and Guo, Fengjun and Jin, Lianwen},
  journal  = {IEEE Transactions on Artificial Intelligence},
  title    = {Appearance Enhancement for Camera-Captured Document Images in the Wild},
  year     = {2024},
  volume   = {5},
  number   = {5},
  pages    = {2319-2330},
  doi      = {10.1109/TAI.2023.3321257}
}

@inproceedings{zhang.etal2025,
  author    = {Zhang, Weiguang and Lu, Huangcheng and Ning, Maizhen and Huang, Xiaowei and Wang, Wei and Huang, Kaizhu and Wang, Qiufeng},
  title     = {DvD: Unleashing a Generative Paradigm for Document Dewarping via Coordinates-based Diffusion Model},
  year      = {2025},
  isbn      = {9798400721373},
  publisher = {Association for Computing Machinery},
  address   = {New York, NY, USA},
  doi       = {10.1145/3757377.3763913},
  booktitle = {Proceedings of the SIGGRAPH Asia 2025 Conference Papers},
  articleno = {62},
  numpages  = {12},
  location  = {Hong Kong},
  series    = {SA Conference Papers '25}
}

@inproceedings{zhao.etal2025,
  author    = {Zhao, Fangmin and Zeng, Weichao and Li, Zhenhang and Yang, Dongbao and Li, Binbin and Bi, Xiaojun and Zhou, Yu},
  title     = {Uni-DocDiff: A Unified Document Restoration Model Based on Diffusion},
  year      = {2025},
  isbn      = {9798400720352},
  publisher = {Association for Computing Machinery},
  address   = {New York, NY, USA},
  doi       = {10.1145/3746027.3755362},
  booktitle = {Proceedings of the 33rd ACM International Conference on Multimedia},
  pages     = {8204–8213},
  numpages  = {10},
  location  = {Dublin, Ireland},
  series    = {MM '25}
}

@inproceedings{zhou.etal2025,
  author    = {Zhou, Xinyue
               and Li, Guanting
               and Jiang, Nanfeng
               and Wang, Da-Han
               and Zhang, Xu-Yao
               and Zhu, ShunZhi},
  editor    = {Antonacopoulos, Apostolos
               and Chaudhuri, Subhasis
               and Chellappa, Rama
               and Liu, Cheng-Lin
               and Bhattacharya, Saumik
               and Pal, Umapada},
  title     = {DocHFormer: Document Image Dewarping via Harmonized Modeling of Hierarchical Priors},
  booktitle = {Pattern Recognition},
  year      = {2025},
  publisher = {Springer Nature Switzerland},
  address   = {Cham},
  pages     = {29--44},
  isbn      = {978-3-031-78119-3}
}

@inproceedings{kingma.ba2015,
  title     = {Adam: A Method for Stochastic Optimization},
  booktitle = {3rd International Conference on Learning Representations, ICLR 2015, San Diego, CA, USA, May 7-9, 2015, Conference Track Proceedings},
  author    = {Kingma, Diederik P. and Ba, Jimmy},
  editor    = {Bengio, Yoshua and LeCun, Yann},
  year      = {2015},
  url       = {http://arxiv.org/abs/1412.6980}
}

@inproceedings{loshchilov.hutter2018,
  title     = {Decoupled Weight Decay Regularization},
  author    = {Ilya Loshchilov and Frank Hutter},
  booktitle = {International Conference on Learning Representations},
  year      = {2019},
  url       = {https://openreview.net/forum?id=Bkg6RiCqY7}
}

@article{feng.etal2025,
  author     = {Feng, Hao and Zhou, Wengang and Deng, Jiajun and Tian, Qi and Li, Houqiang},
  title      = {DocScanner: Robust Document Image Rectification with Progressive Learning},
  year       = {2025},
  issue_date = {Aug 2025},
  publisher  = {Kluwer Academic Publishers},
  address    = {USA},
  volume     = {133},
  number     = {8},
  issn       = {0920-5691},
  url        = {https://doi.org/10.1007/s11263-025-02431-5},
  doi        = {10.1007/s11263-025-02431-5},
  journal    = {Int. J. Comput. Vision},
  month      = may,
  pages      = {5343–5362},
  numpages   = {20}
}

@misc{wang.etal2025a,
  title         = {Axis-Aligned Document Dewarping},
  author        = {Chaoyun Wang and I-Chao Shen and Takeo Igarashi and Caigui Jiang},
  year          = {2025},
  eprint        = {2507.15000},
  archiveprefix = {arXiv},
  primaryclass  = {cs.CV},
  url           = {https://arxiv.org/abs/2507.15000}
}
\end{document}